%% file: GraphLearning-main-file.tex
\documentclass[examplefnt,biber]{nowfnt} 

\usepackage[utf8]{inputenc}
\usepackage{multirow}
\usepackage{tabularx}
\usepackage{makecell}
\usepackage{booktabs}
\usepackage{amsmath}
\usepackage{amssymb}
\usepackage{mathrsfs}
\usepackage{makecell}
\usepackage{forest}
\usepackage{float}
\title{Graph Learning}

\maintitleauthorlist{
Feng Xia \\
RMIT University\\
f.xia@ieee.org
\and
Ciyuan Peng\\
Federation University Australia\\
ciyuan.p@ieee.org
\and
Jing Ren\\
RMIT University\\
jing.ren@ieee.org
\and
Falih Gozi Febrinanto\\
Federation University Australia\\
f.febrinanto@federation.edu.au
\and
Renqiang Luo\\
Jilin University\\
lrenqiang@outlook.com
\and
Vidya Saikrishna\\
Federation University Australia\\
v.saikrishna@federation.edu.au
\and
Shuo Yu\\
Dalian University of Technology\\
shuo.yu@ieee.org
\and
Xiangjie Kong\\
Zhejiang University of Technology\\
xjkong@ieee.org
}

\usepackage{mwe}

\author[1]{Feng Xia}
\author[2]{Ciyuan Peng}
\author[3]{Jing Ren}
\author[4]{Falih Gozi Febrinanto}
\author[5]{Renqiang Luo}
\author[6]{Vidya Saikrishna}
\author[7]{Shuo Yu}
\author[8]{Xiangjie Kong}

\affil[1]{RMIT University; f.xia@ieee.org (Corresponding author)}
\affil[2]{Federation University Australia; ciyuan.p@ieee.org}
\affil[3]{RMIT University; jing.ren@ieee.org}
\affil[4]{Federation University Australia; f.febrinanto@federation.edu.au}
\affil[5]{Jilin University; lrenqiang@outlook.com}
\affil[6]{Federation University Australia; v.saikrishna@federation.edu.au}
\affil[7]{Dalian University of Technology; shuo.yu@ieee.org (Corresponding author)}
\affil[8]{Zhejiang University of Technology; xjkong@ieee.org}

\articledatabox{\nowfntstandardcitation}

\begin{document}

\makeabstracttitle

\begin{abstract}
Graph learning has rapidly evolved into a critical subfield of machine learning and artificial intelligence (AI), enabling the modeling of complex relational data that arises in domains such as social networks, biology, finance, transportation, and knowledge representation. Its development began with early graph-theoretic methods, gaining significant momentum with the advent of graph neural networks (GNNs). Over the past decade, progress in scalable architectures, dynamic graph modeling, multimodal learning, generative AI, explainable AI (XAI), and responsible AI has broadened the applicability of graph learning to various challenging environments. Graph learning is significant due to its ability to model complex, non-Euclidean relationships that traditional machine learning struggles to capture, thus better supporting real-world applications ranging from drug discovery and fraud detection to recommender systems and scientific reasoning. However, challenges like scalability, generalization, heterogeneity, interpretability, and trustworthiness must be addressed to unlock its full potential. This survey provides a comprehensive introduction to graph learning, focusing on key dimensions including scalable, temporal, multimodal, generative, explainable, and responsible graph learning. We review state-of-the-art techniques for efficiently handling large-scale graphs, capturing dynamic temporal dependencies, integrating heterogeneous data modalities, generating novel graph samples, and enhancing interpretability to foster trust and transparency. We also explore ethical considerations, such as privacy and fairness, to ensure responsible deployment of graph learning models. Additionally, we identify and discuss emerging topics, highlighting recent integration of graph learning and other AI paradigms and offering insights into future directions. This survey serves as a valuable resource for researchers and practitioners seeking to navigate the rapidly evolving landscape of graph learning.

\end{abstract}

\chapter{Introduction}\label{chapter_into}
\input{chapters/1-Introduction}

\chapter{Scalable Graph Learning}\label{chapter_sgl}
\input{chapters/2-sgl}

\chapter{Temporal Graph Learning}\label{chapter_tgl}
\input{chapters/3-tgl}

\chapter{Multimodal Graph Learning}\label{chapter_mgl}
\input{chapters/4-mgl}

\chapter{Generative Graph Learning}\label{chapter_ggl}
\input{chapters/5-ggl}

\chapter{Explainable Graph Learning}
\input{chapters/6-xgl}

\chapter{Responsible Graph Learning}
\input{chapters/7-rgl}

\chapter{Emerging Topics}
\input{chapters/8-et}

\chapter{Conclusion}
\input{chapters/9-Conclusion}

\begin{acknowledgements}
The authors are grateful to Yicong Li, Tingting Wang, Yingbo Wang, Yanting Li, Yuhao Li (Dalian University of Technology), Wenchao Weng, Wenyi Zhang, Chenhao Ni (Zhejiang University of Technology), and Huafei Huang (University of South Australia) for their assistance with drafting and editing early versions of certain sections of this survey. 
\end{acknowledgements}

\backmatter  

\printbibliography

\end{document}

%% file: chapters/1-Introduction.tex

\begin{figure}[ht]
  \centering
  \includegraphics[width=0.9\linewidth]{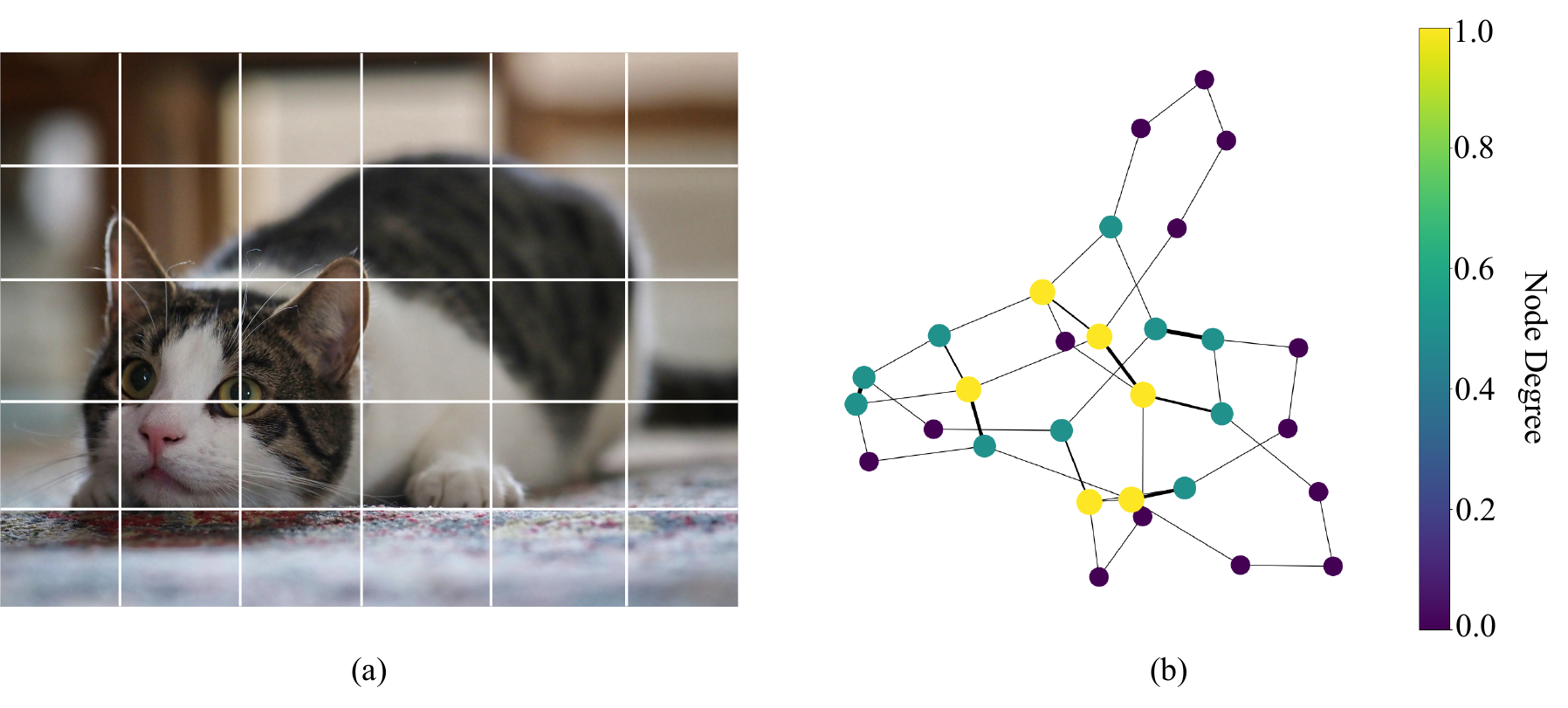}
  \caption{Examples of Euclidean Data and Graph Data. (a) Image of Euclidean data and (b) image of non-Euclidean data.}
  \label{Example of Euclidean data and graph data.}
\end{figure}

Graph learning has emerged as a pivotal artificial intelligence (AI) solution, driven by the increasing recognition of graph structures in modeling complex relationships within data~\citep{xia2021graph_GLsurvey, DBLP:journals/tbd/ZhangYZZ20, shehzad2024graph}. Graphs, which consist of nodes and edges, are inherently suited to represent a diverse range of complex systems such as social systems~\citep{zhou2023hierarchical_socialNetworks, yucage}, knowledge graphs~\citep{peng2023knowledge_knowledgeGraph,10880109}, academic networks~\citep{liu2019shifu2,liu2022deep}, urban dynamics~\citep{kong2022exploring_transportation,10992276}, and biological networks~\citep{yu2024long,peng2025biologically}. Unlike other Euclidean data such as images and text, graphs are typically non-Euclidean data, which exhibit irregular structures, such as varying node degrees and edge weights (see Fig.~\ref{Example of Euclidean data and graph data.}). 

Graph learning techniques are specifically designed to leverage the structural information embedded in graphs. Besides, graph learning is not limited to providing solutions for graph-structured data, but also Euclidean data \citep{DBLP:conf/cvpr/BrynteIOK24}. With nodes and relations extracted from Euclidean data, the graph structure can be first constructed and then be represented by graph learning solutions. For example, image segmentation and object detection can be implemented with graph learning by constructing a connection graph between pixels \citep{DBLP:conf/aaai/HeWZZL22,DBLP:conf/cvpr/BrynteIOK24}. Moreover, graph learning can be applied to disease prediction by constructing a similarity graph of symptoms between patients. The efficiency of analyzing graphs is closely tied to how they are represented. 

Generally speaking, graph learning refers to machine learning on graphs, and its target is to extract the desired features of a graph. Most graph learning approaches utilize deep learning techniques to encode and represent graph data as vectors in a continuous space. This representation allows graphs to be seamlessly integrated into downstream tasks, making graph learning a highly effective tool for various AI applications.  

\begin{figure}[ht]
  \centering
  \includegraphics[width=1\linewidth]{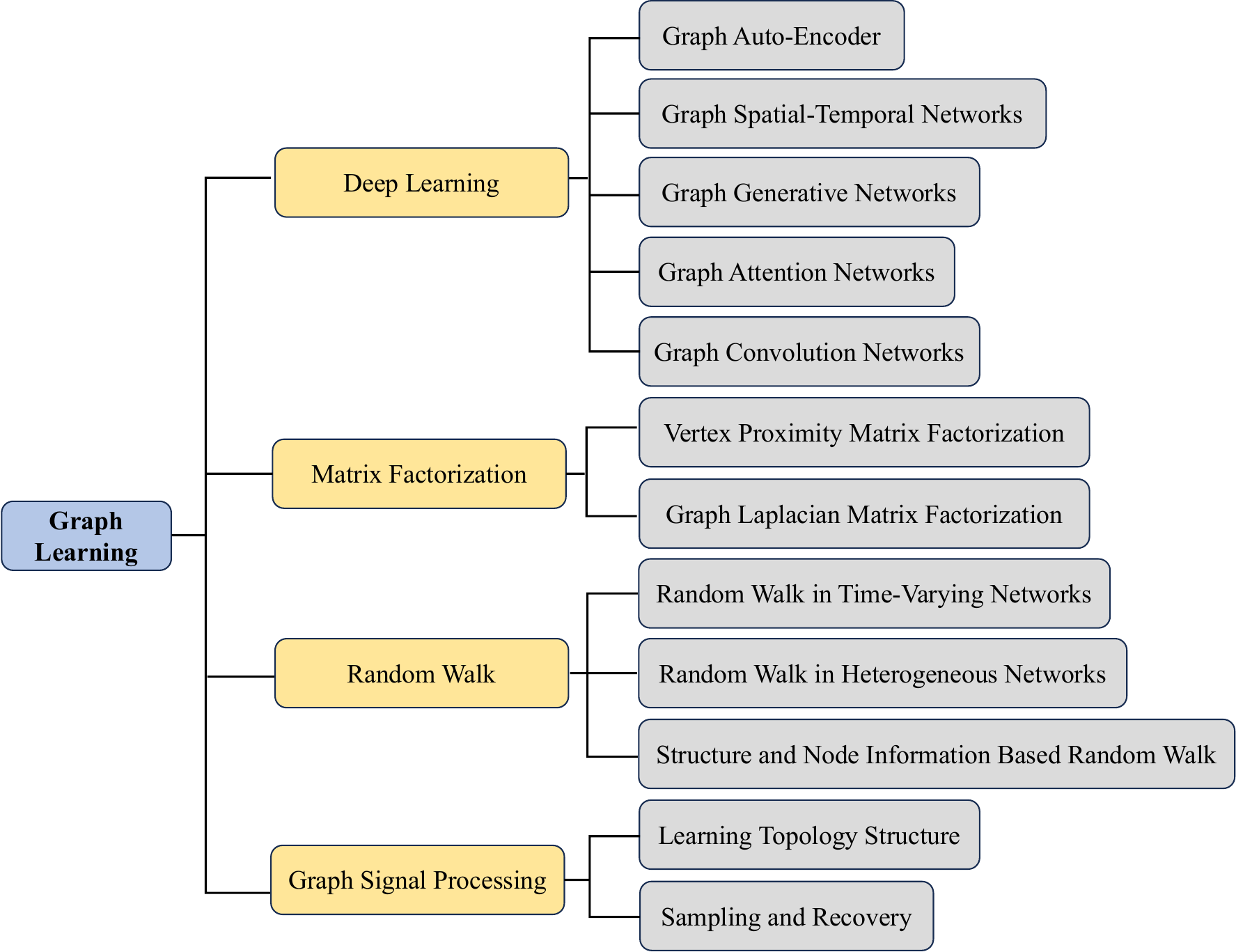}
  \caption{Some Representative Graph Learning Methods.}
  \label{fig:gl_categorization}
\end{figure}

As illustrated in Figure~\ref{fig:gl_categorization}, existing graph learning methods fall roughly into the following four categories~\citep{xia2021graph_GLsurvey}: methods based on deep learning, matrix factorization, random walk, and graph signal processing (GSP). Deep learning-based methods include, for example, graph convolutional networks (GCNs), graph attention networks (GATs), graph auto-encoders (GAEs), graph generative networks, and graph spatial-temporal networks ~\citep{DBLP:journals/tcyb/PanHFLJZ20,DBLP:journals/corr/BrunaZSL13,DBLP:conf/nips/VaswaniSPUJGKP17,luan2024heterophilic}. Matrix factorization techniques can be categorized into two main types: graph Laplacian matrix factorization and vertex proximity matrix factorization ~\citep{choquette2024amplified,DBLP:books/ox/07/GolubR07,wan2023z,li2025revisiting}. Random walk-based methods include, for example, structure-based random walks, random walks that incorporate both structure and node information, random walks in heterogeneous networks, and those in time-varying networks ~\citep{DBLP:conf/ijcai/YangLZSC15,DBLP:journals/ml/LaoC10,nguyen2018continuous_CDNE,xia2019random_randomWalk}. Finally, GSP focuses on the sampling and recovery of graphs, as well as inferring the topological structure of data ~\citep{levie2024graphon,DBLP:journals/tsp/SandryhailaM13,DBLP:journals/tsp/ChenVSK15,DBLP:conf/icassp/NarangGO13}. It is worth noting that graph learning~\citep{xia2021graph_GLsurvey} and graph neural networks (GNNs)~\citep{corso2024graph,zhou2022graph} are not exactly the same thing, with GNNs being a popular type of graph learning approaches. 

To better illustrate diverse graph learning approaches, we start with a classic task: community detection.  In social networks, community detection aims to identify groups of users who interact more frequently with each other than with those outside their group. We frame the community detection task using the following graph learning methods.
\begin{itemize}
    \item \textbf{Deep learning based methods:} Consider a social network represented as a graph \( G = (V, E) \), where \( V \) is the set of nodes (vertices) and \( E \) is the set of edges (links). Each edge \( (u, v) \) can be weighted to reflect the frequency or strength of interaction between users \( u \) and \( v \). GNNs further extend these concepts through a message-passing paradigm~\citep{xia2022cengcn}. In a GNN, each node iteratively aggregates information from its neighbors. This framework allows the model to learn representations that are sensitive to the topology of the graph. Formally, in layer \( l \), the representation \( h_v^{(l)} \) for the node \( v \) is updated as:
\begin{equation}
    h_v^{(l)} = \sigma\left( W^{(l)} h_v^{(l-1)} + \sum_{u \in N(v)} \frac{1}{c_{uv}} W^{(l)} h_u^{(l-1)} \right),
\end{equation}

where \( N(v) \) represents the neighbors of \( v \), \( W^{(l)} \) represents learnable weights, and \( c_{uv} \) is the normalization factor. This iterative process enables the model to capture intricate community structures by considering multi-hop neighbor interactions. 

GATs adaptively assign importance to neighboring nodes, unlike GNNs, which treat all connections with equal weight~\citep{velickovic2018graph}. To improve the model's stability and expressiveness, GAT often employs a multi-head attention mechanism. For each head \(k\), different weight matrices and attention vectors are used to calculate distinct attention coefficients and new node characteristics. The representation \(h_v^{new}\) for the node \(v\) is updated as: 

\begin{equation}
h_v^{new}= \sigma \left( \frac{1}{K} \sum_{k=1}^K \sum_{u\in N(v)} \alpha_{uv}^k 
h_u^{’} \right),
\end{equation}

where \(N(v)\) represents the neighbors of \(v\), \(K\) represents the number of heads, \(\alpha_{uv}^k \) is the normalized attention coefficients, and \(h_u^{’}\) is the original feature vector. This iterative process allows the model to focus on the most informative connections, thereby enhancing its performance.

GAEs contain encoders and decoders. In GAEs, GNNs are employed as encoders to embed network vertices into low-dimensional vectors, and decoders convert these vectors into reconstructed graph data. Specific auto-encoders for graphs are designed by combining either GCN with Generative Adversarial Network (GAN) or Long Short-Term Memory (LSTM) with GAN. GAEs can learn both topology structure and content features ~\citep{DBLP:conf/aaai/CaoLX16,DBLP:conf/kdd/WangC016}. The loss function of GAEs is the mean squared error loss represented as:

\begin{equation}
\mathcal{L} = \frac{1}{N^2} \sum_{u,v=1}^N {(A_{uv}-\hat{A}_{uv})}^2,
\end{equation}

where \(N\) is the number of nodes, \(A_{uv}\) is the adjacency matrix, and \(\hat{A}_{uv}\) is the reconstructed adjacency matrix. To enhance the robustness of graph embeddings, the Adversarially Regularized Variational Graph Autoencoder (ARVGA) introduces an adversarial training process \citep{thomas2016variational}.

In reality, community structures within graphs are often incomplete. To address this, graph generative networks have been developed to generate graphs based on observed data. One category of graph generative networks considers the generation process as the formation of vertices and edges, and another category of them is to employ generative adversarial training.   
  
Social networks evolve over time, with each vertex's attributes changing dynamically. Graph spatial-temporal networks are designed to capture both spatial and temporal dependencies within graphs. Recent studies~\citep{bhatti2023deep} have integrated GCNs with RNNs, CNNs, or LSTMs to effectively model the spatial structures and temporal evolution of information in networks. RNN effectively captures temporal information to model the dependency of node features over time~\citep{qin2024hierarchically}. LSTM is a variant of RNN to address the vanishing or exploding gradient problems encountered by the traditional RNNs when processing long sequence data \citep{van2020review}.  

    \item \textbf{Matrix factorization based methods:} Matrix factorization techniques utilize spectral graph theory to examine community structures within graphs \citep{wan2023z}. A key tool in this analysis is the Laplacian matrix  \( L \) , defined as  \( L = D - A \) , where  \( D \) represents the degree matrix and  \( A \) is the adjacency matrix. The eigenvalues and eigenvectors of \( L \) offer valuable insights into the graph's connectivity. In spectral clustering, the eigenvectors associated with the smallest eigenvalues are particularly informative, as they can reveal the underlying community structure.  

    In addition to the generalized eigenvalue problem, another matrix factorization approach is the vertex proximity matrix factorization.  This method emphasizes the direct similarities or relationships between nodes, capturing their immediate connections and interactions.   Unlike the Laplacian matrix, which primarily reflects the structural information of nodes, vertex proximity focuses on the relational aspects.  Vertex proximity can be approximated in a low dimensional space, and the objective of preserving vertex proximity is to minimize the error as follows:  
  \begin{equation}
    min_{U,V} \lVert A-UV^T \rVert _F^2,
\end{equation}
    where \(U\) and \(V\) are embeddings in a low dimensional space, $T$ refers to transpose and $F$ represents norm.
    
    \item \textbf{Random walks based methods:} Random walks play a crucial role in understanding the flow of information across the graph~\citep{xia2019random_randomWalk}. Random walk-based methods, which are classified as structure- and node information-based random walk methods, utilize random walks to generate node embeddings to capture local and global structures. Algorithms such as DeepWalk~\citep{perozzi2014deepwalk} and Node2Vec~\citep{grover2016node2vec} work in alignment with such techniques. By optimizing a skip-gram model over these walks, nodes that frequently appear together in walks are embedded closely in the vector space, allowing for effective clustering based on proximity in the embedding space. These structure-based random-walk methods leverage structural information in networks for community detection.

    In addition to network topology, content or label of vertex also improves the accuracy of representation, eventually guiding the random walk process. Structure and node information-based random walks methods regard vertex attribute information as supplementary information. For example, TADW~\citep{DBLP:conf/ijcai/YangLZSC15} and MMDW~\citep{tu2016max} are algorithms based on DeepWalk, leveraging text and labeling information of vertices to enhance the performance, respectively. Thus, the transition probabilities of moving from one node to another are adjusted based on both the structural properties of networks and the attributes of vertex.

    Random walk-based approaches also benefit heterogeneous social networks \citep{DBLP:journals/tkde/LuoBYYHLZ23}, as these networks often have various types of nodes and relations. Compared to meta path-based methods that require prior knowledge for optimal meta path selection, random walk approaches use jump and stay strategies to overcome biases \citep{DBLP:conf/cikm/HusseinYC18,DBLP:conf/aaai/WanDPH20,DBLP:conf/kdd/El-KishkyMPVKEM22}. Especially,  when encountering different types of nodes and relations, random walk methods not only consider immediate neighbors but also capture the global structure of the network. This comprehensive exploration enhances the understanding of interconnections among nodes. 
    
    Networks are dynamic entities, with new vertices and relationships emerging over time, while existing ones may become obsolete. Effectively capturing these temporal behaviors is crucial for comprehensive network analysis. Random walk-based methods have proven instrumental in this context, as they traverse time-series paths in dynamic networks, thus encapsulating essential temporal characteristics~\citep{regnier2023universal,zelenkovski2023random}. 
    
    \item \textbf{Graph signal processing based methods:} 
    GSP provides a new perspective of the spectral analysis of graphs \citep{ortega2018graph}. GSP usually defines graph signal models with adjacency matrices \(A\) (or Laplacian matrices \(L\)) as shifts \citep{DBLP:journals/tsp/SandryhailaM13,DBLP:journals/spm/ShumanNFOV13}. Taking adjacency matrix-based GSP as an example, the eigenvalue decomposition of the shift \(A\) is defined as \(A=V\Lambda V^{-1}\), where \(V\) is the matrix of eigenvectors and \(\Lambda\) is a diagonal matrix of eigenvalues. Eigenvectors corresponding to larger eigenvalues can be used to construct low-frequency filters to capture fundamental characteristics, and eigenvectors associated with smaller eigenvalues can be used to capture the variation between neighbor nodes \citep{DBLP:journals/tgrs/LuoLJH25}.

    Sampling and recovery are another direction in GSP. The sampling problem can be defined as reconstructing signals from samples on a subset of vertices, and the signals in it are usually limited by band \citep{DBLP:journals/tsp/ChenVSK15}. The reconstruction task on graphs can also be interpreted as a data interpolation problem. Least squares reconstruction and smoothness technique are available methods in practice. They define a generative model for signal recovery by Gaussian random field (GRF) and a covariance matrix on graphs. Reconstruction of graph signals can be seen as the maximum posterior inference of GRF with low-rank approximation \citep{DBLP:journals/sigpro/HashemiSVM22}.

Inferring the topology of a graph from data involves estimating the graph's structure, often represented by its Laplacian matrix, to identify missing links between entities. Typically, the graphs under consideration exhibit properties of sparsity and smoothness.  In the context of GSP, smooth signals are predominantly composed of low-frequency components. When such signals are subjected to a graph shift operator, the resulting output tends to emphasize these low-frequency components, effectively filtering out higher-frequency variations.  

\end{itemize}

The community detection task can be framed as a supervised or unsupervised learning problem, depending on the availability of labeled data. Loss functions such as cross-entropy for classification tasks or contrastive loss for embedding tasks guide the optimization process, leveraging techniques like the stochastic gradient descent. By integrating theoretical concepts from graph theory, statistical learning, and neural network architectures, graph learning provides robust methodologies for community detection in social networks. This synthesis of theory and application not only enhances our understanding of user interactions but also informs strategic decisions in marketing and social engagement.

Graph learning offers significant advantages across various domains and thus shows great potential. By effectively modeling intricate relationships among entities, graph learning captures the nuances of interactions that previous methods often overlook. For example, in fields such as social networks, graph learning enhances community detection, enabling the identification of tightly-knit user groups based on interaction patterns. In addition, it facilitates recommendations by analyzing user behavior and suggesting connections that are likely to be meaningful. This capability is particularly valuable in mining implicit relations from explicit interactions \citep{liu2023mirror}. In the realm of biology, graph learning plays a critical role in drug discovery by analyzing protein-protein interaction networks, allowing scientists to uncover potential therapeutic targets and understand the underlying biological processes. Furthermore, the adaptability of graph learning (such as GNNs) makes it possible to learn rich representations of entities while considering their local and global contexts. The iterative process not only enhances the model’s understanding of relationships but also improves its robustness in tasks like node classification and link prediction  \citep{RenGL4AA2023,RenMatching2021,wang2020model}. As the research advances to improve scalability and efficiency, the potential for graph learning to drive insights and innovation across diverse industries becomes increasingly promising. Techniques that enable the processing of large-scale graphs have made it feasible to apply graph learning to real-world problems, from fraud detection in finance to route optimization in transportation. Moreover, the integration of graph learning with other technologies, such as natural language processing (NLP) and computer vision, opens new avenues for developing comprehensive models that leverage both relational and contextual information \citep{ren2024survey,ma2025class,li2025graph,guo2025counterfactual,yu2025graph2text,zeng2025edge,wu2025graph}. 

Graph learning has advanced significantly in recent years, yet several critical challenges remain. For instance, scalability remains a primary concern, as real-world graphs, like social networks or biological systems, often contain billions of nodes and edges, straining computational resources and necessitating efficient algorithms for processing large-scale data \citep{ju2024survey}. Dynamic and temporal graphs \citep{febrinanto2023graph_GLL}, which evolve over time, introduce complexity in modeling time-dependent relationships and maintaining real-time updates, critical for applications like financial fraud detection. Multimodal graphs, integrating diverse data types (e.g., text, images, and numerical features), require robust methods to align and fuse heterogeneous information effectively \citep{peng2024learning}. In parallel, the growing influence of generative AI opens new opportunities and risks for graph learning \citep{li2025generative}, particularly in generating realistic graphs or augmenting sparse data, but ensuring validity and fidelity remains difficult. Explainability is another pressing issue \citep{huang2025sehg,nandan2025graphxai}, as the black-box nature of models like GNNs obscures decision-making, limiting trust in high-stakes domains like healthcare. This ties closely with responsible AI concerns, including, e.g., fairness, robustness, and privacy. For instance, bias in graph structures or node attributes can propagate through models \citep{luo2024algorithmic}, while sensitive relational data demands privacy-preserving graph learning techniques. Addressing these challenges is essential for deploying graph learning systems that are not only effective but also trustworthy and socially responsible.
Furthermore, we offer a summary table as shown in Table~\ref{update:summarytable}. It summarizes each branch, application domains, and representative methods.

This survey provides a comprehensive overview of recent advances in graph learning that tackle the key challenges outlined above. Specifically, we cover scalable graph learning (Section 2), temporal graph learning (Section 3), multimodal graph learning (Section 4), generative graph learning (Section 5), explainable graph learning (Section 6), and responsible graph learning (Section 7). Furthermore, we highlight several emerging topics that are gaining increasing attention in the research community (Section 8).

\begin{table}[htbp]
  \centering
  \caption{Overview of All Sections}
    	\renewcommand\arraystretch{1.5}
\resizebox{\textwidth}{!}{
        \begin{tabular}{|c|c|c|}
    \hline
    \multicolumn{1}{|c|}{\textbf{Taxonomy}} & \multicolumn{1}{c|}{\textbf{Application Domains}} & \textbf{Representative methods} \\
    \midrule
    \multicolumn{1}{|c|}{\multirow{3}[6]{*}{\makecell{Scalable\\Graph Learning}}} & \multicolumn{1}{c|}{\multirow{3}[6]{*}{\makecell{Social Network Analysis, \\Recommendation Systems}}} & Graph Data Summarization \\
\cmidrule{3-3}          &       & Computational Sampling Methods \\
\cmidrule{3-3}          &       & Distributed Graph Learning \\
    \midrule
    \multicolumn{1}{|c|}{\multirow{2}[4]{*}{\makecell{Temporal \\Graph Learning}}} & \multicolumn{1}{c|}{\multirow{2}[4]{*}{\makecell{Real-time Traffic Forecasting, \\Epidemic Spread Prediction}}} & Spatiotemporal Graph Learning \\
\cmidrule{3-3}          &       & Dynamic Graph Learning \\
    \midrule
    \multicolumn{1}{|c|}{\multirow{2}[4]{*}{\makecell{Multimodal \\Graph Learning}}} & \multicolumn{1}{c|}{\multirow{2}[4]{*}{Multimedia Content Analysis}} & Graph-driven Multimodal Learning \\
\cmidrule{3-3}          &       & Learning on Multimodal Graphs \\
    \midrule
    \multicolumn{1}{|c|}{\multirow{2}[4]{*}{\makecell{Generative \\Graph Learning}}} & \multicolumn{1}{c|}{\multirow{2}[4]{*}{\makecell{Drug Discovery,  \\Molecular Material Design}}} & Unconditional Generative Graph Learning \\
\cmidrule{3-3}          &       & Conditional Generative Graph Learning \\
    \midrule
    \multicolumn{1}{|c|}{\multirow{2}[4]{*}{\makecell{Explainable \\Graph Learning}}} & \multicolumn{1}{c|}{\multirow{2}[4]{*}{\makecell{Medical Diagnostics, \\Financial Risk Control}}} & Post-hoc Explanation for Graph Learning \\
\cmidrule{3-3}          &       & Self-explanatory Graph Learning \\
    \midrule
    \multicolumn{1}{|c|}{\multirow{2}[4]{*}{\makecell{Responsible \\Graph Learning}}} & \multicolumn{1}{c|}{\multirow{2}[4]{*}{\makecell{Credit Analysis, \\Judicial Risk Assessment, \\Public Policy Formulation}}} & Privacy-Preserving Graph Learning \\
\cmidrule{3-3}          &       & Fairness in Graph Learning \\
    \bottomrule
    \end{tabular}}
  \label{update:summarytable}%
\end{table}%

%% file: chapters/2-sgl.tex

The rapid increase in the size and complexity of graphs poses significant challenges for graph learning techniques~\citep{duan2022comprehensive_scalableGLIntro,xia2022cengcn,shehzad2025multiscale}. The traditional approach to graph learning is not well-suited for large graphs, primarily due to scalability challenges stemming from high model complexity and excessive memory demands. Firstly, processing large-scale graphs requires immense computational resources. Conventional graph learning methods involve operations like matrix multiplication, which do not scale efficiently as the graph size increases. This can lead to bottlenecks, prolonged processing times, and difficulties in training models within a feasible timeframe. Secondly, graphs with millions of nodes and billions of edges demand substantial memory allocation for storing adjacency matrices and associated features. As the graph size grows, memory consumption can exceed system capacity, making storage and processing impractical.

\begin{figure*}[ht!]
    \centering
    \includegraphics[width=\textwidth]{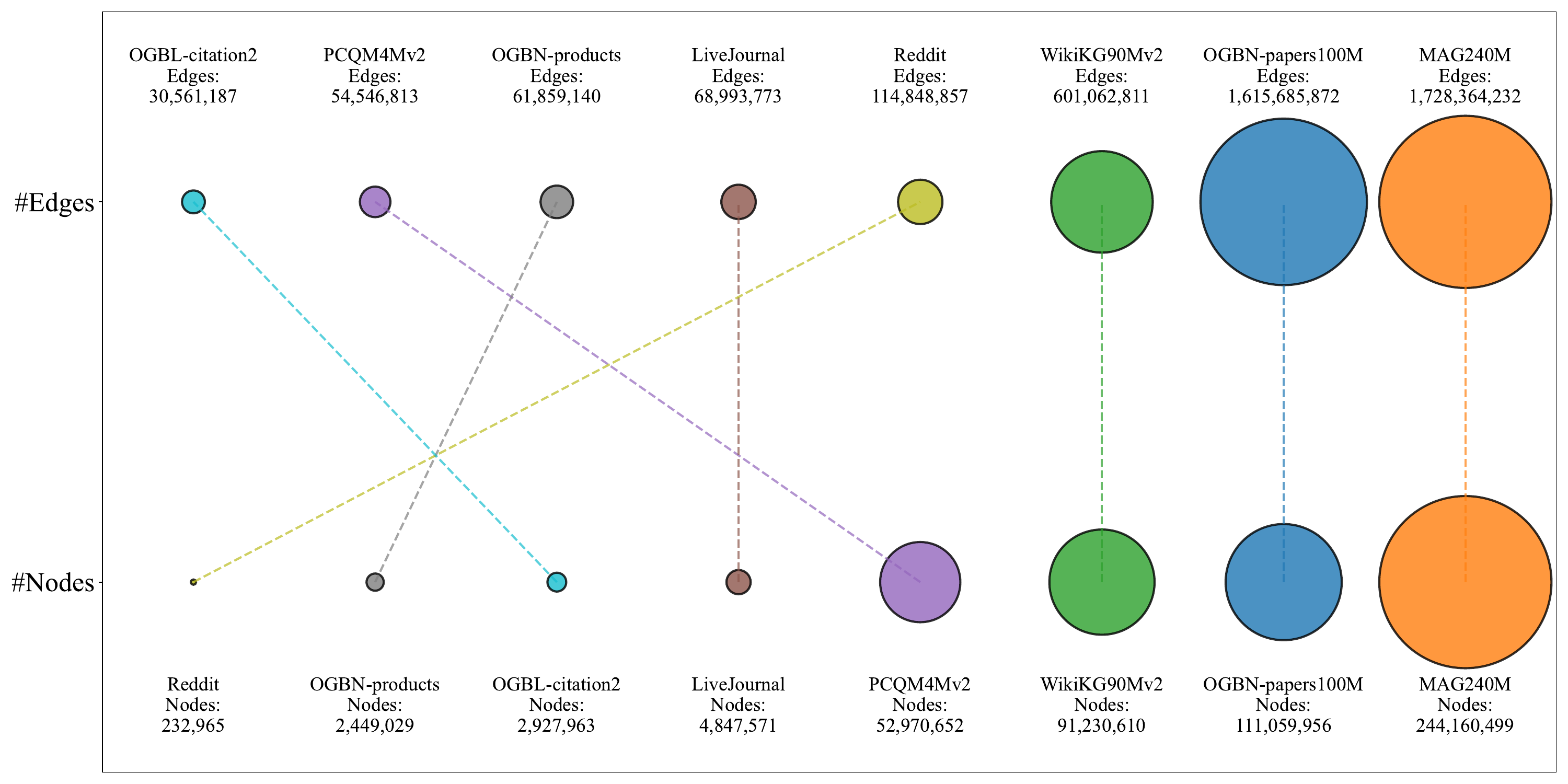}
    \caption{Comparison of Nodes vs. Edges in Large Graph Datasets.}
    \label{fig:comparison_nodes_edges_number}
\end{figure*}

Several large, publicly available graph datasets illustrate the scale and complexity of modern graph learning challenges, as shown in Figure~\ref{fig:comparison_nodes_edges_number}.  These datasets include MAG240M~\citep{hu2020open_MAG240M_OGBN}, OGBN~\citep{hu2020open_MAG240M_OGBN}, WikiKG90Mv2~\citep{hu2ogb_MAG240M_WikiKG}, Reddit~\citep{hamilton2017inductive_reddit_graphSage}, and LiveJournal~\citep{yang2012defining_liveJournal}.  Among them, MAG240M is the largest, containing over 240 million nodes and 1.7 billion edges, making it particularly challenging for traditional graph learning methods. Even smaller datasets, such as Reddit, still consist of a significant number of nodes and edges, highlighting the need for scalable and efficient graph learning techniques.

\begin{figure}[ht]
  \centering
  \includegraphics[width=1\linewidth]{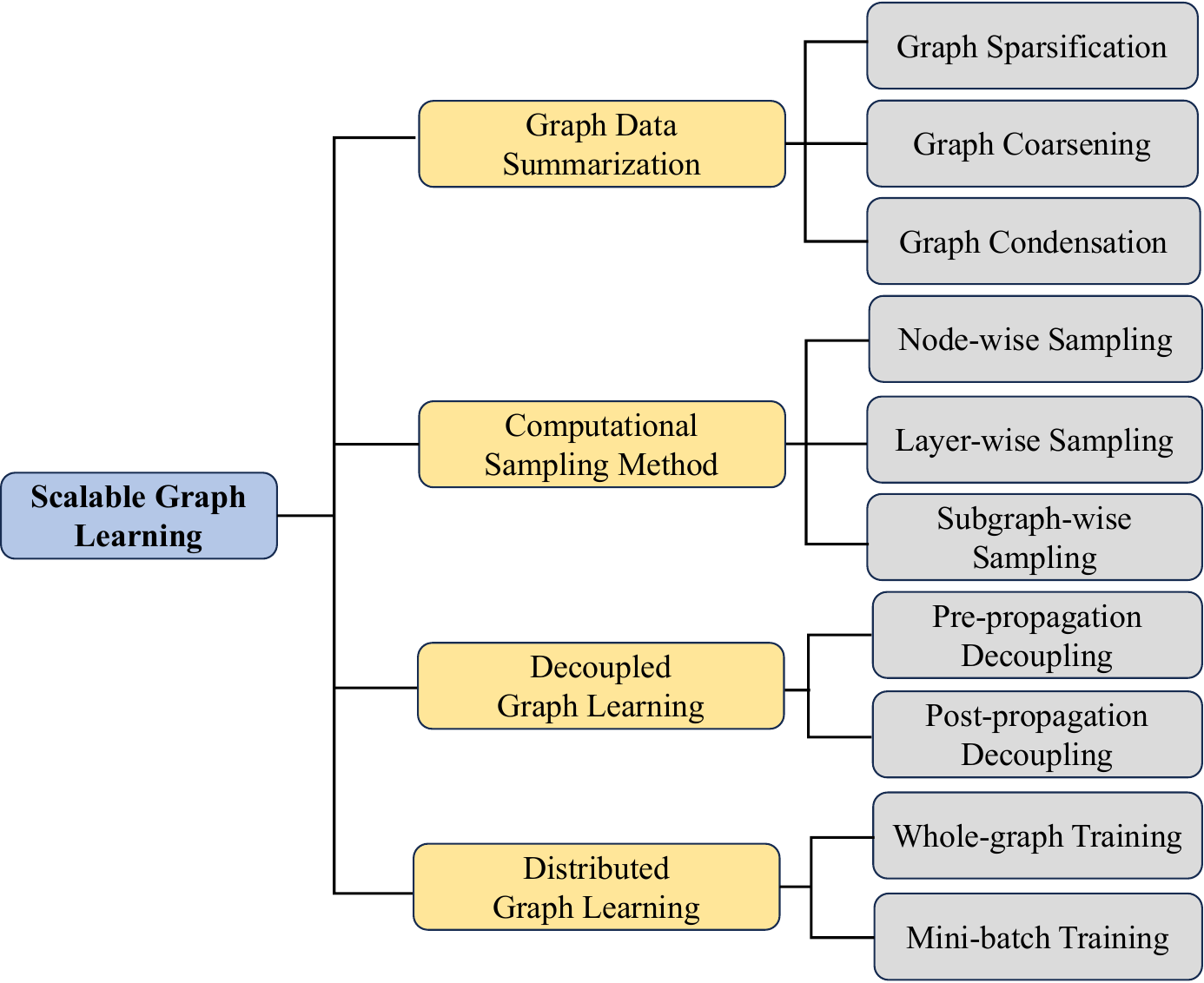}
  \caption{Strategy Taxonomy of Scalable Graph Learning.}
  \label{fig:sgl_categorization}
\end{figure}

The concept of scalable graph learning~\citep{ma2022graph_scalableGLSurvey} has emerged as a solution to address the scalability limitations of traditional graph learning methods. Its primary goal is to efficiently process large-scale static graphs, such as social networks with over 1 billion edges and more than 100 million nodes, especially for tasks like node and edge classification. By optimizing computational efficiency and memory usage, scalable graph learning enables practical applications in massive graphs in the real world.  

In this survey, as illustrated in Figure~\ref{fig:sgl_categorization}, we focus on existing strategies that can be used to improve the scalability of graph learning, which fall into the following four categories: graph data summarization, computational sampling methods, decoupled graph learning, and distributed graph learning. Graph data summarization~\citep{shabani2024comprehensive_graphSummarizationSurvey} deals with graph sparsification, graph condensation, and graph coarsening. Computational sampling methods~\citep{ma2022graph_scalableGLSurvey} include several strategies: node-wise, layer-wise, and subgraph-wise. Decoupled graph learning~\citep{ma2022graph_scalableGLSurvey} can be divided into pre-propagation and post-propagation decoupling. Lastly, distributed graph learning~\citep{besta2024parallel_distributedGraphLearningSurvey} includes whole-graph training and mini-batch training strategies. Essentially, these categorizations differ in solving the scalability challenge in graph learning. This survey presents an extensive review of representative state-of-the-art scalable graph learning techniques.

\begin{figure*}[ht!]
    \centering
    \includegraphics[width=\textwidth]{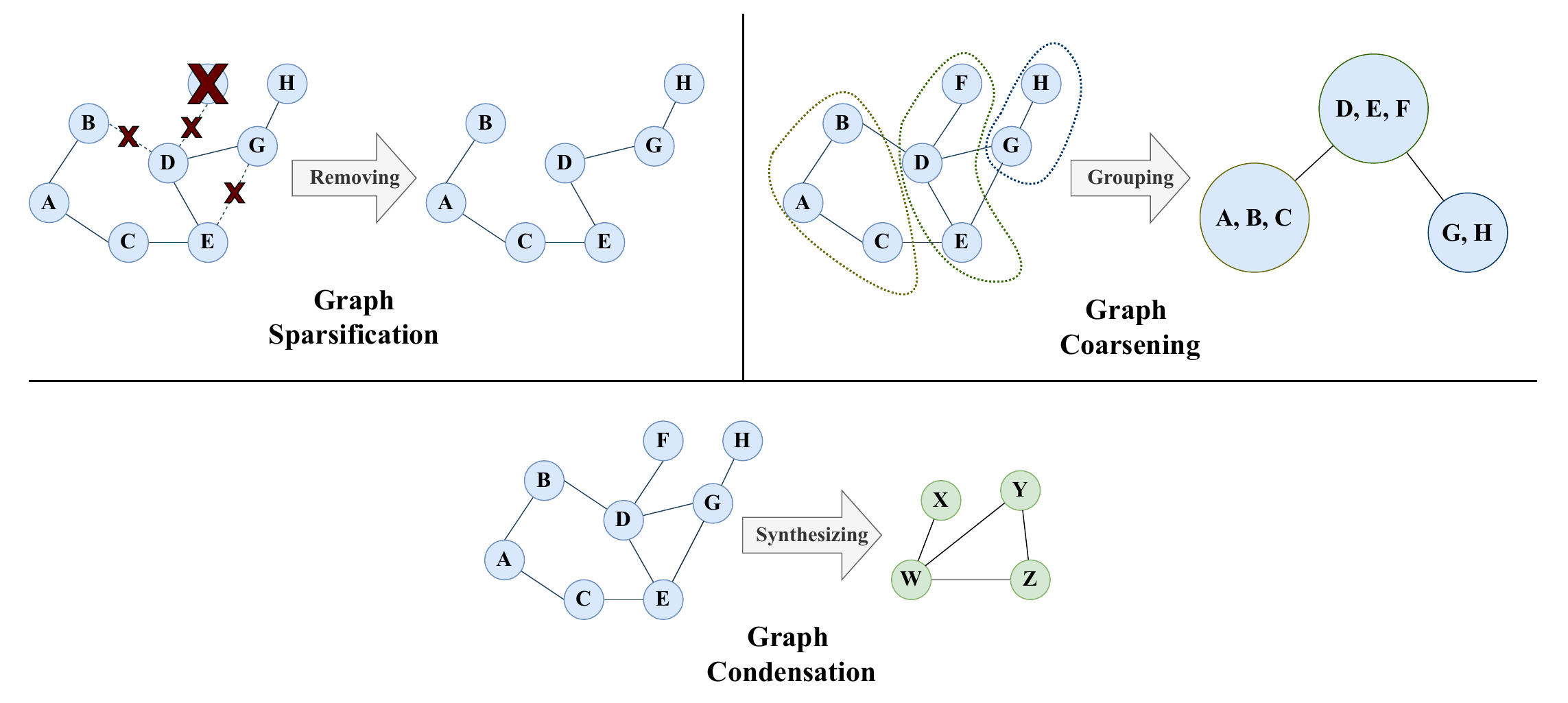}
    \caption{Data Summarization Techniques.}
    \label{fig:sgl_summarization}
\end{figure*}

\section{Graph Data Summarization}
Graph data summarization, also known as graph reduction or simplification, refers to reducing the size of a graph while preserving its essential properties and information~\citep{shabani2024comprehensive_graphSummarizationSurvey}. This process is important for handling large-scale graphs by making a smaller and more manageable graph structure that could enhance scalability in graph learning.

This section categorizes the graph data summarization techniques into three major groups: graph sparsification, condensation, and coarsening. Figure~\ref{fig:sgl_summarization} compares the three graph data summarization strategies. Graph sparsification selects a subset of nodes and edges from the original graph to preserve critical properties, such as essential edges or node centrality~\citep{chen2023demystifying_sparsificationSimplest}. On the other hand, graph coarsening groups and aggregates similar nodes into \textit{super-nodes} and their connections into \textit{super-edges} instead of removing nodes~\citep{chen2023gromov_GWGRAPH}. Graph condensation~\citep{sun2024gc_graphCondensationBenchmark}, a more recent method, differs from sparsification and coarsening by creating a new, smaller synthetic graph that retains the task performance of graph learning while reducing graph size. We discuss each of the techniques in the sub-sections below:

\subsection{Graph Sparsification}
Graph sparsification aims to improve scalability in graph learning by selecting important nodes or edges to result in a smaller graph that still maintains the structural quality of the original graph. Given a graph $ G = (V, E) $, graph sparsification selects representative nodes or edges from the original graph $ G $ to produce a graph $ G' = (V', E') $, where $ V' $ or $ E' $ are subsets of $ V $ or $E $. The simplest way to perform sparsification is by randomly sampling a subset of edges. This technique is called a random sparsifier, which samples all edges in the graph with equal probability~\citep{chen2023demystifying_sparsificationSimplest}.

\textit{Cut and Distance-Preserving Sparsifiers.} Traditional sparsification techniques are based on the concept of cut sparsifiers~\citep{abboud2022friendly_cuts}. These methods aim to preserve the total weight of edges that separate two distinct groups of nodes, known as a cut, ensuring that the sparsified graph closely approximates the original graph's cut properties. This approach reduces the number of edges while maintaining the overall structural integrity of the graph, making it more computationally efficient for various graph learning tasks.~\citet{althofer1990generating_spanner} developed an algorithm called Spanner, which starts with an empty edge set and creates a smaller version of the graph by adding only the most important edges. In the spanner algorithm, the smaller graph retains the approximate distances between nodes, similar to the original graph, but with fewer connections. This means that while the sparsified graph may have significantly fewer edges, the overall structure and relationships between nodes are preserved, allowing for efficient computations and graph learning tasks without losing critical information. They proved that the original graph could be reduced to a smaller graph called a \textit{2t}-Spanner, which has fewer edges, around $ O(N^{1+\frac{1}{t}}) $, where $N$ is the number of nodes.

\textit{Spectral sparsifiers.} Other techniques have also been proposed. Unlike methods that maintain pairwise distances, some focus on preserving cuts or spectral properties, known as spectral sparsifiers~\citep{chen2023demystifying_sparsificationSimplest, spielman2014nearly_spectralSparsifiers}. These methods ensure that the sparsified graph retains important spectral properties, such as those encoded in the graph Laplacian. The effective resistance sparsifier (ER)~\citep{spielman2008graph_ER} is a representative strategy. ER is derived from an analogy with an electrical circuit where edges represent resistors. The method measures the effective resistance of an edge in terms of how much "effort" it takes for electric current to flow from one node to another. ER has demonstrated that the quadratic form of the Laplacian in the sparsified graph remains close to that of the original graph, as captured by the following inequality:

\begin{equation}
\forall x \in \mathbb{R}^{|V|} \quad (1 - \epsilon) x^T L x \leq x^T \tilde{L} x \leq (1 + \epsilon) x^T L x,
\end{equation}
where $x$ is a vector with dimension equal to the number of nodes $|V|$ in the graph, $\tilde{L}$ is the Laplacian of the sparsified graph, and $\epsilon$ is a small number. The expression measures the impact of the graph structure on the vector $x$, showing that this impact remains consistent after the graph is sparsified. The \textit{Twice Ramanujan Sparsifier} (TRS)~\citep{batson2009twice_TRS} is another technique that has proven, for any small number $\epsilon$ between 0 and 1, that it is possible to generate a smaller graph $G'$ from any undirected graph $G$. The sparsified graph $G'$ has fewer edges, with the number of edges being linear in the number of vertices of $G$ while closely approximating the spectral properties of the original graph.

\textit{Similarity-based Sparsifiers.} Another group of sparsifiers incorporates similarity evaluation as a metric to assess the quality of the sparsified graph. For example, methods inspired by the structural clustering algorithm for networks (SCAN)~\citep{xu2007scan_SCAN} use structural similarity measures to decide which edges to retain, keeping those that connect nodes sharing many neighbors while discarding edges with low similarity. If two nodes share many neighbors, their connection is likely retained, as they belong to the same cluster. In contrast, edges between nodes with low similarity are deleted since it is considered less important. Another work, the local Jaccard similarity sparsifier (L-Spar)~\citep{satuluri2011local_LSPAR}, focuses on a localized approach to edge sparsification. L-Spar selects edges for each node based on its local neighborhood structure. It ranks the edges connected to a node by Jaccard similarity and retains the top-ranked ones, ensuring that nodes stay connected to their most similar neighbors. This approach helps preserve local cluster structures while significantly reducing the overall edge set of the graph. 

\textit{Deep learning-based Sparsifiers.} Following the significant advances in deep learning, some approaches combine sparsification strategies with deep learning techniques to improve the performance of downstream tasks. Examples include DropEdge~\citep{rong2019dropedge}, randomly removing edges from the input graph during training in each epoch. NeuralSparse~\citep{zheng2020robust_neuralsparse} develops edge pruning and deep learning to parameterize the sparsification process. SparRL~\citep{wickman2022generic_SPARRL} uses reinforcement learning by adjusting reward functions to perform sparsification while maintaining the quality of pairwise node distances. WIS (Walk Index Sparsification) by~\citet{razin2023ability_wis} removes edges based on the walk index, preserving the ability of graph learning to model interactions between vertices. While graph sparsification effectively reduces the number of edges, it may not drastically reduce the overall graph structure. In contrast, more recent strategies like graph coarsening and condensation can significantly change the original structure. Graph coarsening combines nodes into super-nodes and edges into super-edges, while graph condensation creates new synthetic graph structures inspired by the original graph. 

\subsection{Graph Coarsening}
Graph coarsening is a promising direction for improving the scalability of graph processing by summarizing a very large graph into a smaller, tractable one. The selection of nodes or edges in sparsification methods can lead to the loss of information. Graph coarsening ensures that sufficient valuable information from the graph structure is preserved by grouping and aggregating nodes or edges. Given \( G = (V, E) \), graph coarsening produces new graph \( G' = (V', E') \) which contains super-nodes $V'$ and super-edges $E'$, where $|V'| < |V|$ and $|E'| < |E|$. The coarse graph should be tractable and mapped from the original graph, which can be represented by a one-hot matrix $C \in \{0,1\}^{|V| \times |V' |}$ to denote the membership of a group of nodes. The adjacency matrix of the coarsened graph is $A' = C^\intercal AC$, and the Laplacian matrix is $L' = C^\intercal LC$.

\textit{Classical Coarsening Techniques}. The algebraic multigrid literature has influenced the development of graph coarsening~\citep{chen2023gromov_GWGRAPH}. The classical graph coarsening techniques include some techniques that use traditional graph partitioning, clustering, and spectral properties to group nodes into supernodes while reducing graph size. One example of clustering is spectral clustering~\citep{jin2020graph_spectralClustering}, which leverages the eigenvectors of the graph Laplacian to find dense partitions within the graph, summarizing its structure. An earlier technique, Edge Matching, introduced by~\citet{dhillon2007weighted_edgeMatching}, merges node pairs by maximizing their total edge weight. This approach typically computes the weight for each pair of nodes $(v_i, v_j)$, defined as $\frac{W_{ij}}{\max(d_i, d_j)}$, where $d_i$ and $d_j$ represent the degrees of the nodes. This ensures that significant connections in the graph are preserved during the reduction process. Kron Reduction~\citep{dorfler2012kron} is an algebraic method that reduces a graph by applying the Schur complement to its Laplacian matrix, preserving essential physical properties like effective resistance, which is critical in applications such as electrical networks. CoarseNet~\citep{purohit2014fast_coarseNet} focuses on retaining diffusion properties by grouping nodes based on their influence within the network, enabling substantial graph reduction without losing key information on how processes propagate. Compressive network embedding (COSINE) by~\citet{zhang2020cosine_cosine} performs coarsening by partitioning nodes into groups for parameter sharing in compressive embeddings.~\citet{deng2020graphzoom} introduced GraphZoom, which uses spectral properties to aggregate nodes with high similarity and builds multi-level coarsened graphs to improve the scalability of graph embedding tasks.~\citet{fahrbach2020faster_schurComplement} used Schur Complement and Random Contraction techniques to eliminate or combine nodes into supernodes, maintaining the graph's spectral and structural properties to speed up embedding and simplify graph structure. Similarly, featured graph coarsening (FGC) by~\citet{kumar2023featured_FGC} compresses both node features and graph structure to preserve the node relationships, which is particularly useful for tasks like node classification.

\textit{Deep Learning-Based Coarsening Techniques}. On the other hand, Deep Learning-Based Coarsening techniques such as convolution matching (ConvMatch) and its approximate variant, A-ConvMatch, focus on preserving the outputs of graph convolutions in GNNs~\citep{dickens2024graph_convmatch}. These methods merge nodes based on minimizing the difference in convolution results between the original and coarsened graphs. The k-MIS-based coarsening leverages maximal independent sets to ensure even node distribution~\citep{bacciu2023generalizing_kMIS}. It is applied to downsample graphs for GNN pooling layers (Bacciu et al., 2023). Supervised coarsened adjacency learning (SCAL) by~\citet{huang2021scaling_scal} trains the GNN model on coarsened graphs for downstream tasks, especially node classification. Other specific graph coarsening methods based on deep learning techniques specifically address scalability issues, such as variable node grouping (VNG)~\citep{si2023serving_VNG}, which uses weighted k-means to match forward propagation and generate an effective mapping matrix. For applications with large-scale web graphs,~\citet{cao2024graph_graphSkeleton} proposed Graph-Skeleton fetches and condenses critical background nodes into synthetic supernodes to retain structural and feature correlations essential for classification tasks while reducing the graph size by up to 99\%.

\subsection{Graph Condensation}

Graph condensation is a strategy for creating a smaller, highly informative graph from a larger original graph~\citep{sun2024gc_graphCondensationBenchmark,gao2025graph}. Unlike graph sparsification and coarsening, which simplify the existing graph, graph condensation generates a new synthetic graph designed to achieve comparable performance to the original graph on various downstream tasks. Typically, graph condensation leverages deep learning methods, specifically GNNs, to produce this condensed graph. The primary goal is to reduce the size and complexity of the graph while preserving its essential information.

Given a graph $\mathcal{T} = (V, E)$, the aim is to learn a small synthetic graph $\mathcal{S} = (V', E')$ where $|V'| < |V|$ and $|E'| < |E|$, such that models trained on $\mathcal{S}$ perform comparably to those trained on $\mathcal{T}$. To establish a meaningful representation between the original graph $\mathcal{T}$ and the synthetic graph $\mathcal{S}$, many approaches employ a shared GNN model $\textup{GNN}_\theta$, parameterized by $\theta$ to encode both graphs and compute task-specific losses for each graph input as follows:

\begin{equation} \begin{aligned} \mathcal{L}^\mathcal{T} (\theta) &= \ell \left(\textup{GNN}_\theta(\mathcal{T}), Y\right), \\ \mathcal{L}^\mathcal{S} (\theta) &= \ell \left(\textup{GNN}_\theta(\mathcal{S}), Y'\right), \end{aligned} \end{equation}
where $\ell$ represents the task-specific objective, such as cross-entropy for classification tasks, and $Y$ and $Y'$ are the task-specific labels for the original graph $\mathcal{T}$ and $\mathcal{S}$, respectively. Eventually, the objective of graph condensation can be formulated as a two-level optimization problem, or bi-level problem, as follows:

\begin{equation} \label{eq} \mathcal{S} = \min_\mathcal{S} \mathcal{L}^\mathcal{T}\left(\theta^\mathcal{S}\right) \quad \text{s.t.} \quad \theta^\mathcal{S} = \arg\min_\theta \mathcal{L}^\mathcal{S}(\theta). \end{equation}

Equation~\ref{eq} states that in order to find the best possible smaller graph $\mathcal{S}$ to train a model using optimal parameters $\theta^\mathcal{S}$, the performance, as measured by $\mathcal{L}^\mathcal{T}$, should be as good as possible on the original large graph $\mathcal{T}$. It also requires that the parameters $\theta^\mathcal{S}$ represent the best solution for the smaller graph $\mathcal{S}$ itself, evaluated by $\mathcal{L}^\mathcal{S}$.

\textit{Gradient Matching-based Techniques}. Several popular methods have been proposed to perform graph condensation. The first taxonomy discussed in this survey paper is the gradient matching strategy. Gradient matching is an example in graph condensation where it aims to match the gradient of the model trained on the condensed graph with that of the model trained on the original graph. The first attempt by~\citet{zhao2021dataset_condensationGradientMatching} proposed gradient matching for image classification datasets, where the optimized parameters of these gradient matching models are trained on two datasets. Inspired by an early work of~\citet{zhao2021dataset_condensationGradientMatching}, graph condensation (GCond)~\citep{jingraph_GCond} proposed to match model gradients at each training step $t$. This approach allows the training path on the condensed data to imitate the original data, leading to similar solutions between models trained on both datasets, given the following optimization solution:

\begin{equation} \begin{aligned} \min_\mathcal{S} \mathbb{E}_{\theta_0 \sim \Theta} \left[\sum_{t=0}^{T-1} \mathcal{D} \left( \nabla_{\theta_t} \mathcal{L}^\mathcal{T}(\theta_t), \nabla_{\theta_t} \mathcal{L}^\mathcal{S}(\theta_t) \right) \right], \end{aligned} \end{equation}
where $D(\cdot,\cdot)$ represents the distance calculation, $T$ denotes the training steps, and $t$ represents each step. The equation aims to minimize the difference between the training path of a model on the smaller synthetic graph $\mathcal{S}$ and the original large graph $\mathcal{T}$ overall training iterations or epochs. The objective is to ensure that at each training epoch $t$, the model's gradient on the synthetic graph closely matches the model's gradient on the original graph. One-step gradient matching for graph condensation (DosCond) is proposed by~\citet{jin2022condensing_DosCond}, which accelerates graph condensation by matching network gradients only at the initialization of the model, bypassing the need to track the full training trajectory. It further models the graph structure as a probabilistic entity using Bernoulli distributions. Structure-broadcasting graph dataset distillation (SGDD)~\citep{yang2024does_SGDD} preserves the original graph's structure through graphon approximation and minimizes Laplacian Energy Distribution (LED) shift using optimal transport to optimize the condensed graph. CTRL (CrafTing RationaL Gradient Matching) by~\citet{zhang2024two_CTRL} uses clustering to initialize synthetic graphs and minimizes matching errors by considering both gradient direction and magnitude during condensation.~\citet{fang2024exgc} proposed efficient and explainable graph condensation (EXGC), which refines gradient matching by selecting the most important nodes via the gradient information bottleneck (GDIB) module, using explainability techniques to justify the selections and employing variational approximations to boost efficiency. This combination ensures that the condensed graph preserves the most relevant gradient information while being computationally efficient and has good explainability.

\textit{Trajectory Matching-Based Techniques}. The second taxonomy of graph condensation is trajectory matching. Unlike gradient matching, which focuses on single-step gradient alignment, trajectory matching considers the entire training trajectory of the model. It aims to ensure that the model's learning path on the small graph closely follows that on the original large graph. Structure-free graph condensation (SFGC)\citep{zheng2024structure_SFGC} introduces trajectory matching by aligning the long-term learning behaviors between the original large graph and the condensed, structure-free data, improving knowledge transfer without explicitly maintaining the graph structure. Graph condensation via expanding window matching (GEOM), proposed by~\citet{zhangnavigating_GEOM}, tackles the issue of limited supervision in SFGC, especially when dealing with complex nodes. GEOM evaluates these nodes based on homophily and uses curriculum learning~\citep{wang2021survey_curriculumLearning} to gradually increase the complexity of the matching process. This approach ensures a better transfer of information from both easy and difficult nodes, helping the model generalize more effectively and capture a broader range of graph properties.

\textit{Distribution Matching-Based Techniques}. The third taxonomy of graph condensation is distribution matching. While training a graph learning model on a small condensed graph can be extremely fast, creating a synthetic condensed graph structure from the original graph might be expensive due to complex bi-level optimization and second-order derivative computation. Thus,~\citet{zhao2023dataset_distributionMatching} proposed an alternative optimization principle called distribution matching, which was applied to image data, focusing on matching the feature distribution of the condensed and original data. It aims to generate condensed data with a feature distribution closely approximating the original data using metrics such as Maximum Mean Discrepancy (MMD). A framework called condense and train (CaT)~\citep{liu2023cat_CATdistributionMatching} implements the distribution matching strategy in graph data using the following objective function:

\begin{equation} \begin{aligned} \ell_{\text{MMD}} = \sum_{c \in C} r_c \cdot \left| \text{Mean}(E_c) - \text{Mean}(\tilde{E}_c) \right|^2 \end{aligned}. \end{equation}

Here, $C$ represents the different classes of nodes, $E_c$ and $\tilde{E}_c$ are the embeddings of nodes with class $c$ in the original and condensed graphs, respectively, and $r_c$ is the ratio of class $c$. This approach ensures that the synthetic graph retains essential properties from the original graph. Graph condensation via receptive field distribution matching (GCDM) by~\citet{liu2022graph_GCDM} expands this by matching the distribution of receptive fields between the original and condensed graphs, leveraging Maximum Mean Discrepancy (MMD) to ensure similarity. Furthermore, graph distillation with eigenbasis matching (GDEM)~\citep{liugraph_GDEM} introduces an innovative eigenbasis matching technique that directly aligns the eigenbasis and node features between real and synthetic graphs, preventing spectrum bias.~\citet{liu2023puma} introduced pseudo-label guided Memory Bank (PUMA), which extends distribution matching-based techniques to continual learning, using distribution matching to efficiently balance the retention of old knowledge while learning from new data.

\textit{Kernel Ridge Regression (KRR) Techniques}. The fourth taxonomy of graph condensation is Kernel Ridge Regression (KRR)~\citep{nguyendataset_KKR}. It offers an alternative to the complex bi-level optimization typical of neural network approaches by formulating the condensation as a regression problem with a closed-form solution. This can lead to more efficient optimization when the number of condensed samples is small, though it may trade off some capacity to capture non-linear relationships compared to deep learning methods. The optimization solution of KRR can be formulated as follows:

\begin{equation} 
    \begin{aligned} \mathcal{L}_{\text{cond}} = \frac{1}{2} \left| \mathbf{Y} - K_{\mathcal{T}\mathcal{S}} \left( K_{\mathcal{S}\mathcal{S}} + \epsilon I \right)^{-1} \mathbf{Y}' \right|^2, \end{aligned} \end{equation}
where $K_{\mathcal{T}\mathcal{S}}$ represents the kernel matrix of graph $\mathcal{T}$ and graph $\mathcal{S}$, and $K_{\mathcal{S}\mathcal{S}}$ is the kernel matrix of graph $\mathcal{S}$. In kernel ridge regression, inverting a matrix is required to solve the optimization problem, which can be very computationally expensive. However, in this technique, the matrix inversion only depends on the number of condensed samples, which is considered small, not the original large graph. Several representative methods for KRR, such as GC-SNTK (Graph Condensation with Structure-Based Neural Tangent Kernel)~\citep{wang2024fast_GCSNTK}, simplify and stabilize the optimization of condensed graphs by replacing the GNN training based on the bi-level optimization process with kernel ridge regression. This approach eliminates the need for multiple parameter initializations, making the condensation process more efficient and stable. KiDD (Kernel Ridge Regression-Based Graph Dataset Distillation)~\citep{xu2023kernel_KiDD} builds on this by reducing non-linearities in the kernel, simplifying the computation process and improving efficiency.

\begin{figure*}[!ht]
    \centering
    \includegraphics[width=\textwidth]{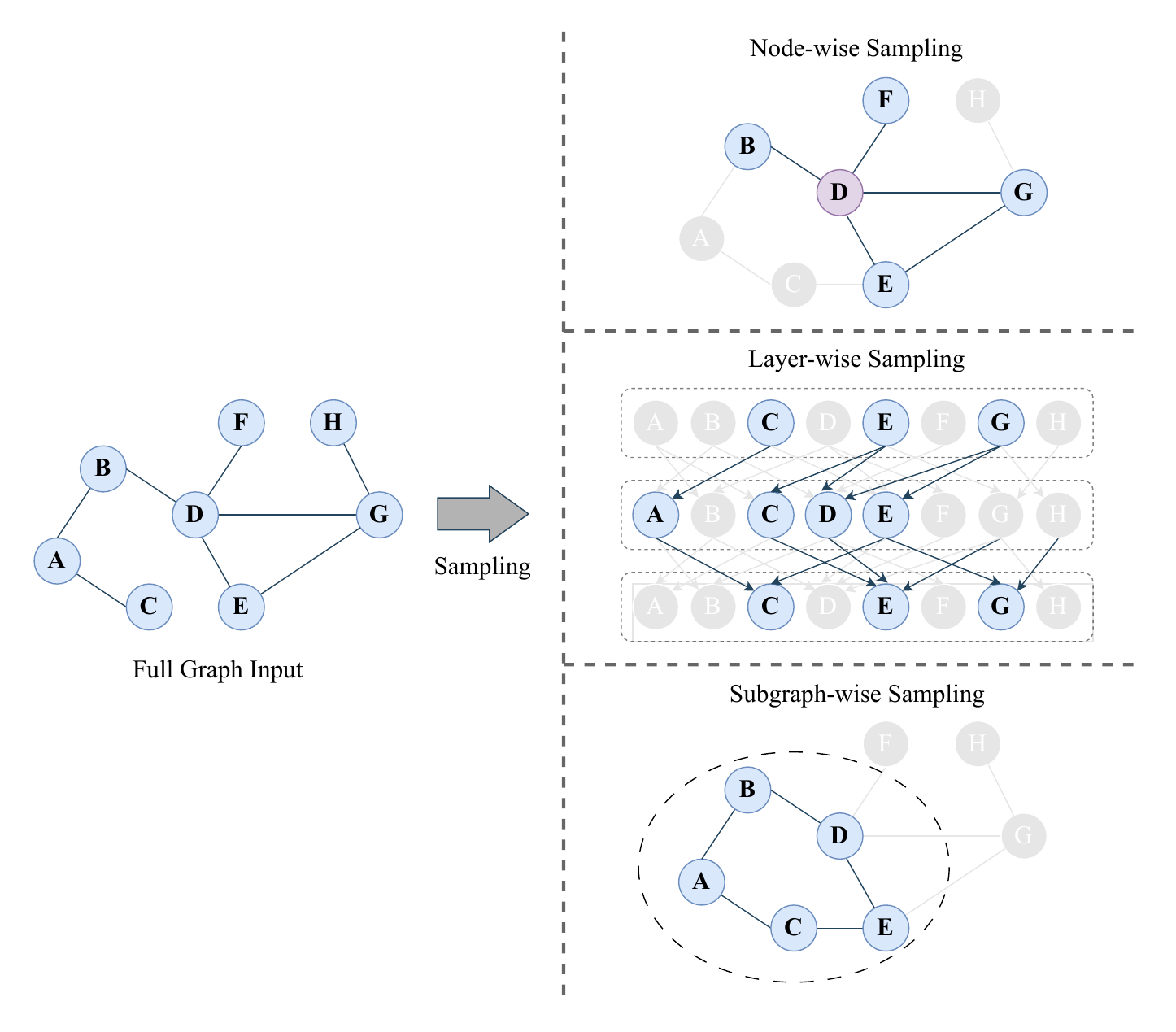}
    \caption{Computational Sampling Techniques.}
    \label{fig:sgl_sampling}
\end{figure*}

\section{Computational Sampling Methods}

Computational sampling methods aim to process graph data while reducing computational complexity and memory usage~\citep{ma2022graph_scalableGLSurvey}. In traditional graph learning techniques, such as random walks, matrix factorization, and other classical methods~\citep{goyal2018graph_classicalGraphLearningSurvey}, these approaches help represent or select subsets of graph properties during the training phase. This enables efficient management of large-scale graphs by reducing the need for excessive resource consumption. These methods ensure that only essential parts of the graph are processed, making the training process more computationally feasible.  In modern GNNs~\citep{xia2021graph_GLsurvey}, computational sampling also mitigates the cost of training and memory usage. Additionally, one of the methods addresses the \textit{neighbor explosion} problem~\citep{balin2023layer_neighborExplosion_labor}, where the number of nodes and edges grows exponentially with the number of GNN layers. 

In this section, we examine key computational sampling strategies used in both traditional graph learning and GNN-based approaches. These methods are categorized into three main types: node-wise sampling, layer-wise sampling, and subgraph-wise sampling. As illustrated in Figure~\ref{fig:sgl_sampling}, node-wise sampling selects neighbors for individual nodes, layer-wise sampling chooses nodes at each layer to manage computational complexity, and subgraph-wise sampling divides the graph into smaller subgraphs for independent processing during GNN training. 

\subsection{Node-wise Sampling}
In traditional graph learning methods, node-wise sampling is often applied in algorithms like random walks~\citep{xia2019random_randomWalk}, PageRank~\citep{brin1998pagerank}, DeepWalk~\citep{perozzi2014deepwalk}, and Node2Vec~\citep{grover2016node2vec}, which focus on selecting individual nodes and their direct neighborhoods in the process of graph traversal or exploration and to learn graph representations. In contrast, in GNNs, node-wise sampling specifically aims to reduce the computational load during the aggregation of neighbor features.

Random Walk traverses a graph by moving from one node to a randomly selected neighbor, capturing both local and global connectivity. PageRank builds on this concept by introducing a transition mechanism that allows the walker to jump to random nodes, assigning importance to nodes based on their visit probabilities and sampling them accordingly. DeepWalk refines this approach by performing truncated random walks, limiting the number of steps to balance computational efficiency while generating node sequences. These sequences are then treated like sentences in a language model and embedded into a vector space using the Skip-Gram model. Node2Vec further enhances DeepWalk by incorporating a biased random walk strategy that blends breadth-first search (BFS) and depth-first search (DFS). This hybrid approach enables the model to capture both local neighborhood structures and long-range dependencies, leading to more informative node representations. 

In the GNN context, a node-wise sampling strategy aims to aggregate information from the neighbors of selected nodes. This prevents computational overhead and reduces memory usage. GraphSAGE~\citep{hamilton2017inductive_reddit_graphSage} employs a sampling strategy to select a set of neighbors $N_s(v)$ for each node $v$ by aggregating their features to update the node's embedding in each GNN layer. The following is the updated node embedding at layer $k$:

\begin{equation} 
h_v^{(k)} = \sigma \left( W^{(k)} \cdot \text{AGGREGATE} \left( \{ h_v^{(k-1)} \} \cup \{ h_u^{(k-1)} \mid u \in N_s(v) \} \right) \right),
\end{equation}
where $h_v^{(k)}$ is the embedding of node $v$ at layer $k$, $\sigma$ is an activation function like ReLU, $W^{(k)}$ is the trainable matrix, and $\text{AGGREGATE}$ is a function to aggregate the features of sampled neighbor node embeddings $h_u^{(k-1)}$. Using this, GraphSAGE reduces the computational process by avoiding using all neighbors, which is useful for dealing with large and dense graphs. However, this node-wise sampling can lead to high variance in gradient estimates, making training unstable and potentially missing important information if the sampled neighbors are not representative. VR-GCN (Vectorized Relational Graph Convolutional Network) by~\citet{chen2017stochastic_VRGCN} uses node-wise sampling and applies a relational convolutional mechanism, which samples nodes and their relations for learning embeddings in multi-relational networks. Minimal Variance Sampling (MVS-GNN)~\citep{cong2020minimal_MVSGNN} introduces an adaptive node sampling based on their gradient information, minimizing both stochastic gradient and embedding approximation variance. This leads to faster convergence and more stable training for large-scale GNNs.

\subsection{Layer-wise Sampling}
Unlike node-wise sampling, layer-wise sampling methods focus on GNNs rather than traditional graph learning. GNNs typically work with a layered architecture across several network layers. In node-wise sampling for GNNs, the problem of the neighbor explosion is not fully addressed due to the exponential growth of the computational graph with the number of layers~\citep{balin2023layer_neighborExplosion_labor}. Suppose we have $K$ nodes sampled for each target node in each layer, the total number of nodes involved still grows exponentially as $K^L$, where $L$ is the number of layers. This makes it challenging to train deeper networks or those with a large number of $K$. Layer-wise sampling addresses this by selecting a fixed set of nodes at each layer independently, thereby controlling the overall computational budget. This approach addresses the neighborhood expansion problem by maintaining a consistent computational budget across layers, which improves the scalability and efficiency of training deep GNNs on larger graphs. 

Several techniques have been proposed to implement layer-wise sampling strategies to enhance the scalability of GNNs. FastGCN~\citep{chen2018fastgcn} introduces layer-wise importance sampling by approximating it using the \textit{Monte Carlo} method~\citep{gilmer2017neural}. It samples a fixed number of nodes in each layer based on the importance probabilities of the node features, simplifying the complexity of the GNN process. However, FastGCN can result in unstable model outputs due to high variance, as the nodes are sampled independently at each layer without considering their connections in other layers. ASGCN~\citep{huang2018adaptive_ASGCN} improves FastGCN by adaptively adjusting the sampling process based on the current model parameters to control high variance. As a result, the model focuses on the most relevant sampled nodes across layers, leading to more accurate and efficient learning on large-scale graphs.~\citet{zou2019layer_LADIES} propose a layer-dependent importance sampling (LADIES), which addresses the limitations of both by ensuring that the sampled nodes at each layer are connected to those in the previous layer. This concept preserves inter-layer dependencies and local graph structure while maintaining low variance, resulting in more stable training of deep GNNs on large-scale graphs. A recent technique, layer-neighbor sampling (LABOR) by~\citet{balin2023layer_neighborExplosion_labor}, combines neighbor and layer sampling, using Poisson sampling~\citep{ohlsson1998sequential_PoissonSampling} to significantly reduce the number of vertices and edges sampled while maintaining performance quality, making the training process much more efficient.

\subsection{Subgraph-wise Sampling}
Instead of processing the entire graph or sampling individual nodes or layers, subgraph sampling divides a large graph into smaller, manageable subgraphs to be processed independently. It works by selecting portions of the graph that represent the whole while maintaining the connectivity and structural properties within each subgraph. A traditional method that achieves a similar objective is spectral clustering~\citep{jin2020graph_spectralClustering}, which leverages eigenvalues and eigenvectors of the graph's Laplacian to detect community structures within the graph. Partitioning the graph into smaller groups called communities reduces computational complexity while preserving local graph properties. This enables algorithms to operate on these subgraphs independently.

In more sophisticated approaches like the GNNs, several techniques have been applied to enable subgraph-wise sampling to improve scalability. Cluster-GCN~\citep{chiang2019cluster_clusterGCN} uses graph partitioning algorithms to divide the graph into manageable clusters that serve as mini-batches during training. Cluster-GCN ensures that the nodes within each cluster are relevant and highly connected to preserve neighborhood information while also limiting interactions with remote nodes that are far away or in different parts of the graph. GraphSAINT~\citep{zeng2019graphsaint} is another technique that uses stochastic sampling methods like random walks, edge sampling, or node-based sampling to develop subgraphs with a good representation of the original. The recent technique called  SubInfer (subgraph training and inferring) by~\citet{zhou2024efficient_SubInfer} is designed specifically to address large heterogeneous graphs by breaking them into smaller subgraphs based on meta-paths, which are patterns that represent specific relationships between different types of nodes and the sub-graphs are completed with important nodes to fill in missing information and balance label distributions across subgraphs.

\subsection{Post-Propagation Decoupling}
The first strategy is the post-propagation decoupling approach, where the graph propagation process is performed only at a later stage. For instance, when using a multi-layer perceptron (MLP) for feature transformation, it can be expressed as:

\begin{equation}
\mathbf{H}^{(l+1)} = \sigma(\mathbf{H}^{(l)}\mathbf{W}^{(l)}), \quad l = 0, 1, \dots, L-2,
\end{equation}
where $\mathbf{H}^{(l)}$ is the node feature matrix at layer $l$, $\mathbf{W}^{(l)}$ is the learnable weight matrix, and $\sigma$ is the activation function. In post-propagation decoupling, the propagation or message-passing process is delayed until later in the network (often consolidated into the final layer) after the initial layers have focused on feature transformation. This separation can simplify the learning process and reduce computational overhead, even though some architectures may integrate propagation with intermediate feature transformations. This technique mathematically can be expressed as follows:

\begin{equation}
\mathbf{H}^{(L)} = \sigma(\mathbf{\Pi}\mathbf{H}^{(L-1)}\mathbf{W}^{(L-1)}),
\end{equation}
where $\mathbf{\Pi}$ is the precomputed graph propagation matrix, for example, can be derived from techniques such as Personalize PageRank (PPR)~\citep{yang2024efficient_personalizedPageRank}.

Several graph learning techniques have been proposed to implement the post-propagation strategy. Approximate personalized propagation of neural predictions (APPNP)~\citep{gasteiger2018predict_APPNP} introduces the PPR matrix as the feature propagation step. The computed graph propagation matrix in APPNP can be represented as follows:

\begin{equation} \hat{\mathbf{\Pi}} = \sum_{l=0}^{L_P} \alpha (1 - \alpha)^l \tilde{\mathbf{A}}^l, \end{equation}
where $\hat{\mathbf{\Pi}}$ represents the approximate PPR matrix computed over $L_P$ propagation hops. Each term in the sum applies the normalized adjacency matrix $\tilde{\mathbf{A}}$ raised to the $l$-th power and is weighted by $\alpha (1 - \alpha)^l$. In this formulation, the parameter $\alpha$ controls the balance between local and overall graph information. PPRGo introduced by~\citet{bojchevski2020scaling_PPRGo} implements the post-propagation technique by maintaining only the top-K entities per node when precomputing a sparse approximation of the PPR using the forward push algorithm.~\citet{liao2024scalable_SCARA} presente SCARA, introducing feature-oriented optimizations like \textit{Feature-Push} and \textit{Feature-Reuse}. This approach avoids redundant propagation by reusing previously computed features, ensuring that propagation is performed efficiently only after feature computations. As a result, it reduces both time complexity and memory overhead.  

\subsection{Pre-Propagation Decoupling}
The second strategy is a pre-propagation decoupling. The graph propagation phase is conducted before the learning phase by precomputing node representations based on the graph structure. This removes the need for repeated message-passing processes in multiple iterations. The precomputed node representation is denoted as the feature matrix $\mathbf{\Pi}$, which later serves as the input to the learning process, such as a neural network. This can be achieved by using the equation:

\begin{equation}
H^{(0)} = \mathbf{\Pi} = \sum_{l=0}^{L_P} a_l \tilde{A}^l \cdot X.
\end{equation}

Here, $\mathbf{\Pi}$ also serves as an input embedding of $H^{(0)}$, where this propagation step is processed through $L_P$ hops using the normalized adjacency matrix $\tilde{A}$. The coefficient $a_l$ controls the contribution of each hop, and $X$ is the original feature matrix. This propagation process is calculated only once before the downstream training process, reducing the computational cost during training. After precomputing the embeddings $\mathbf{\Pi}$, the learning process can be expressed as:

\begin{equation}
\mathbf{H}^{(l+1)} = \sigma\left( \mathbf{H}^{(l)} \mathbf{W}^{(l)} \right), \quad l = 0, 1, \dots, L - 1,
\end{equation}
where $\mathbf{H}^{(l)}$ represents the hidden layer at layer $l$ and $H^{(0)} = \mathbf{\Pi}$. $\mathbf{W}^{(l)}$ is a learnable weight matrix at layer $l$, and $\sigma$ is a non-linear activation function.

There are several representative techniques for this pre-propagation decoupling approach. The first is SGC by \citet{wu2019simplifying}, which aims to simplify the traditional GCN. This approach removes nonlinearities and collapses multiple GCN layers into a single linear transformation with the following propagation step: $\tilde{A}^{L_P}X$, where $\tilde{A}$ is the normalized adjacency matrix, $L_P$ denotes the number of propagation steps, and $X$ represents the node features. This propagation step is computed before the training phase, simplifying and speeding up the learning process. S2GC~\citep{zhu2021simple_S2GC} further enhances SGC by aggregating the results of multiple propagation steps, capturing both local and global information, thus providing a more comprehensive representation for each node. Graph Diffusion Convolution (GDC), developed by~\citet{gasteiger2019diffusion_GDC}, generalizes the pre-propagation concept by introducing weighted aggregation, where the coefficients $a_l$ control the contribution of each hop of propagation. These coefficients are derived from diffusion processes like the heat kernel or PPR. Another approach is GNN via Bidirectional Propagation (GBP) by~\citet{chen2020scalable_GBP}, which performs bidirectional propagation via a generalized PageRank mechanism, propagating from both node features and labeled nodes to create unbiased estimations of node representations. AGP (Approximate Graph Propagation)~\citep{wang2021approximate_AGP} introduces a randomized algorithm to approximate feature propagation across various proximity measures, such as heat kernel PageRank (HKPR) used in SGC~\citep{wu2019simplifying} or PPR employed in GBP~\citep{chen2020scalable_GBP}, enabling efficient precomputation before feeding into GNN models.

\section{Distributed Graph Learning}
Another strategy to improve the scalability of graph learning is distributed graph learning, which addresses the challenges of handling large-scale graphs that can not be processed on a single machine due to memory and computational constraints~\citep{besta2024parallel_distributedGraphLearningSurvey}. Distributing the graph and its training process across multiple machines enables scalable and efficient processing of large-scale graphs. This section reviews two main categories of distributed graph learning: whole-graph training and mini-batch training.

\subsection{Whole-graph Training}
Whole-graph training is a method where the entire graph is kept in memory across multiple computational machines. The distribution mechanism of whole-graph training partitions the graph across multiple machines. Each machine performs local forward and backward propagation on its graph partition. To capture the global structure, these machines must frequently synchronize their gradients. However, managing this communication overhead is challenging and requires careful strategies to balance computation and data exchange. This allows the model to update consistently across all partitions. To improve the quality of whole-graph training, two aspects need to be considered: strategies for managing communication overhead and strategies to ensure efficient partitioning to balance the computational load across machines.

Current techniques for distributed whole-graph training have been proposed. NeuGraph~\citep{ma2019neugraph} and ROC~\citep{jia2020improving_roc} architectures allow the system to split the graph and distribute it to different GPUs. However, NeuGraph does not dynamically assign tasks based on GPU load. Instead, it emphasizes partitioning the graph into chunks and uses a pipeline to overlap data transfer and computation between the CPU and GPU, with a topology-aware scheduling mechanism to balance workloads across multiple GPUs. On the other hand, ROC uses a dynamic approach for graph partitioning, where it balances tasks dynamically by monitoring memory usage and data flow to optimize GPU utilization. Both frameworks aim to speed up training by optimizing hardware better than static methods, where each GPU might be assigned a fixed portion of the graph.

Other techniques perform resource management in advance rather than being managed dynamically during training. These techniques involve a predetermined graph partition and allocating computational tasks to different machines or GPUs. Architectures like FlexGraph~\citep{wang2021flexgraph}, MG-GCN (Multi-GPU Graph Convolutional Network)~\citep{balin2022mg_mggcn}, and Dorylus~\citep{thorpe2021dorylus} employ the strategy of partitioning in advance to minimize communication overhead between machines. FlexGraph splits the graph at the start to balance the workload and minimize communications among different workers (machines). MG-GCN optimizes key mathematical operations and efficiently manages memory and data flow across GPUs. Dorylus breaks graphs and runs them on scalable cloud functions. Another framework, CAGNET (Communication-Avoiding Graph Neural Network)~\citep{tripathy2020reducing_CAGNET} employs a communication-avoiding approach through advanced partitioning strategies that reduce communication between GPUs, further enhancing scalability on distributed clusters. A more recent approach proposed by~\citet{wanpipegcn_PipeGCN}, called pipelined graph convolutional network (PipeGCN), focuses on pipelining inter-partition communication to avoid overheads during full-graph training.

\subsection{Mini-batch Training}
Unlike whole-graph training, mini-batch training samples smaller and manageable subgraphs from a larger graph and processes these in parallel across different machines. This allows for scalable and efficient training while reducing memory requirements. The key difference is that mini-batch training typically involves less communication by focusing on smaller subsets of the data. However, some communication is still necessary for tasks like synchronizing model parameters. Mini-batch training strategies are closely related to computational sampling methods~\citep{ma2022graph_scalableGLSurvey} that address scalability issues, specifically through subgraph-wise sampling strategies~\citep{chiang2019cluster_clusterGCN, zeng2019graphsaint}. The main difference is that distributed mini-batch training focuses on parallelizing and distributing the training process across multiple machines or GPUs. In contrast, graph-wise sampling focuses on creating efficient and representative samples of the larger graph to improve the quality and stability of training.

Several representative strategies for implementing distributed mini-batch training in graph learning systems exist. The first is AliGraph by~\citet{zhu12aligraph} which divides large-scale graphs across multiple CPUs and implements optimized sampling methods such as \textit{neighborhood}, \textit{traverse}, and \textit{negative} sampling along with caching strategies to minimize communication overheads between nodes to efficiently support large-scale, dynamic, and heterogeneous graphs. AGL~\citep{zhang2020agl_agl} accelerates training by constructing K-hop neighborhoods around target nodes, which serve as independent subgraphs for training, and applies a MapReduce-like pipeline~\citep{sun2023survey_mapreduce} to efficiently handle and model industrial graph data. DistDGL~\citep{zheng2020distdgl} extends the Deep Graph Library~\citep{wang2019deep_deepGraphLibrary} to support distributed mini-batch training on both CPU and GPU clusters. DistDGL employs graph partitioning with techniques to balance workloads and reduce network communication during training. DistDGLv2~\citep{zheng2022distributed_DistDGLv2} further enhances its first version by combining hybrid CPU and GPU training, introducing asynchronous mini-batch generation to ensure full hardware utilization across CPU, GPU, and network while maintaining model performance. A more recent framework called DGTP (Distributed GNN Training Planning) by~\citet{luo2022optimizing_DGTP} focuses on task placement and online scheduling for distributed GNN training, which optimizes data transmission and resource allocation. On top of that, it also minimizes communication bottlenecks and maximizes parallel task execution. 

%% file: chapters/3-tgl.tex
The primary graph learning techniques focus on static graphs where the structure and node features remain constant over time. General graph learning methods, such as matrix factorization~\citep{goyal2018graph_classicalGraphLearningSurvey}, are often used for node embedding and link prediction, but traditional matrix factorization techniques may struggle with evolving graph structures. It is also similar to random walk-based methods such as DeepWalk~\citep{perozzi2014deepwalk} and Node2Vec~\citep{grover2016node2vec}, which capture global and local connectivity within static graphs. GNNs, such as GAT~\citep{velickovic2018graph} and GCN~\citep{KipfW17gcn}, were also initially proposed to model static graphs. These approaches have demonstrated significant success in many real-world downstream tasks, such as node classification, link prediction, and community detection, by effectively mining information from static graphs.

Despite their effectiveness, typical graph learning methods face challenges when applied to dynamic or evolving graphs. Static models assume that the graph structure and node features do not change over time, which is often unrealistic in real-world scenarios where graph structures continuously evolve~\citep{barros2021survey_dynamicgraphsurvey, gravina2024deep_dynamicgraphSurveyBenchmark, febrinanto2023graph_GLL,zheng2025survey}.
Such phenomena are common across many domains, prompting researchers to propose a variety of datasets and benchmarks for further investigation.

Temporal graph learning aims to capture dynamic interactions and evolving features in graphs, which are important for many real-world tasks, such as real-time recommendation systems~\citep{sharma2024survey_graphRecommenderSystem}, human mobility~\citep{kong2022exploring_transportation,10992276}, epidemic modeling~\citep{yu2022pandora_graphEpidemic}, anomaly detection~\citep{RenEAGLE2023, RenGL4AA2023,shehzad2025dynamic}, and others~\citep{xia2021graph_GLsurvey}. By adapting to the temporal aspects of graph structures, graph learning models can perform more accurate predictions and provide more robust solutions for broader tasks. 

\begin{figure}[ht]
  \centering
  \includegraphics[width=1\linewidth]{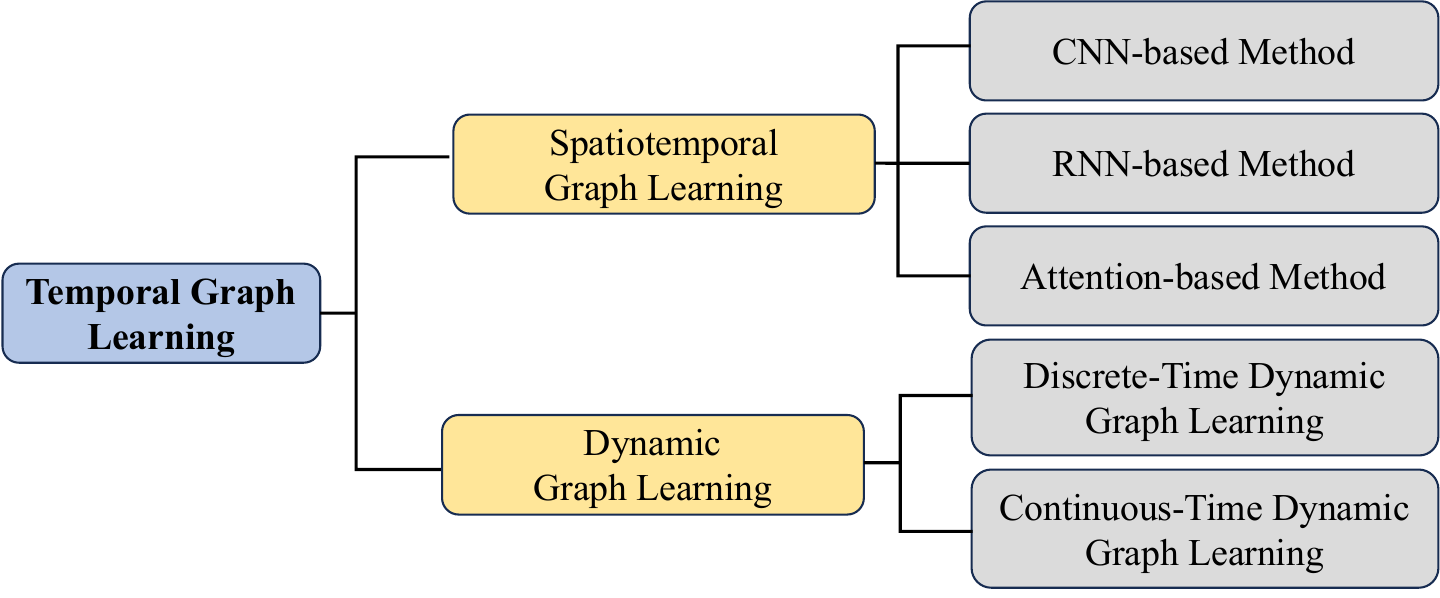}
  \caption{Strategy Taxonomy of Temporal Graph Learning.}
  \label{fig:tgl_categorization}
\end{figure}

In this survey, as shown in Figure~\ref{fig:tgl_categorization}, we review current approaches that address both the evolution of graph structure and the changes in node or edge features over time. These approaches are categorized into dynamic graph learning and spatiotemporal graph learning. Dynamic graph learning focuses on modeling structural changes, such as node or edge additions and deletions. Dynamic graph learning is further divided into two main tasks: discrete-time dynamic graph learning, which models structural changes at discrete time intervals, and continuous-time dynamic graph learning. In contrast, spatiotemporal graph learning incorporates temporal modeling techniques to capture both the spatial relationships within the graph structure and the temporal dynamics of feature changes in nodes or edges. We examine three core strategies that enhance temporal graph learning: convolutional neural networks (CNN-based methods), recurrent neural networks (RNN-based methods), and attention-based methods. This survey provides a detailed review of representative state-of-the-art techniques for temporal graph learning models.

\section{Dynamic Graph Learning}
Dynamic graph learning aims to model the evolving graph structure over time~\citep{barros2021survey_dynamicgraphsurvey, gravina2024deep_dynamicgraphSurveyBenchmark,shehzad2025dynamic}. Unlike static graphs, where the structure remains unchanged, dynamic graph learning captures changes in graph structure, such as node addition or removal, edge addition or removal, and changes in feature attributes associated with nodes or edges. Handling dynamic and evolving graph structures is important because it enables the detection of patterns and responses to real-time events in evolving graphs, which static models cannot capture. Dynamic graph learning often focuses on tasks like node classification and link prediction, driven by the availability of relevant datasets~\citep{gravina2024deep_dynamicgraphSurveyBenchmark}.

In this section, we categorize dynamic graph learning techniques into two groups: discrete-time and continuous-time dynamic graph learning focusing on \textit{discrete dynamic graph (DDG) types} and \textit{continuous dynamic graph (CDG) types}, respectively.

\subsection{Discrete-Time Dynamic Graph Learning} \textit{Discrete dynamic graph (DDG) types}. This type of graph represents a dynamic graph as a sequence of snapshots, which capture the state of the graph at specific time intervals~\citep{tang2024tcgc, zhu2023WinGNN, luo2022neighborhood_NAT}. In a discrete dynamic graph setting, the graph at time $t$ is represented as $G_{t} = (V_{t}, E_{t})$, where $V_{t}$ is the set of nodes and $E_{t}$ represents the set of edges at time $t$. Formally, DDG is represented as:

\begin{equation} \mathcal{G} = { G_0, G_1, \dots, G_T}, \end{equation}
where $\mathcal{G}$ is a dynamic graph and $T$ is the total number of snapshots. The time granularity to represent $t$ depends on the downstream tasks, such as hourly, daily, or monthly. The most straightforward way to model DDG is to apply a static graph learning model in each snapshot. In this case, each snapshot creates its own independent embedding space. However, this approach fails to capture the temporal evolution of the DDG, as embeddings at different time points remain disconnected. To address this limitation and avoid learning from scratch at every time step when changes occur, several works have proposed methods for modeling DDGs. In this survey, we refer to these methods as discrete-time dynamic graph learning. We review both advanced GNN-based techniques and non-GNN-based methods that build upon classical graph learning approaches.

\textit{GNNs-based strategy}. Unlike static models of GNNs such as GCN~\citep{KipfW17gcn} and GAT~\citep{velickovic2018graph}, several GNNs are proposed to model discrete dynamic graphs. Several representative techniques, such as EvolveGCN by~\citet{pareja2020evolvegcn}, address dynamic graphs by evolving the parameters of GCNs over time using recurrent neural networks, focusing on updating GCN weights to capture the graph's evolution. DySAT~\citep{sankar2020dysat} introduces a model that employs self-attention mechanisms along both structural and temporal dimensions to learn node representations, effectively capturing local structural properties and temporal changes in dynamic graphs. ROLAND~\citep{you2022roland} extends static GNNs to dynamic settings through a hierarchical node state update mechanism to enable a live-update evaluation to mimic real-world use cases.~\citet{qin2023seign} introduced SEIGN, which proposes a scalable evolving inception GNN for large dynamic graphs, utilizing a parameter-free message-passing mechanism and evolving graph filters to reduce memory usage and capture changing dynamics efficiently. WinGNN~\citep{zhu2023WinGNN} eliminates the need for temporal encoders by employing a random gradient aggregation window, capturing long-term temporal dependencies efficiently without additional encoders. TCGC~\citep{tang2024tcgc} introduces a framework for dynamic recommendation systems that incorporates a temporal collaboration-aware indicator and a co-evolving graph network, enhancing the model's ability to capture both short-term event dynamics and long-term historical influences on user preferences.

\textit{Non-GNNs-based strategy}. Discrete-Time Dynamic Graph Learning can also be processed using general deep learning or classical machine learning. The first example of a representative technique is NetWalk by~\citet{yu2018netwalk}, which employs a deep autoencoder with reservoir sampling to learn dynamic embeddings efficiently. They focus on anomaly detection tasks while maintaining memory efficiency in evolving networks. DynamicTriad introduced by~\citet{zhou2018dynamic_DynamicTriad}, models the triadic closure process to capture both structural and temporal evolutions by considering how to open triads (two nodes connected to a common third node) close over time. NAT (Neighborhood-Aware Temporal network)~\citep{luo2022neighborhood_NAT} uses dictionary-type neighborhood representations and an N-cache structure to efficiently update embeddings based on joint neighborhood information, enabling scalable and accurate modeling of large-scale DDGs. Dyngraph2vec~\citep{goyal2020dyngraph2vec} combines dense and recurrent layers to learn individual graph snapshots and the transition between them. This technique has three variations: autoencoder, recurrent, and a combination of autoencoder and recurrent.

\subsection{Continuous-Time Dynamic Graph Learning} \textit{Continuous dynamic graph (CDG) types}. This type of graph captures continuous changes in the graph over time, which can be denoted as $\mathcal{G} = (G, S)$, where the initial graph $G$ at time $t=0$ might be empty or have some initial structure. The initial graph $G$ at time $t=0$ does not include any past events. The changes in the graph are captured by a set of events $S = {S_1, S_2, \dots, S_T}$. Similar to DDG, static graph learning models can be adapted to model CDG. However, they fail to capture the sequence of events and consequently lose important temporal information. Additionally, static methods must be rerun from scratch to learn the new graph structure whenever a change occurs.

\textit{GNN-based strategy}. To effectively model CDG, several GNN models have been developed, each introducing unique mechanisms to capture temporal dynamics in the graph. JODIE~\citep{kumar2019predicting_JODIE} models user-item interactions using coupled RNNs that learn the trajectory of dynamic embeddings. It employs a projection operator to predict future interactions within the CDG setting, effectively capturing short-term and long-term user behaviors. Building upon attention mechanisms,~\citet{xu2020inductive_TGAT} propose the Temporal Graph Attention Network (TGAT), which uses a temporal graph attention mechanism with functional time encoding to generate node embeddings, handling unseen nodes and future interactions without the re-training process. TGN (Temporal Graph Networks)~\citep{rossi2020temporal_TGN} focuses on maintaining nodes' memory states to capture long-term dependencies, using message passing and temporal attention mechanisms for efficient prediction on continuous-time dynamic graphs. Graph Hawkes Transformer (GHT)~\citep{sun2022graph_GHT} processes CDG by leveraging a transformer-based model integrated with a neural temporal point process~\citep{hawkes1971point} to capture structural and temporal evolution for event prediction in temporal knowledge graphs. Another technique, TREND~\citep{wen2022trend}, integrates the event and node dynamics through a Hawkes process~\citep{hawkes1971point} based on GNN, capturing temporal characteristics and long-term interactions with inductive capabilities. Zebra~\citep{li2023zebra}, leveraging Temporal Personalized PageRank (T-PPR), addresses computational complexity in Temporal Graph Neural Networks (T-GNNs) by selectively aggregating important neighbors, improving efficiency and prediction accuracy in large-scale continuous dynamic graphs.~\citet{tian2024latent_conda} presented Conda, a latent diffusion-based data augmentation method for CDG that leverages a Variational Auto-Encoder (VAE) and conditional diffusion to enhance node embeddings and also improves performance in data-scarce environments.

\textit{Non-GNN-based strategy}. Several representative methods implementing general machine learning have also been proposed to model CDG. Know-Evolve~\citep{trivedi2017know_knowevolve} introduces a powerful method for reasoning over continuous dynamic knowledge graphs by modeling future interactions' occurrence and timing.~\citet{nguyen2018continuous_CDNE} introduced a continuous-time dynamic network embedding (CDNE) framework that leverages time-respecting traditional random walks to preserve temporal dependencies, effectively modeling DDGs in continuous time. DyRep (inductive deep representation learning), proposed by~\citet{trivedi2019dyrep}, utilizes multivariate point processes to model continuous dynamic graphs by distinguishing between communication (transient interactions) and association (long-term structural changes) events, which then update embeddings in real-time to reflect these changes. 

\begin{figure*}[ht!]
    \centering
    \includegraphics[width=\textwidth]{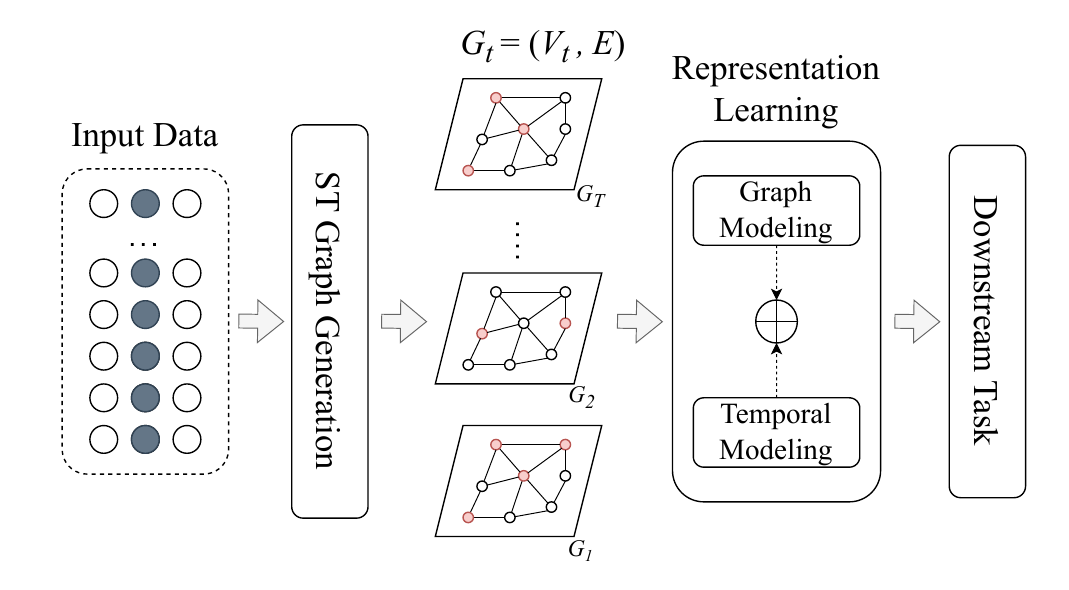}
    \caption{Spatiotemporal Graph Learning Setting.}
    \label{fig:spatio_temporal_graph_learing}
\end{figure*}

\section{Spatiotemporal Graph Learning} 
Another branch of temporal graph learning is spatiotemporal graph learning, which extends traditional graph learning to model dynamic graphs where node features evolve over time~\citep{jin2023spatio_stgraphSurvey, jin2024survey_stgraphSurvey}. This type of graph, known as a spatiotemporal graph (ST Graph), contains both the static spatial relationships between nodes and the dynamic temporal changes in node features. Spatiotemporal graph learning combines spatial modeling through graph learning with temporal analysis to capture how node features evolve over time, as illustrated in Figure~\ref{fig:spatio_temporal_graph_learing}. Several downstream tasks of spatiotemporal graph learning include traffic forecasting~\citep{shang2020discrete_gts}, disease spread prediction~\citep{yu2022pandora_graphEpidemic}, multivariate time series anomaly detection~\citep{deng2021graph_GDN}, and human activity recognition.

\textit{ST Graph}. Unlike static graphs, an ST graph contains temporal attributes of nodes that dynamically change over time. At the $t$-th time step, the ST graph can be formulated as $G_t = (V_t, E)$, where $V_t$ is a finite set of nodes corresponding to the observation at time $t$. Spatiotemporal graph learning aims to integrate graph learning techniques to capture the spatial structure of the graph while analyzing the temporal changes in node features.

\textit{ST Graph Generations}. The graph structure represents the spatial aspect of the ST graph. When relationships between nodes are not available, a graph generation process is required to establish connections between the entities. For example, determining relationships between the sensors that capture dynamic changes in each sensor's values. Several techniques in spatiotemporal graph learning propose solutions for graph generation. The most straightforward approach is to create a fully connected graph, though this method lacks generalization and suffers from high computational complexity up to $O(N^2)$. Some methods use knowledge-driven construction techniques, such as 2s-STGCN (Two-Stream Spatial-Temporal GCN)~\citep{bai2021two_biologicalAssumptionGraph}, which manually connects nodes in facial landmark points based on biological assumptions. HSTGCNN (Hierarchical spatiotemporal Graph Convolutional Neural Network)~\citep{zeng2021hierarchical_videoAD} builds graph structures based on human interactions for video anomaly detection. This type of knowledge-driven technique, which manually labels the relations, is suitable for small graphs.

Rather than manually constructing the graph, some methods aim to automate graph construction. A common approach is to start with random initialization and refine the graph structure through end-to-end learning during the training process. One way to achieve this is by using cosine similarity (or other distance metrics) between the feature vectors of each node~\citep{deng2021graph_GDN,wu2019graph_graphwavenet}. This technique employs a top-$K$ filter to maintain the sparsity of the resulting graph structure. Another approach leverages the Gumbel-Softmax trick~\citep{jang2017categorical_gumbel} to generate differentiable graphs by sampling edges from the Gumbel distribution~\citep{shang2020discrete_gts,kipf2018neural_nri}, as shown in:

\begin{equation}
    \label{eq:gumbel}
    A_{u,v} = \frac{\exp\left( \frac{ \log \widetilde{A}_{u,v} + \epsilon_v }{ \tau } \right)}{ \sum_{w \in \mathcal{N}_u} \exp\left( \frac{ \log \widetilde{A}_{u,v} + \epsilon_{w} }{ \tau } \right) }.
\end{equation}

Here, $A_{u,v}$ is the produced adjacency matrix for the relations between nodes $u$ and $v$, $\widetilde{A}_{u,v}$ is an intermediate adjacency matrix that encodes the unnormalized probabilities of the edges (or relations) between nodes $u$ and $v$, $\epsilon = -\log (-\log (s))$ is drawn from a Gumbel distribution with noise $s \sim $ Uniform(0,1), and $\iota$ is the temperature parameter.

In addition to the end-to-end learning process, another option for building the ST graph structure is to pre-calculate the probability of relations using causality discovery techniques~\citep{zhang2023spatial_causalityGraphStructure}. Once the graph structure is built, graph learning techniques, along with temporal modeling techniques, can model the ST graph effectively.

Overall, in this section, we review several representative methods for spatiotemporal graph learning. These methods are categorized based on how they model the temporal evolution of node features. We group them into three main categories: CNN-based methods, RNN-based methods, and Attention-based methods.

\subsection{CNN-based Methods}
\textit{Spectral-based graph learning}. Methods that belong to this strategy use the graph Laplacian and its eigenvalues to transform the original graph structure, enabling convolution operations similar to those used in classical signal processing tasks~\citep{jin2020graph_spectralClustering}. Spatiotemporal Graph Convolutional Network (STGCN) by~\citet{yu2018spatio_STGCN} captures spatial relationships across the graph using Chebyshev polynomial approximations and combines this with temporal gated convolutions to model time-series data. Spectral Temporal Graph Neural Network (StemGNN) by~\citet{cao2020spectral_STEMGNN} extends this approach by fully modeling both spatial and temporal dependencies in the spectral domain, using the Graph Fourier Transform (GFT) for spatial correlations and the Discrete Fourier Transform (DFT) for temporal patterns. On top of that, StemGNN also learns the temporal patterns of dynamic features in the graph via 1D convolutions and GLU (Gated Linear Unit) layers. 

\textit{Spatial-based graph learning}. Methods included in this strategy take a different approach by modeling spatial dependencies through direct aggregation of information from neighboring nodes in the graph. These models focus on the local neighborhood structures in graphs and dynamically learn spatial relationships without relying on spectral transformations. For instance,~\citet{wu2019graph_graphwavenet} developed GraphWaveNet, which introduces a self-adaptive adjacency matrix to directly learn spatial dependencies from data and combines this with dilated causal convolutions to capture long-term temporal features. To extend GraphWaveNet, ~\citet{wu2020connecting_MTGNN} introduced MTGNN, which also learns the optimal graph structure through end-to-end learning and uses spatial and temporal convolutions to handle multivariate time-series forecasting. Another work, STG-NCDE~\citep{choi2022graph_STGNCDE}, integrates spatial graph convolutions with neural-controlled differential equations to model temporal dynamics in continuous time. Similarly, st-GNN~\citep{karimi2021spatiotemporal_STGNN} constructs graphs based on spatial proximity and employs spatial graph convolutions and temporal convolutional layers to capture spatiotemporal dynamics in the graph structure. CST-GL~\citep{zheng2023correlation_CSTGL} learns dynamic adjacency matrices to model spatial correlations and uses spatial-temporal graph neural networks to capture spatial and temporal features through convolution operations.

\subsection{RNN-based Methods}
\textit{Spectral-based graph learning}. This part reviews several representative methods that use spectral concepts in RNN-based spatiotemporal graph learning. The first method is ST-MGCN (spatiotemporal Multi-Graph Convolution Network)~\citep{geng2019spatiotemporal_STMGCN}. In this method, multiple graph convolutions based on Chebyshev polynomials of the graph Laplacian are used to capture various types of spatial relationships, including neighborhood proximity, functional similarity, and transportation connectivity. Additionally, it employs a contextual gated recurrent neural network (CGRNN) to model temporal dependencies on global contextual information. A work by~\citet{simeunovic2021spatio_gclstm_gctrafo} proposes two variants of spatiotemporal graph neural networks, one of which is Graph-Convolutional Long Short Term Memory (GCLSTM). GCLSTM applies spectral graph convolution, which operates in the graph Fourier domain and uses long short-term memory (LSTM) units to capture temporal dependencies in the time series data.~\citet{deng2022graph_STGAN} proposed STGAN, which integrates traditional graph convolution based on spectral graph theory within their Graph Convolutional Gated Recurrent Units (GCGRUs) to jointly model spatiotemporal patterns for anomaly detection. RGSL (Regularized Graph Structure Learning)~\citep{yu2022regularized_RGSL} combines graph convolutions with GRUs, leveraging both prior information and dynamically learned graph structures. It further enhances forecasting task performance through the Laplacian Matrix Mix-up Module (LM3), which fuses explicit relationships from domain knowledge with implicit graphs generated by the Regularized Graph Generation (RGG) module.

\textit{Spatial-based graph learning}. Several advanced models have been proposed to integrate spatial graph learning with recurrent neural networks (RNNs) for spatiotemporal graph learning. For example, Diffusion Convolutional Recurrent Neural Network (DCRNN)~\citep{li2018diffusion_dcrnn} captures spatial dependencies in traffic networks using diffusion convolution and models temporal dynamics through gated recurrent units (GRUs). Similarly,~\citet{chen2019gated_MResRGNN} proposed Gated Residual Recurrent Graph Neural Network (MRes-RGNN) combines graph convolutions, residual connections, and recurrent units to improve the modeling of spatiotemporal patterns in traffic data. Adaptive Graph Convolutional Recurrent Network (AGCRN) by~\citet{bai2020adaptive_AGCRN} dynamically learns both node-specific parameters and an adaptive graph structure, which are integrated into a recurrent framework for time series forecasting. Graph for Time Series (GTS) proposed by~\citet{shang2020discrete_gts} learns the graph structure using the Gumbel-Softmax trick~\citep{jang2017categorical_gumbel} and employs graph convolutions with RNNs for multivariate time series forecasting.~\citet{cini2023scalable_sgp} deployed Scalable spatiotemporal GNNs that use echo-state networks and graph convolutions for efficient spatiotemporal embeddings for large-scale forecasting tasks. Hierarchical Spatiotemporal Graph Neural Network (HST-GNN) by~\citet{kan2022sign_HSTGNN} models complex sign language translation using a hierarchical graph structure and RNNs to capture both spatial and temporal dynamics. Lastly,~\citet{he2020spatiotemporal_topomad} proposed TopoMAD, which combines graph neural networks with LSTMs in a variational autoencoder framework for unsupervised anomaly detection in cloud systems.

\subsection{Attention-based Methods}
\textit{Spectral-based graph learning}. Unlike CNN-based and RNN-based approaches, only a few studies have explored spectral-based graph learning for analyzing spatial relationships in spatiotemporal attention models. One such method is GCTrafo (Graph-Convolutional Transformer) by ~\citet{simeunovic2021spatio_gclstm_gctrafo}, which utilizes spectral graph convolution in the graph Fourier domain. It combines this with Transformer-based attention mechanisms to capture temporal dependencies across different time steps for each node. Another spectral-based approach is TPGNN (Temporal Polynomial Graph Neural Network) by ~\citet{liu2022multivariate_TPGNN}, which represents temporal dependencies as matrix polynomials in the spectral domain. By employing spectral graph convolutions, TPGNN effectively models time-varying relationships and captures dynamic spatial correlations in multivariate time-series data.

\textit{Spatial-based graph learning}. Several methods in the attention-based category lie in spatial graph learning.~\citet{li2020forecaster} proposed a forecaster that learns spatial dependencies using Gaussian Markov Random Fields and incorporates a sparsified Transformer with attention mechanisms to capture long-range temporal dependencies. MTAD-GAT by~\citet{zhao2020multivariate_MTADGAT} utilizes two parallel layers of GAT, feature-oriented GAT and time-oriented GAT, to model dependencies between features and across time steps using an attention mechanism for anomaly detection in multivariate time series.~\citet{zheng2020gman} introduced GMAN, which employs spatial and temporal attention mechanisms within a graph framework to model dynamic spatial correlations and nonlinear temporal relationships for traffic prediction. Another work, SPIN (spatiotemporal Point Inference Network)~\citep{marisca2022learning_SPIN} uses a sparse spatiotemporal attention mechanism combining temporal self-attention and spatiotemporal cross-attention to reconstruct missing data in spatiotemporal graphs. Work by~\citet{shao2022pre_STEP} enhances the spatial-temporal graph neural network model by pre-training a Transformer-based model. The pre-trained model captures long-term temporal patterns and refines graph structures using attention-based temporal representations. RAINDROP is a spatiotemporal graph learning method proposed by~\citet{zhanggraph_raindrop}, which handles irregularly sampled multivariate time series by learning dynamic graphs and considers the importance of sequential sensor observations via temporal attention to model time-varying dependencies. A more recent approach, TransMOT~\citep{chu2023transmot}, introduces a Spatial-Temporal Graph Transformer that uses attention mechanisms within a graph learning framework to model spatial and temporal relationships for multiple object tracking. 

%% file: chapters/4-mgl.tex

Data in real-world scenarios often come from different sources. 
Multimodal fusion aims to integrate information from diverse data sources, such as visual, textual, and acoustic data. However, multimodal data often exhibits intricate relational dependencies, including both inter-modal and intra-modal correlations. Traditional multimodal learning methods that rely on basic modality alignment and simple fusion techniques often fail to effectively capture and integrate the rich, complementary information across different modalities. To address this challenge, multimodal graph learning has emerged as a powerful approach~\citep{ektefaie2023multimodal,peng2024learning,xie2024multimodal,he2025unigraph2}. It leverages graph structures to model and learn the intricate inter-modal and intra-modal correlations, enabling a more nuanced integration of multimodal data. 

Compared to other types of multimodal learning, where data typically follows a consistent and well-defined structure, multimodal graph learning deals with the challenge of processing and fusing data with complex and heterogeneous graph topologies. This complexity is further amplified when the graph structure varies between different modalities. As a result, numerous studies have been proposed to address these challenges in multimodal graph learning, targeting a wide range of tasks such as recommendation, medical diagnosis, and knowledge graph completion~\citep{kim2023heterogeneous,zhao2024deep,ai2024gcn,peng2024mbgt}. According to the structural characteristics of multimodal data, we classify multimodal graph learning into two categories: graph-driven multimodal learning and learning on multimodal graphs. Section~\ref{mgl1} and Section~\ref{mgl2} will explore these approaches in detail.

\section{Graph-driven Multimodal Learning }
\label{mgl1}

In graph-driven multimodal learning, data from each modality is represented as unimodal graphs (e.g., textual graphs, visual graphs). Various graph learning models are applied to process these graphs, facilitating multimodal fusion. Based on the fusion strategy, graph-driven multimodal learning can be categorized into two types: separate unimodal learning and joint multimodal learning. Table~\ref{MGtabel1} lists recent representative graph-driven multimodal learning methods.

\begin{table}
	\centering
	\renewcommand\arraystretch{1}
	\small
	\caption{Repretentative methods for graph-driven multimodal learning.}
	\begin{tabular}{|>{\centering\arraybackslash}p{2cm}|>{\centering\arraybackslash}p{2.8cm}|>{\centering\arraybackslash}p{3cm}|>{\centering\arraybackslash}p{3cm}|}
		\hline
		\textbf{Type} & \textbf{Method}& \textbf{Model}& \textbf{Description} \\ \hline
		\multirow{6}{*}{\makecell{Separate \\ Unimodal \\ Learning}} 
		& \citet{tao2020mgat} & Gated Attention-Graph Neural Network &  Textual and Visual User-Item Graphs \\ \cline{2-4} 
		& \citet{chen2022hybrid}  & Graph Transformer& Textual and Visual Knowledge Graphs\\ \cline{2-4} 
		& \citet{lin2022modeling} & Graph Contrastive Learning& Hierarchical Textual, Visual, Acoustic Graphs \\ \cline{2-4} 
		& \citet{liu2023multimodal}  & Graph Contrastive Learning & Textual and Visual User-Item Graphs \\ \cline{2-4} 
		&\citet{guo2024lgmrec} &Hypergraph Embedding&Textual and Visual User-Item Graphs  \\ \cline{2-4} 
		&\citet{cai2024multimodal}&Out-of-distribution Generalized Multimodal Graph Neural Network &Textual and Visual Graphs  \\ \cline{1-4}

		\multirow{7}{*}{\makecell{Joint \\ Multimodal \\ Learning}} 
		& \citet{saqur2020multimodal}  & Shared Graph Neural Network & Textual and Visual Graphs \\ \cline{2-4} 
		& \citet{wang2023hypergraph} & Multimodal Phenotypic Graph Diffusion  & Multimodal Brain Graphs \\ \cline{2-4} 
		&\citet{liu2024multimodal} & Spatio-Temporal Graph Convolution &Inter-industry, Inter-company, Inter-day Graphs  \\ \cline{2-4} 
		& \citet{kim2020hypergraph}& Hypergraph Attention Networks  & Textual and Visual Symbolic Graphs \\ \cline{2-4}
		& \citet{cai2022graph}  & Graph Transformer & Structural and Functional Brain Graphs \\ \cline{2-4} 
		& \citet{he2023multimodal}  & Graph Transformer& Textual, Semantic, Dense Region Graphs \\ \cline{2-4} 
		&\citet{liu2024muse} &Cross-Graph Transformer &Gene Graphs  \\ \cline{1-4} 

	\end{tabular}
	\label{MGtabel1}
\end{table}

\subsection{Separate Unimodal Learning }

\begin{figure}[h]
	\centering
	\includegraphics[width=\textwidth]{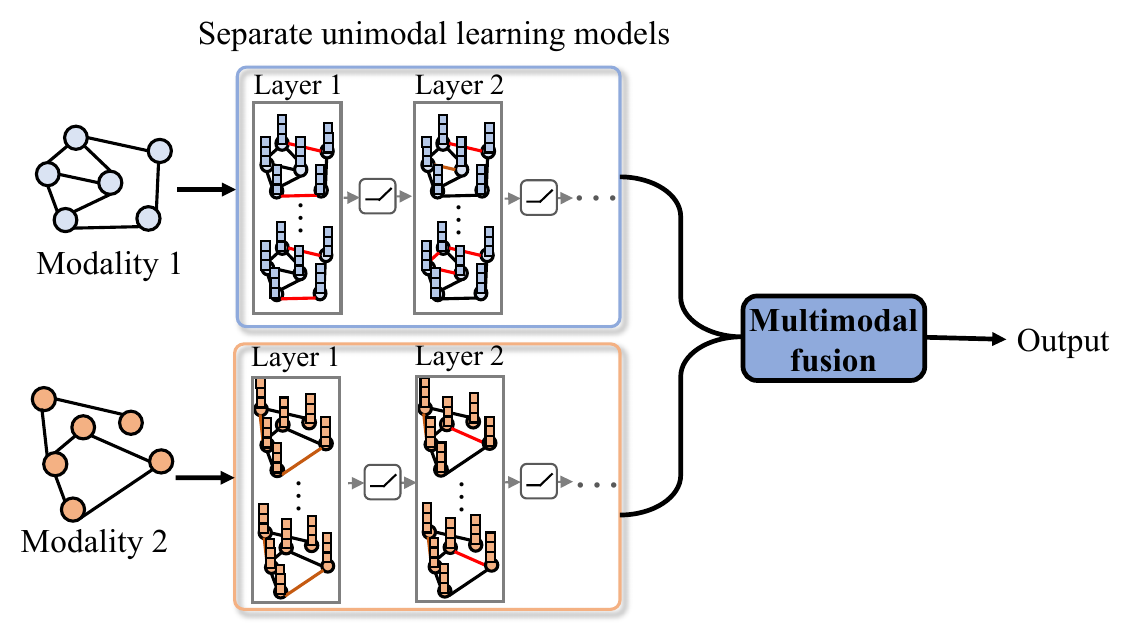}
	\caption{The illustration of separate unimodal learning.}
	\label{graphdriven1}
\end{figure}

Separate unimodal learning involves the use of distinct graph learning architectures, each tailored to a specific data modality. The outputs from these architectures are then aggregated to achieve multimodal fusion. In other words, separate unimodal learning first handles each modality independently and then performs the fusion of the modalities (see Figure~\ref{graphdriven1}). Given $m$ graphs, each corresponding to a modality, the graph for modality $i$ is denoted as $\mathcal{G}_i$. The process of separate unimodal learning is as:
\begin{equation}
	\centering
	\label{eq1}
	\mathbf{\hat{Z}} = Fusion (\mathbf{Z}_1, \mathbf{Z}_2, \dots, \mathbf{Z}_m).
\end{equation}
Here, $\mathbf{Z}_i$ indicates the representation of $\mathcal{G}_i$ learned by the unimodal graph learning architecture, and $\mathbf{\hat{Z}} $ is the multimodal representation. $Fusion (\cdot)$ stands as the fusion operation. Various graph learning models, such as GCNs, GATs, GAEs, and others, can be applied to each unimodal learning architecture. These models enable effective learning from the structured data within each modality before the multimodal fusion process.

\textbf{Modality-specific Representation.}~Separate unimodal learning exhibits the advantage of capturing modality-specific representation.
Specifically, the modality-specific representation $\mathbf{Z}_i$ for the modality $i$ can be effectively extracted using various independent graph learning models. For example, MGAT~\citep{tao2020mgat} applies embedding propagation layers to perform message passing on each single-modal user item graph, capturing user preferences on each modality. \citet{guo2024lgmrec} proposed local and global graph learning-guided multimodal recommender (LGMRec) for learning on multimodal user-item graphs. It introduces a local graph embedding module to perform message propagation on textual and visual user-item graphs separately, capturing intra-modal local user interests and generating modality-specific representations. MKGformer~\citep{chen2022hybrid} is a hybrid Transformer model designed for the multimodal knowledge graph completion task. In MKGformer, two types of knowledge graph are constructed: visual knowledge graphs and textual knowledge graphs. A visual Transformer and a textual Transformer, both equipped with self-attention modules, are then applied to these knowledge graphs to learn modality-specific representations. 

\textbf{Multimodal Fusion.}~After obtaining modality-specific representations of all modalities, multimodal fusion operation $Fusion (\cdot)$ is applied to integrate them. 
\begin{itemize}
    \item Typical fusion methods: Three typical fusion operations, including concatenation, summation, and element-wise multiplication, are widely adopted for multimodal fusion. Concatenation concatenates the representations of each modality along a specified dimension, creating a unified representation that preserves all the features of the individual modality. It is denoted as:
    $\mathbf{\hat{Z}} = \mathbf{Z}_1\oplus\mathbf{Z}_2\oplus, \dots, \oplus\mathbf{Z}_m$. Summation aggregates the multimodal representations by summing them element-wise, denoted as: $\mathbf{\hat{Z}} = \sum_{i=1}^m \mathbf{Z}_i$. Element-wise multiplication captures interactions between modalities at each feature dimension, denoted as: $\mathbf{\hat{Z}} = \prod_{i=1}^m \mathbf{Z}_i$.    
    \item Global learning module: Many studies apply a global learning module to fuse multimodal representations generated by separate learning models, aiming to enhance overall feature integration. For example, LGMRec~\citep{guo2024lgmrec} designs a global hypergraph embedding module to model global cross-modality dependency relations of modality-specific representations produced by multiple local graph embedding modules.
    OMG-NAS~\citep{cai2024multimodal} maintains global weights across learning architectures for all modalities and fuses multimodal information.
    \item Attention-based fusion: Some studies apply attention mechanisms, such as cross-attention, on multimodal representations, capturing cross-modality interactions. For instance, MGAT~\citep{tao2020mgat} leverages gated attention to aggregate information across multiple modalities.     
    \item Contrastive learning-based fusion: Some works present multimodal graph contrastive learning frameworks. These frameworks apply a contrastive objective strategy to capture inter-modal and intra-modal similarities and differences, enabling multimodal fusion. For example, \citet{liu2023multimodal} proposed MGCL-RA,  which presents a contrastive loss of cross-modality to capture similarities between visual and textual modalities. \citet{lin2022modeling} presented a hierarchical graph contrastive learning (HGraph-CL) framework for multimodal sentiment analysis. It explores sentiment relations across different modalities, employing a graph contrastive learning strategy at the inter-modal level.
\end{itemize}

Due to independent learning in each modality, the unimodal representations obtained from unimodal architectures are highly modality specific, preserving the unique characteristics of each modality. Therefore, separate unimodal learning can fully encode intra-modal correlations and effectively protect modality-specific information during the fusion process.

\subsection{Joint Multimodal Learning}

\begin{figure}[h]
	\centering
	\includegraphics[height=4.8cm]{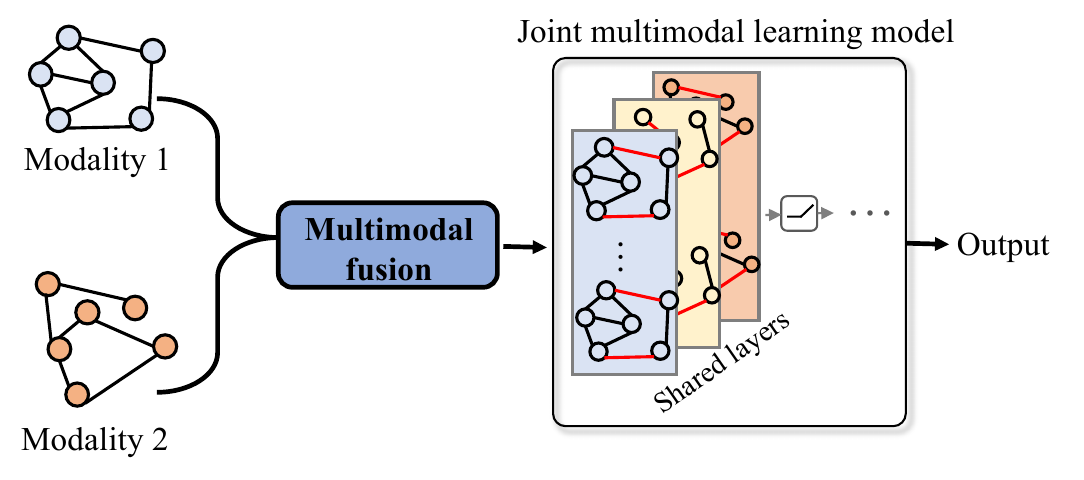}
	\caption{The illustration of joint unimodal learning.}
	\label{graphdriven2}
\end{figure}

Joint multimodal learning first exploits a multimodal fusion step to integrate various graphs. Then, an all-in-one multimodal architecture, which includes shared representation layers and processes the fused multimodal information (see Figure~\ref{graphdriven2}). Unlike separate unimodal learning, joint multimodal learning first performs graph fusion and then conducts the learning process. Formally, the process of joint multimodal learning can be expressed as:
\begin{equation}
	\centering
	\label{eq2}
	\begin{aligned}
		\mathcal{\hat{G}} &= Fusion(\mathcal{G}_1, \mathcal{G}_2, \dots, \mathcal{G}_m), \\
		\mathbf{\hat{Z}} &= Joint(\mathcal{\hat{G}}).
	\end{aligned}
\end{equation}
Here, $\mathcal{\hat{G}}$ is the fused multimodal graph after the fusion operation. $Joint (\cdot)$ stands as joint learning process. Many researchers have focused on joint multimodal learning, particularly in designing shared representation layers to optimize multimodal representations.

\textbf{Inter-modal Correlation Modeling.}~By integrating multimodal graphs into a shared representation space, the joint learning process $Join(\cdot)$ can effectively model inter-modal correlations. \citet{saqur2020multimodal} proposed a multimodal graph network (MGN) architecture that includes a graph parser and a graph matcher. In the graph parser, they employed a two-layer graph isomorphism network to learn from two multimodal graphs: textual graphs and visual graphs. Subsequently, in the graph matcher, a shared graph neural network is utilized to learn from the multimodal representations. \citet{wang2023hypergraph} designed a unified multimodal phenotypic graph diffusion method to process multimodal brain graphs, capturing inter-modal correlations between different brain graphs (e.g., brain structural graphs and brain functional graphs). 

Some works focus on dynamic graphs. For example, Melody-GCN~\citep{liu2024multimodal} is a multimodal dynamic graph convolutional network designed for stock price prediction tasks. This model specifically addresses three types of graphs, including inter-industry graphs, inter-company graphs, and inter-day graphs. Melody-GCN introduces a multimodal fusing-diffusing block to effectively fuse multimodal representations, enhancing the model's ability to capture temporal dependencies and interactions.

With a strong relation modeling ability, such as cross-modality relation modeling, the attention mechanism is particularly representative for joint multimodal learning. \citet{kim2020hypergraph} proposed a hypergraph attention network for integrating symbolic information extracted from multimodal symbolic graphs, specifically textual and visual graphs. They developed a semantic similarity measurement-based co-attention mapping function to facilitate the fusion of multimodal information. Graph transformers have gained significant attention because of their ability to effectively capture complex dependencies within graphs. Consequently, many studies have applied multimodal graph transformers to achieve joint multimodal learning. For example, GTGL~\citep{cai2022graph} is a graph transformer geometric learning framework to learn on multimodal brain graphs. GTGL implements a graph transformer network with the multi-head self-attention mechanism to jointly learn structural and functional brain graphs. \citet{he2023multimodal} developed a multimodal graph transformer that incorporates a graph-involved, plug-and-play quasi-attention mechanism for multimodal fusion. Their model operates on three types of graphs, including textual graphs, semantic graphs, and dense region graphs. Specifically, they set a cross-modal mask as an all-zero matrix to learn cross-attention between multimodal features. Moreover, MuSe-GNN~\citep{liu2024muse}, which features a cross-graph Transformer architecture, is proposed for multimodal biological data integration from gene graphs. In the cross-graph Transformer, a graph neural network is integrated with a multi-head self-attention module, incorporating co-expression information to obtain multimodal gene representations. This design enhances the ability to capture relationships across different gene modalities and improves the quality of multimodal biological data fusion.

Different from separate unimodal learning, joint multimodal learning models have a unified learning framework for all modalities. Shared layers can simultaneously process graphs from multiple modalities and obtain multimodal representations. Therefore, joint multimodal learning models are relatively lightweight, with low computational complexity. Furthermore, by processing multimodal graphs in a unified manner, joint multimodal learning has the advantage of capturing inter-modal correlations more effectively.

\section{Learning on Multimodal Graphs}
\label{mgl2}

\begin{figure}[h]
	\centering
	\includegraphics[height=4.5cm]{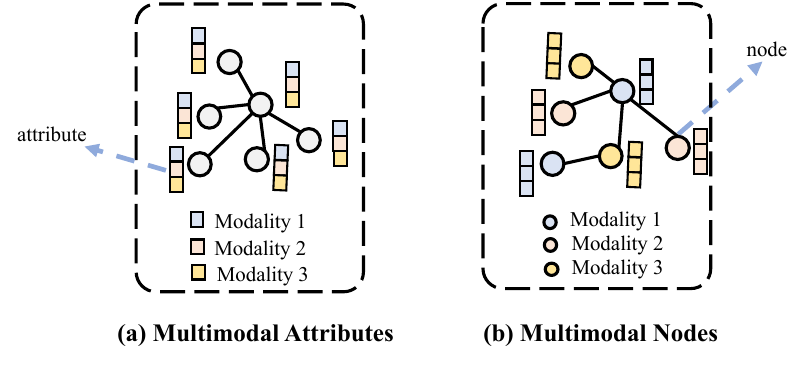}
	\caption{Two types of multimodal graphs.}
		\label{multimodal graph}
\end{figure}

Learning on multimodal graphs refers to applying graph learning models to multimodal graphs, where multimodal information from different sources is integrated into a unified graph structure. This approach allows for the simultaneous processing of inter-modal and intra-modal relations within a single graph representation. According to the integration manner of multimodal information, multimodal graphs can be divided into two types: graphs with multimodal attributes and graphs with multimodal nodes. Figure~\ref{multimodal graph} illustrates examples of these two multimodal graph types, and Table~\ref{MGtabel2} lists recent representative methods for each type.

\begin{table}
	\centering
	\caption{Representative methods for learning on multimodal graphs.}
	\renewcommand\arraystretch{0.1}
	\setlength{\tabcolsep}{0.4mm}{
		\small
		\begin{tabular}{|>{\centering\arraybackslash}p{2cm}|>{\centering\arraybackslash}p{3cm}|>{\centering\arraybackslash}p{3.5cm}|>{\centering\arraybackslash}p{3.5cm}|} \hline 
			\textbf{Type}&\textbf{Method}&\textbf{Model}&\textbf{Description}\\ \hline
			\multirow{10}{*}{\makecell{Learning on \\ Multimodal \\ Attributes}} 
			&\citet{lian2023gcnet}&Graph Complete Network &Multimodal Conversational Interactions\\ \cline{2-4} 
			&\citet{9810283}&Graph Convolutional Network&Multimodal Brain Graphs\\\cline{2-4} 
			&\citet{TianZGMMC22}& Heterogeneous Graph Neural Network&Heterogeneous Recipe Graphs\\ \cline{2-4} 
			&\citet{000100LDC23}&Graph Transformer&Multimodal Knowledge Graphs\\ \cline{2-4} 
			&\citet{chen2022multi} &Graph Contrastive Learning&Multimodal Knowledge Graphs\\ \cline{2-4} 			
			&\citet{li2023joyful}&Graph Contrastive Learning&Multimodal Conversational Interactions\\ \cline{2-4} 
			&\citet{cao2022otkge}& Relational Transformation&Textual and Visual Knowledge Graphs\\ \cline{2-4} 
			&\citet{meng2024masked} & Graph Convolutional Network& Textual, Visual, Acoustic Graphs \\ \cline{2-4} 
			&\citet{zhou2024disentangled} &Graph Variational Auto-Encoder&Textual and Visual Item-Item Graphs \\ \cline{2-4} 
			&\citet{li2024multimodal} &Spectrum-based Graph Learning&Textual, Visual, Acoustic Graphs\\ \cline{1-4} 
			
			\multirow{9}{*}{\makecell{Learning on \\ Multimodal \\ Nodes}} 
			& \citet{wang2020multimodal}& Graph Convolutional Network&Multimodal Knowledge Graphs\\		\cline{2-4} 		
			&\citet{wen2022graph}&Cell-feature Graph Convolution &Cell-feature Graphs\\ \cline{2-4} 
			&\citet{hu2023graph}&Graph Fusion Neural Network & Freezing of Gait Graphs\\ \cline{2-4} 
			&\citet{zeng2023multi}&Hypergraph Convolutional Network  &Multimodal Knowledge Hypergraphs\\ \cline{2-4} 
			&\citet{jin2018learning}& Junction Tree Encoder-Decoder&Molecular Graphs\\ \cline{2-4} 
			&\citet{lyu2021mdnn}&Graph Neural Network &Multimodal Drug Knowledge Graphs\\ \cline{2-4} 
			&\citet{song2024scene} &Graph Convolutional Network&Multimodal Knowledge Graphs\\ \cline{2-4} 
			&\citet{kim2024self}& Graph Convolutional Network &Textual and Visual User-Item Graphs \\ \cline{2-4} 
			&\citet{tu2024adaptive}&Adaptive Interactive Graph Network & Multimodal Conversational Interactions \\ \cline{1-4} 
			
	\end{tabular}}
	\label{MGtabel2}
\end{table}

\subsection{Learning on Multimodal Attributes}

In graphs with multimodal attributes, the features of each node consist of multiple modalities (see Figure~\ref{multimodal graph} (a)). Notably, these graphs can only be constructed when data from each modality shares the same underlying graph structure. Thus, a graph with $n$ types of multimodal attributes can be viewed as  $n$ structurally consistent unimodal graphs, each representing a single modality. For example, a graph where nodes have attributes from 3 modalities can be decomposed into 3 unimodal graphs, all maintaining the same structure across the modalities. Formally, a graph with multimodal attributes can be expressed as:
\begin{equation}
	\centering
	\label{eq3}
	\mathcal{G}=\bigg(\mathcal{V}, \mathcal{E}, \{\mathbf{x}_i\}^{|\mathcal{V}|}_{i=1}\bigg).
\end{equation}
Here, $\mathcal{V}$ and $\mathcal{E}$ indicate node and edge sets, respectively. $\mathbf{x}_i=[\mathbf{x}_i^1,\mathbf{x}_i^2,\cdots,\mathbf{x}_i^m]$ is the feature of node $i$, where $\mathbf{x}_i^j$ denotes the feature from modality $j$.
 
\textbf{Multimodal graph convolutional networks} are typical models for learning on multimodal attributes. For example, \citet{lian2023gcnet} proposed GCNet (a graph convolutional network-based graph complete model) for the conversation completion task. In GCNet, multimodal graphs are constructed based on conversation contents, incorporating modalities such as text and images. Each node in the graph contains multimodal features, though some features may be incomplete due to missing modalities. Then, the authors used a relational graph convolutional network to aggregate neighborhood information, enabling more advanced conversation understanding despite incomplete data.  \citet{9810283} propose a multicenter and multichannel pooling-based graph convolutional network for multimodal brain graph learning. For each brain region, they integrated information extracted from multimodal brain imaging data and used it as the feature vector for the brain region. A multimodal brain graph is then constructed, where nodes represent different brain regions. To effectively process this graph, the authors introduce a multicenter attention mechanism and multichannel pooling within the graph convolutional network. More recently, \citet{meng2024masked} proposed a method to fuse multimodal features into node attributes. They present an MGLRA (masked graph learning with a recurrent alignment) model, which aims to iteratively refine unimodal representations from multiple modalities. The model first employs a graph convolutional network with a random masking mechanism to facilitate multimodal fusion. Particularly, a cross-modal multi-head attention mechanism is then introduced to learn from the constructed graphs, effectively capturing interactive semantic information across modalities. 

\textbf{Transformer-based and contrastive learning-based methods} also attract great attention for learning on multimodal attributes. MarT~\citep{000100LDC23} is a Transformer model for multimodal analogical reasoning over knowledge graphs. The authors constructed a multimodal knowledge graph (MarKG), where each entity possesses multimodal attributes. This enables the model to reason over relations and analogies between entities across different modalities. \citet{chen2022multi} designed an MSNEA (multimodal siamese network for entity alignment) framework for entity alignment in multimodal knowledge graphs. They first fused entity features from three different modalities, including visual, relational, and attribute modalities, to gain multimodal entity representations. Subsequently, they present a multimodal contrastive learning module to achieve inter-modal enhancement fusion and avoid the overwhelming impact of weak modalities. To address the multimodal emotion recognition task, \citet{li2023joyful} proposed a joint modality fusion-based graph contrastive learning method. Specifically, they designed a multimodal fusion mechanism to construct multimodal graphs, where each node contains both unimodal-specific features and global contextual features, enabling a comprehensive information integration from different modalities. 

In addition, some other graph learning models are widely proposed for learning on multimodal attributes. \citet{TianZGMMC22} introduced a multimodal recipe representation learning model, Recipe2Vec, which leverages a heterogeneous graph neural network to embed multimodal information. They constructed heterogeneous recipe graphs by fusing visual, textual, and relational information as node attributes.
DGVAE~\citep{zhou2024disentangled} converts the visual and textual information of items into consistent embeddings, which are then used to construct item-item graphs where items are represented as nodes. To reconstruct the users' rating matrix and their textual preferences, DGVAE inputs the word vectors and rating vectors into the graph encoder independently. This approach captures multimodal information to enhance the effectiveness of multimodal recommendations. 3D-HaarFrame~\citep{li2024multimodal} is a multimodal graph learning model designed for student engagement prediction. It captures inter-modal correlations within textual, visual, and acoustic data by embedding the multimodal information into a unified graph. The model then applies spectrum-based graph learning to decompose the graph into a set of coefficients at different scales and orientations, efficiently compacting the multimodal representations for improved prediction accuracy.

\subsection{Learning on Multimodal Nodes}

Graphs with multimodal nodes consist of nodes derived from different modalities (see Figure~\ref{multimodal graph} (b)). Each node represents a unimodal entity, featuring attributes from a single modality. In these graphs, edges signify either inter-modal or intra-modal correlations. For instance, an edge connecting two nodes from different modalities represents the inter-modal correlation between them. As a result, learning on multimodal nodes is more effective in modeling inter-modal correlations than approaches that rely on multimodal attributes. Formally, edges represent inter-modal and intra-modal correlations are denoted as:
\begin{equation}
	\centering
	\label{eq4}
	e_{\text{inter}}=(v_i^j,v_k^m), e_{\text{intra}}=(v_i^j,v_l^j).
\end{equation}
Here, $v_i^j$ indicates node $i$ from modality $j$.

\textbf{Knowledge Graphs with Multimodal Nodes.}~Studies focusing on knowledge graphs with multimodal nodes are rising as a prominent research topic. For example, \citet{song2024scene} constructed a SceneMMKG (scene-driven multimodal knowledge graph), which incorporates both textual and visual nodes. This graph serves as input to a graph convolutional network to facilitate effective learning and representation. \citet{wang2020multimodal} proposed the MKGCN (multimodal knowledge graph convolutional network) model, focusing on knowledge graphs derived from articles containing both text and images. Nodes represent entities extracted from either text or images. MKGCN utilizes two graph convolutional layers to aggregate node features. This approach allows for the generation of high-quality article representations by effectively modeling the connections within text-text, image-image, and text-image relations. Moreover, \citet{lyu2021mdnn} proposed the MDNN (multimodal deep neural network) model, which incorporates a graph neural network module to learn from multimodal drug knowledge graphs for drug-drug interaction prediction. In this knowledge graph, nodes represent drugs and multimodal targets such as molecules and proteins, enabling the model to capture complex interactions between various drug-related modalities. \citet{zeng2023multi} constructed multimodal knowledge hypergraphs through multimodal instance bagging and concept aggregation. The multimodal instance bagging module selects instances by leveraging higher-order relations of hyperedges. A mixed instance bag, containing both images and texts, enhances the diversity of multi-semantic representations. The authors then employ a hypergraph convolutional layer to encode the entities within the graph.

Likewise, many works also focus on other types of graphs with multimodal nodes. For instance, \citet{wen2022graph} focused on multimodal single-cell graph learning. They introduced scMoGNN, a graph neural network framework that incorporates a cell-feature graph convolution with task-specific heads to analyze multimodal single-cell data across various tasks, such as modality prediction, modality matching, and joint embedding. The model is designed to capture the high-order structural information between cells and modalities. \citet{hu2023graph} designed a graph convolutional network-based graph fusion neural network model to enhance Freezing of Gait (FoG) detection by encoding multimodal information, such as footstep pressure maps and video recordings, into graph nodes. Initially, the features from each modality are encoded as node-level inputs, and a multimodal adjacency matrix is constructed to measure correlations across modalities. Subsequently, They applied graph convolutions to these inputs, focusing on modeling cross-modality relations. SMGCN~\citep{kim2024self} is a self-supervised multimodal graph convolutional network designed to operate on multimodal user-item graphs for recommendation tasks. Specifically, the user-item graphs, where edges between multimodal nodes capture cross-modal user preferences, are generated through collaborative multimodal alignment (CMA) and multimodal consistency regularization (MCR). \citet{tu2024adaptive} proposed AdaIGN (adaptive interactive graph network) for multimodal conversational emotion detection. They emphasized both intra-modal and inter-modal interactions by dynamically selecting nodes and edges across modalities. To achieve this, they employed the Gumbel Softmax trick for adaptive multimodal node selection.

%% file: chapters/5-ggl.tex

The goal of generative graph learning~\citep{guo2022systematic,liu2023generative,cao2024survey,li2025generative} is to understand and learn the distribution of existing graph data, enabling the generation of high-quality graph samples. Generative Graph Learning is significant in studying the underlying structural relationships within graph data, understanding patterns, associations, and hidden information within existing data. 
To this end, researchers have proposed a large number of synthetic datasets and real-world datasets for research purposes.

Depending on whether auxiliary information is required for graph generation, generative graph learning methods can be categorized into  the following two main types: unconditional generative graph learning and conditional generative graph learning.

\section{Unconditional Generative Graph Learning}

Unconditional deep graph generation aims to learn from real graphs $\mathcal{G}$ obtained through sampling, allowing the model's distribution $p_{\text{model}}(\mathcal{G}) $ to approximate the true distribution $p(\mathcal{G})$.
Based on differences in the generation process, these methods can be divided into two main categories: (1) \textbf{One-shot generation}: This approach constructs probabilistic graph models based on matrix representations, generating all nodes and edges simultaneously; (2) \textbf{Sequential generation}: This approach generates nodes and edges one by one in a sequential manner.

\subsection{One-Shot Generative Graph Learning}
The one-shot generative graph learning method maps the entire graph to a latent representation space with a unified probability distribution during the learning phase. In the inference phase, it generates the graph in one step by sampling from this latent representation space.
Based on the approach used for graph generation, this approaches can be further divided  into \textbf{adjacency-matrix-based} methods and \textbf{edge-list-based} methods.

\paragraph{Adjacency-matrix-based Methods}
The topology of a graph is typically represented by an adjacency matrix, node attribute matrix, and edge feature matrix. Adjacency matrix-based methods primarily aim to learn the underlying distribution of these matrices. During the inference phase, these methods generate all three matrices in a single generation process.

 The fundamental challenge lies in achieving a mapping that balances expressiveness and efficiency, often requiring a trade-off between the two. A highly expressive mapping can capture intricate patterns and complex relationships but may come at the cost of increased computational complexity. Conversely, an efficient mapping ensures scalability and faster computations but might sacrifice the ability to represent detailed structural and feature interactions. Finding the optimal balance between these aspects is crucial for developing effective models.
 
 Existing methods based on deep neural network frameworks can be  divided into four types:

\begin{figure}[ht]
  \centering
  \includegraphics[width=0.8\linewidth]{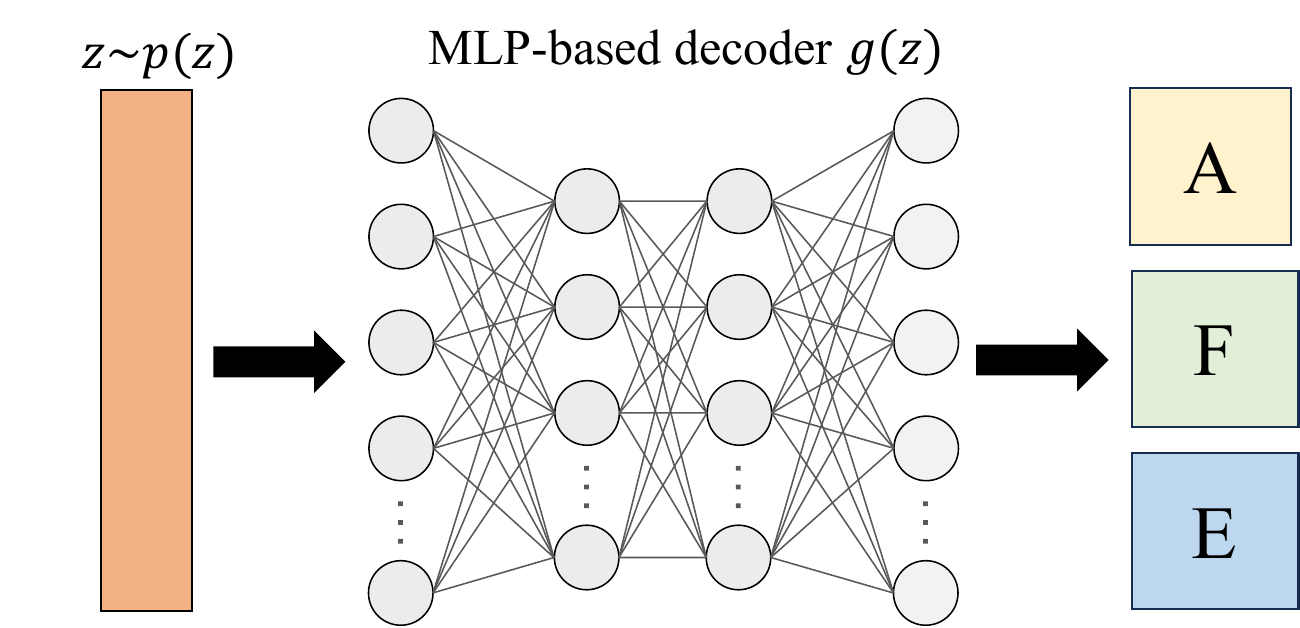}
  \caption{Schematic of MLP-Based Graph Generation.}
  \label{Schematic of MLP-Based Graph Generation.}
\end{figure}

(1) \textbf{MLP-Based Methods.} These methods use simple MLPs~\citep{polsterl2021adversarial,zahirnia2024neural} to construct the graph decoder \( g(\textbf{z}) \). 
As shown in Figure \ref{Schematic of MLP-Based Graph Generation.}, during the inference phase, a latent graph representation \( \textbf{z} \), which is sampled from a statistical distribution \( p(\textbf{z}) \), is passed through an MLP-based decoder \( g(\textbf{z}) \), which generates the adjacency matrix \( \textbf{A} \). Depending on the task requirements, node features \( \textbf{F} \) and edge features \( \textbf{E} \) may also be generated simultaneously.

\begin{figure}[ht]
  \centering
  \includegraphics[width=0.9\linewidth]{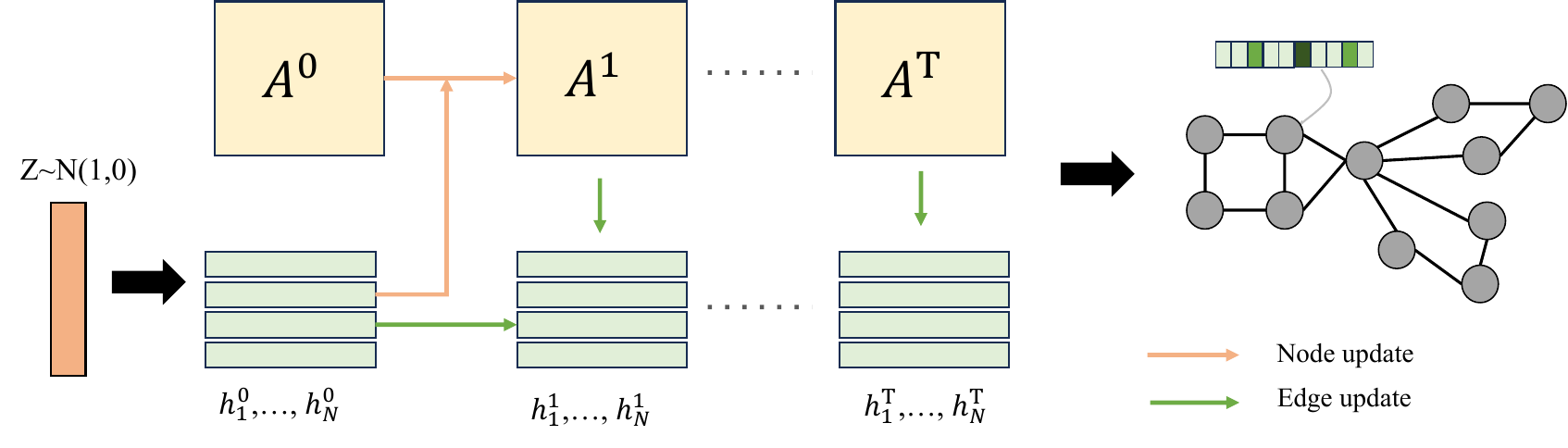}
  \caption{Schematic of Message-Passing-Based Graph Generation.}
  \label{Schematic of Message-Passing-Based Graph Generation.}
\end{figure}

(2) \textbf{Message-Passing-Based Methods.} 
These methods generate graphs by iteratively optimizing the initial graph using Message Passing Neural Networks (MPNN)~\citep{gilmer2017neural}. As shown in Figure \ref{Schematic of Message-Passing-Based Graph Generation.}, an initial adjacency matrix $\textbf{A}_0$   and initial node latent representations \( \textbf{H}_0 \) are generated first using the sampled latent representation \( \textbf{z} \).

The initial matrices \(\textbf{A}^0\) and \(\textbf{H}^0\) are then iteratively updated through multiple MPNN layers, producing \(\textbf{A}^T\) and \(\textbf{H}^T\) in the \(T_{th}\) intermediate layer, which explicitly capture higher-order correlations between nodes and edges. 
Such generators can be built using various learning frameworks, including VAEs, GANs~\citep{flam2021mpgvae}, or simpler score-based frameworks~\citep{niu2020permutation}.

\begin{figure}[ht]
  \centering
  \includegraphics[width=0.9\linewidth]{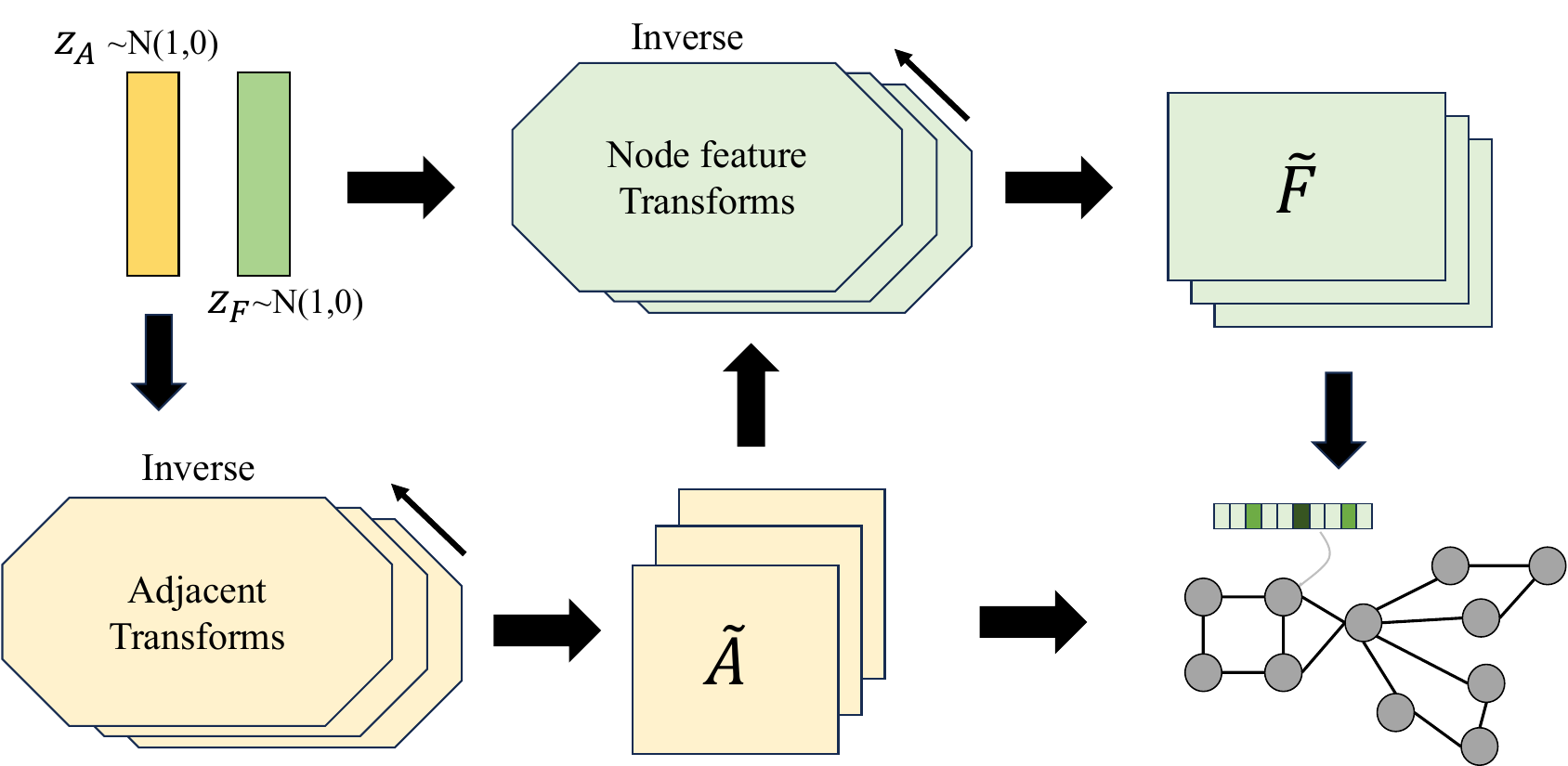}
  \caption{Schematic of Flow-Based Graph Generation.}
  \label{Schematic of Flow-Based Graph Generation.}
\end{figure}

(3) \textbf{Flow-Based Methods.} 
As shown in Figure \ref{Schematic of Flow-Based Graph Generation.}, flow-based methods~\citep{madhawa2019graphnvp} generate the graph \( \mathcal{G} = (\textbf{A}, \textbf{F}) \) and the sampled latent prior \( \textbf{z} \) through a unique invertible function.
Specifically, flow-based methods aim to learn a series of invertible functions \(z = f_L \circ ... \circ f_1(G)\) that convert graph data \(\mathcal{G}\) into a latent space representation \(\textbf{z} \sim P(\textbf{Z})\) with the same dimensionality.
During the inference phase, the model uses the inverse transformations \(\mathcal{G} = f^{-1}_1 \circ ... \circ f^{-1}_L(\textbf{z})\) to reconstruct and generate \(\mathcal{G}\) from the latent representation sampled from \(\textbf{z} \sim P(\textbf{Z})\).

GraphNVP~\citep{madhawa2019graphnvp} is the first to use invertible functions to generate flow-based molecular graphs.
Subsequent research, such as MoFlow~\citep{zang2020moflow}, MolGrow~\citep{kuznetsov2021molgrow}, and EFlow~\citep{zhang2024effective}, refines the design of invertible functions to achieve improved graph generation performance.

(4) \textbf{Diffusion-Based Methods.} In recent years, diffusion models~\citep{cao2024survey} have rapidly emerged as a new generative paradigm and are widely applied across various fields. 
The working mechanism of diffusion models is divided into two stages: forward diffusion and reverse diffusion. 
In the forward diffusion stage, the model starts from a simple initial distribution and progressively adds noise, introducing a structured amount of noise at each step. 
In the reverse diffusion stage, the model begins with noisy data samples and gradually removes the noise, recovering the original data.

Currently, there are three main paradigms for diffusion models: Denoising Diffusion Probabilistic Models (DDPM)~\citep{chen2023efficient}, Noise Conditional Score Networks (NCSN)~\citep{shi2021learning}, and Score Stochastic Differential Equation (Score SDE)~\citep{yang2024graphusion}. DDPM and NCSN leverage the concepts of non-equilibrium thermodynamics and score matching, respectively, to learn the forward and reverse functions of the diffusion process. 
Score SDE further models the diffusion process through SDE, providing a continuous-time framework for diffusion and reverse processes.

\paragraph{Edge-List-Based Methods.} Although adjacency-matrix-based methods possess strong expressive power in learning the distribution of adjacency matrices, they are limited in terms of memory and time efficiency. To address this issue, edge-list-based methods have been proposed. These methods focus on learning local patterns, which are generally more suitable for handling large graphs with simple global patterns.

Edge-list-based methods typically use a generative model to learn the edge probabilities, where all edges are generated independently. These methods are generally applied to learn and then generate a new graph that includes existing nodes within a large-scale graph. 
In edge probability generation methods, existing research can be categorized into the following two types:

(1) \textbf{Random-Walk-Based Methods.} In random-walk-based methods, the edge probabilities are estimated by counting the occurrences of edges across numerous random walks, which are sampled from a learned distribution of the graph. NetGAN~\citep{bojchevski2018netgan} is introduced to learn the random walk distribution on an observed graph, and then generate a set of random walks. 

In random walk-based methods, the edge probability is estimated by counting the occurrences of edges across multiple random walks sampled from the graph's learned distribution. NetGAN~\citep{bojchevski2018netgan} is the first to introduce a method for learning the random walk distribution on an observed graph. During the inference phase, NetGAN generates a set of random walks and then constructs a scoring matrix \( \textbf{S} \in \mathbb{R}^{N \times N} \) based on the frequency of edge occurrences in these walks. 
Subsequently, an edge probability matrix \( \tilde{A} \) is computed from the scoring matrix, where \( \tilde{A}_{i,j} = \frac{S_{i,j}}{\sum_{u,v} S_{u,v}} \). Finally, individual edges \( A_{i,j} \) are generated based on this probability matrix through an efficient sampling process.

Inspired by NetGAN, subsequent research further refines random walks to better suit various tasks. For example, the temporal random walk is used in generative temporal graph learning~\citep{gupta2022tigger}. Similarly, TIARA~\citep{lee2023time} utilizes the time-aware random walks to enhance the dynamic graph.

(2) \textbf{Node-Similarity-Based} 
This method generates the edge probability based on the relationships between given or sampled node embeddings~\citep{thomas2016variational,salha2019degeneracy}. Specifically, let \( \textbf{Z}_i \in \mathbb{R}^L \) be the latent representation of node \( v_i \), then the probabilistic adjacency matrix \( \tilde{\textbf{A}} \) is generated using the node representations \( \textbf{Z} \in \mathbb{R}^{N \times L} \). Finally, edges \( \textbf{A}_{i,j} \) are obtained by sampling from the matrix \( \tilde{\textbf{A}} \). 

The primary goal of existing research is to determine how to generate an effective \(\tilde{A}\).
Some studies~\citep{grover2019graphite} sample the latent representations \( Z \) of nodes from a normal distribution and then compute the inner product of \( Z \) to generate the adjacency matrix \( \tilde{A} \). The adjacency matrix \( A \) is subsequently sampled from \( \tilde{A} \).
Other works~\citep{salha2019gravity} calculate \(\tilde{A}_{i,j}\) by measuring the \(\ell_2\) norm between the representations of two nodes.

In addition to the two categories of research mentioned above, there are also other edge-list-based approaches for graph generation. For instance, a few studies~\citep{jang2023simple} propose edge-list-based representation methods to achieve lightweight graph generation. These approaches often use Depth-First Search (DFS) or Breadth-First Search (BFS) to construct edges, ensuring unique node ordering while maintaining high scalability. Additionally, \citet{luo2023fast} focused on the global information implied by the graph's spectrum. By learning and generating the graph's eigenvalues and eigenvectors, these methods aim to capture the shape information of the graph, leading to an improved generation performance.

\subsection{Sequential Generative Graph Learning}
 
Sequential generation methods view graph generation as a sequence decision process, where the graph \( \mathcal{G} \) is represented as a series of components \( S = \{s_1, \dots, s_N\} \), with each \( s_i \in S \) considered a generative unit. 
During the generation process, different components are generated sequentially based on the already generated subgraph, ultimately forming a complete graph.

One core challenge in sequential generation methods is how to decompose the graph into multiple components to facilitate their sequential generation, i.e., determining the generation order of \( s_i \). Based on the type of generation sequence, sequential generation methods can be further divided into four categories, that are
\textbf{node-Sequence-Based methods}, \textbf{edge-Sequence-based methods}, \textbf{motif-Sequence-based methods}, and \textbf{rule-sequence-based methods}.

\begin{figure}[ht]
  \centering
  \includegraphics[width=0.8\linewidth]{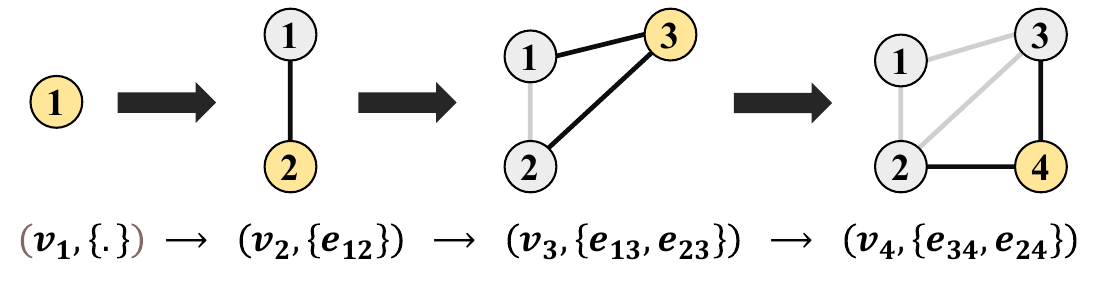}
  \caption{Schematic of Node-Sequence-Based Graph Generation.}
  \label{Schematic of Node-Sequence-Based Graph Generation.}
\end{figure}

\paragraph{Node-Sequence-Based Methods.} As shown in Figure \ref{Schematic of Node-Sequence-Based Graph Generation.}, this method models the graph as a sequence based on a predefined node order \( V = \{v_1, v_2, \dots, v_N\} \). Each generative unit \( s_i \) in the sequence \( S \) can be represented as a tuple \( s_i = (v_i, \{e_{j,i}\}_{j<i}) \). At each step of the generation process, the generator produces a node \( v_i \) along with its corresponding set of edges \( \{e_{j,i}\}_{j<i} \).

In node-sequence-based generation, the generative unit \( s_i \) first generates a node \( v_i \) and a set of associated edges \( \{e_{i,j}\}_{j<i} \) based on the current generated subgraph \( \mathcal{G}_i \). When generating the edges associated with each node, there are primarily two methods:

(1) \textbf{Collective-Associated-Edge-Based Methods.}~\citet{chen2021order} predicted the entire edge set \( \{e_{i,j}\}_{j<i} \) in one operation by considering the relationships between node \( v_i \) and existing nodes in \( G_i \). Most studies rely on the adjacency vector \( \textbf{a}_i \) to predict all potential edges between the newly generated node \( v_i \) and the existing nodes. Therefore, each generative unit can be further represented as \( s_i = (v_i, \textbf{a}_i) \), and the sequence can be expressed as \( Seq(G) = \{(v_1, \textbf{a}_1), \dots, (v_N, \textbf{a}_N)\} \).

The objective is to learn the following distribution:

\[
P(V, A) = \prod_{i=1}^{N} p(v_i | v_{<i}, \textbf{a}_{<i }) p(\textbf{a}_{i} | v_{\leq i}, \textbf{a}_{<i }),
\]

where \( v_{<i} \) represents the nodes generated before \( v_i \), and \( \textbf{a}_{<i } \) represents the adjacency vectors generated before \( \textbf{a}_{i } \). This joint probability can be modeled using sequence-based architectures, such as auto-regressive models~\citep{kong2023autoregressive}.
  
(2) \textbf{Progressive-Associated-Edge-Based Methods.} 
The time complexity of the collective-associated-edge-Based method is \(O(N^2)\), which can be computationally expensive for sparse graphs. To address this, progressive-associated-edge generation~\citep{liu2018constrained,} has been proposed. 
This method incrementally selects nodes from the existing node set \( v_{<i} \) to connect with the current node \( v_i \), continuing until the desired number of connections is reached. 
Specifically, it first determines the size of the edge set \( \{e_{i,j}\}_{j<i} \) for node \( v_i \) and then selects nodes to connect from \( v_{<i} \). If \( M \) represents the number of edges, then the time complexity for this method is denoted by \( O(MN) \). 

\begin{figure}[ht]
  \centering
  \includegraphics[width=0.8\linewidth]{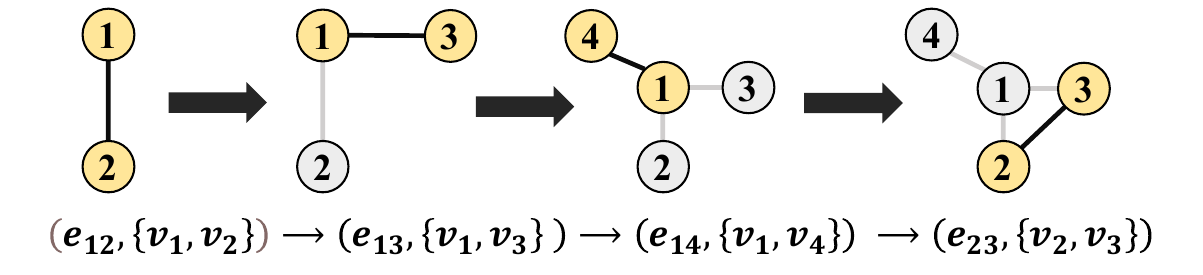}
  \caption{Schematic of Edge-Sequence-Based Graph Generation.}
  \label{Schematic of Edge-Sequence-Based Graph Generation.}
\end{figure}

\paragraph{Edge-Sequence-Based Methods.} As illustrated in Figure \ref{Schematic of Edge-Sequence-Based Graph Generation.}, these methods represent a graph as a sequence of edges, generating each edge along with its two associated nodes step by step. With a total complexity of \( O(M) \), where \(M\) is the number of edges, these approaches are particularly advantageous for handling sparse graphs, as they avoid the overhead associated with processing a full adjacency matrix.
These methods define an order for the edges in the graph \( \mathcal{G} \) and a ranking function \( a(\cdot) \) to index the nodes. In such methods, each unit in the sequence \( S \) is represented as a tuple \( s_i = (a(u), a(v)) \), where \( a(u) \) and \( a(v) \) are the indices of nodes \( u \) and \( v \), respectively.
In existing approaches, the generative unit \( s_i \) can be modeled in two ways: one assumes that \( a(u) \) and \( a(v) \) are independent~\citep{goyal2020graphgen}, while the other assumes they are correlated~\citep{bacciu2019graph,bacciu2020edge}.

\begin{figure}[ht]
  \centering
  \includegraphics[width=0.8\linewidth]{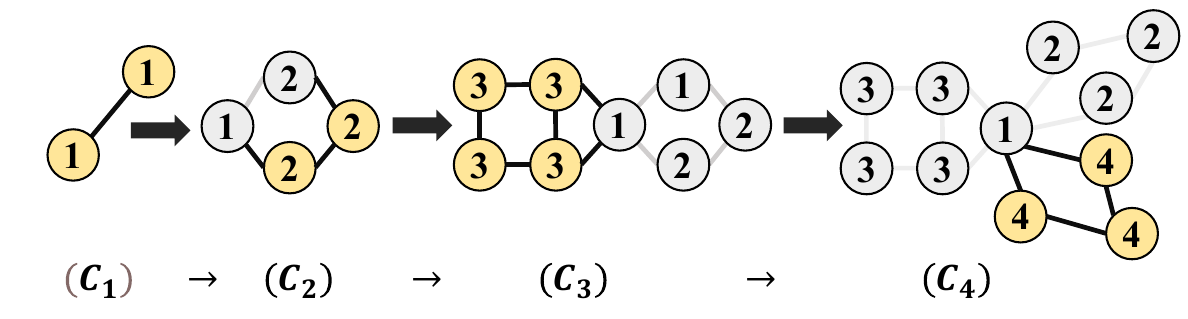}
  \caption{Schematic of Motif-Sequence-Based Graph Generation.}
  \label{Schematic of Motif-Sequence-Based Graph Generation.}
\end{figure}

\paragraph{Motif-Sequence-Based Methods.} Although the aforementioned sequence generation methods perform well in preserving node-to-node relationships, they often struggle to capture higher-order relationships. For instance, recommendation systems, social networks contain numerous triangles, and gene regulatory networks. These scenarios involve sub-graph patterns known as motifs. 

To address this issue, many motif-sequence-based methods have been proposed. As shown in Figure \ref{Schematic of Motif-Sequence-Based Graph Generation.}, these methods represent a graph \( \mathcal{G} \) as a series of graph motifs \( Seq(\mathcal{G}) = \{C_1, ..., C_M\} \). During the generation process, at each step, a new motif block \( C_i \) is generated based on the current graph \( \mathcal{G}_{i-1} \), and then connected to \( \mathcal{G}_{i-1} \) to obtain \( \mathcal{G}_i \).
These methods improve efficiency by generating a set of nodes and edges simultaneously at each step using a series of motifs.

One key challenge faced by these methods is determining how to connect the existing graph \( \mathcal{G}_{i-1} \) with the newly generated graph component \( C_i \).
Based on the different connection strategies, existing research can be divided into the following two types:

(1) \textbf{Domain-Agnostic-Based Methods.}  These methods~\citep{liao2019efficient,jang2023hierarchical} aim to generate general graphs without requiring domain-specific knowledge. Similar to collective-associated-edge-Based methods in node-sequence-based generation, these methods generate an adjacency vector for each edge. However, the key difference is that while collective-associated-edge-Based methods produce a single node at each step, domain-agnostic-based methods generate multiple nodes at each step.

(2) \textbf{Domain-Specific-Based Methods.} Domain-Specific-Based methods~\citep{jin2020hierarchical,liu2023using} involve using domain knowledge to define graph motifs and their connections, such as molecular fragments~\citep{jin2018junction} or structural motifs~\citep{jin2020hierarchical} in molecular graphs, or temporal motifs~\citep{liu2023using} in temporal networks. These motifs help define how to connect the generated motifs to construct a complete graph.

\begin{figure}[ht]
  \centering
  \includegraphics[width=0.8\linewidth]{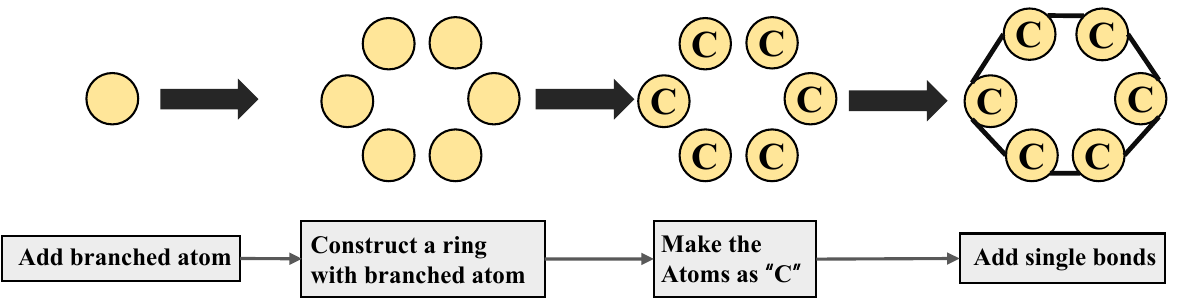}
  \caption{Schematic of Rule-Sequence-Based Graph Generation.}
  \label{Schematic of Rule-Sequence-Based Graph Generation.}
\end{figure}

\paragraph{Rule-Sequence-Based Methods.}  Although the three aforementioned methods are generally effective in end-to-end graph generation, they often fall short in ensuring the generation of "valid" graphs, which adhere to correct syntax and constraints in fields like programming languages and molecular modeling. To address this issue, rule-sequence-based methods are introduced. As shown in Figure \ref{Schematic of Rule-Sequence-Based Graph Generation.}, these methods guide the graph construction process by generating a series of rules or commands. This approach is particularly useful in scenarios where the target graph must meet specific constraints or adhere to certain grammar rules to be valid. For example, molecules must obey fundamental properties such as charge conservation, limiting the choices for node types and edge patterns in molecular graphs.

\citet{dai2018syntax} introduces a semantic constraint component during the parse tree generation phase, enabling the generator to ensure both syntactic and semantic validity. \citet{wang2024grammar} introduced Grammar Prompting, which allows LLMs to use domain-specific constraints and external knowledge during in-context learning. Grammar Prompting enhances model capabilities by incorporating a specialized grammar into each example demonstration.

\section{Conditional Generative Graph Learning}

The goal of conditional generative graph learning is to utilize the condition \( y \) (i.e., auxiliary information) to learn to generate a graph \( \mathcal{G} \). Depending on the learning scenario, the condition \( y \) can be a graph, semantic context, image, etc. Based on the type of condition \( y \), existing conditional generative graph learning methods can generally be divided into the following three categories.

\subsection{Conditional Graphs Generative Graph Learning}

Conditional graph generative learning aims to generate a target graph \( \mathcal{G}_T \) based on an input graph \( \mathcal{G}_S \). These methods are commonly used in tasks involving graph transformation, such as malware network synthesis~\citep{guo2022deep}, molecular graph editing~\citep{zhong2023retrosynthesis}, protein structure generation \citep{guo2020interpretable}, and retrosynthesis prediction \citep{shi2020graph}.

For the retrosynthesis prediction task, \citet{shi2020graph} proposed the G2Gs model, which splits a molecule into synthons by breaking the reaction center. Then, a series of graph transformations is applied to generate the corresponding reactants. \citet{zhong2023retrosynthesis} proposed the Graph2Edits model for retrosynthesis prediction, aimed at performing molecular graph editing. \citet{wu2021graph} formalized the handwritten mathematical expression recognition problem as a graph-to-graph learning task and proposed a G2G model, which has shown significant improvement in experimental results.

In other research directions, \citet{hou2024dag} proposed the DAVA model, which generates social propagation graphs by integrating topological analysis of propagation graphs with corresponding user attributes as prior knowledge. \citet{guo2020interpretable} introduced the NED-VAE model to tackle disentangled representation learning for attributed graphs.

\subsection{Conditional Semantic Context Generative Graph Learning}

Conditional semantic context generative graph learning aims to generate a target graph \( \mathcal{G}_T \) based on an input semantic context. 
When a graph is associated with the semantic context, it enhances the information contained in the graph. As a result, generating graph structures based on given semantic conditions holds significant value.

The semantic context is typically represented as additional meta-features, such as categories, labels, modalities, or any other auxiliary information that can be intuitively expressed as a vector \( \textbf{C} \).
In conditional semantic context generative graph learning, the primary challenge is how to effectively incorporate the semantic context into the generative process to achieve meaningful graph generation. This involves determining the best way to integrate the semantic information \( \textbf{C} \) into the model so that the generated graph \( \mathcal{G}_T \) accurately reflects the conditions provided by the context.

In the field of Computer Vision (CV), scene graph generation aims to generate corresponding scene graphs based on semantic information from images, and several effective methods have been proposed. TextPSG~\citep{zhao2023textpsg} generates scene graphs by using textual descriptions of images as semantic information. TsCM~\citep{sun2023unbiased} integrates semantic conditions through causal inference. PE-Net~\citep{zheng2023prototype} models entities and predicates using compact and distinct representations with prototype alignment, establishing matches between entity pairs and predicates within a shared embedding space for relation recognition. Other methods~\citep{lu2021context,kundu2023ggt} sample from underlying semantic structures between entities in images and use modules like Transformers to guide model learning.

Other research directions also have their own approaches to integrating semantic information. CONDGEN~\citep{yang2019conditional} constructs a conditional graph variational autoencoder by concatenating the semantic context information \(\textbf{C}\) with the latent representation of nodes \(\textbf{Z}_i\) to incorporate the semantic context into the graph generation process.
GPrinFlowNet \citep{mo2024graph} builds on the GFlowNet \citep{bengio2021flow} framework and employs a generation curriculum from low to high frequency. It samples from complex conditional distributions to learn the small eigenvalues of the graph's adjacency matrix and their associated eigenvector information.

\subsection{Conditional Sequence Generative Graph Learning}

The goal of conditional sequence generative graph learning is to  use the input sequence \( X \) to generate a target graph \( \mathcal{G}_T \), which commonly arises in fields such as NLP~\citep{zhang2024coreference}, video analysis \citep{cong2023reltr}, and time series mining \citep{you2022roland}.

In the NLP domain, conditional sequence generative graph learning is applied to tasks such as Semantic Parsing and mind map generation. 
\citet{bai2022graph} considered enhancing AMR parsing and generation by pretraining on semantic graphs. \citet{hu2021efficient} proposed an EMGN that considers document-level semantics through sequence-to-graph modeling. \citet{zhang2024coreference} introduced a CMGN to incorporate external structural knowledge for mind map generation.

In the multimedia domain, generating dynamic scene graphs from videos~\citep{nag2023unbiased,lin2024td2} is more challenging compared to generating scene graphs from images. This is because videos need to be processed as sequences to generate the corresponding scene graphs. The research in this area is still in its early stages, with new sub-problems emerging \citep{yang2023panoptic} and relevant methods \citep{nguyen2024hig} being proposed to address them.
To address the issues of contextual noise and label bias in dynamic scene graph generation from videos, TD²-Net~\citep{lin2024td2} uses the AD-Trans module to address the issue of contextual noise. Additionally, an Asymmetric Re-weighting Loss is introduced to tackle label bias problems.
To address the challenges that existing methods face in tackling the visual interactivity understanding problem, \citet{nguyen2024hig} introduce a new dataset named ASPIRE and proposed a novel Hierarchical Interlacement Graph (HIG) to deeply analyze scene changes across five different tasks.

In other directions, conditional sequence generative graph learning methods have been designed to address specific challenges. For instance, in the context of privacy preservation, \citet{xiang2023privacy} proposed a Gromov-Wasserstein Autoregressive (GWAR) model, which does not require correspondence information, thereby maintaining the privacy of nodes in evolving graphs. For synthetic dataset challenges, \citet{zheng2024temporal} introduced the concept of Time-Bounded Communities to represent communities that are destined to disband. Based on patterns collected from real-world data, they design GTB, a temporal graph Generation method featuring Time-Bound communities.

%% file: chapters/6-xgl.tex

Despite the remarkable success of graph learning, the prevalence of graph neural networks in graph learning has led to the inherent “black box” issue of graph learning.
Insufficient transparency of graph neural networks prevents end-users and model developers from understanding predictions and decision-making processes, leading to a crisis of trust between humans and models.
This limits the possibility of deploying graph learning models into safety-critical scenarios such as healthcare, justice, and finance.

Providing explanations for graph learning models is an effective manner to enhance trust in their predictions.
Consequently, explainable graph learning has received a lot of attention from the graph learning research community in recent years \citep{yucage,huang2025sehg,nandan2025graphxai} and has been applied in various fields.
Unlike explaining traditional deep learning models, there are challenges specific to explaining graph learning models.
First, the topological information of graph data is complex. Its node features and topological relations impact on the prediction results are more difficult to be captured compared to structured data.
Second, the predictions of graph neural networks contain higher-order dependencies. The multi-layer message aggregation mechanism of graph neural networks causes the representation of nodes to depend on the information of multi-hop neighbors, which raises the difficulty of explaining graph neural networks.
As a result of these challenges, explainable graph learning has been a key research topic in the graph learning field, and numerous methods have been presented to improve the explainability of graph learning.
To this end, we consider explainable graph learning as a branch of graph learning and provide a systematic survey of studies within this branch, thereby providing an updated overview.

\begin{figure}[ht]
	\centering
	\includegraphics[width=0.9\linewidth]{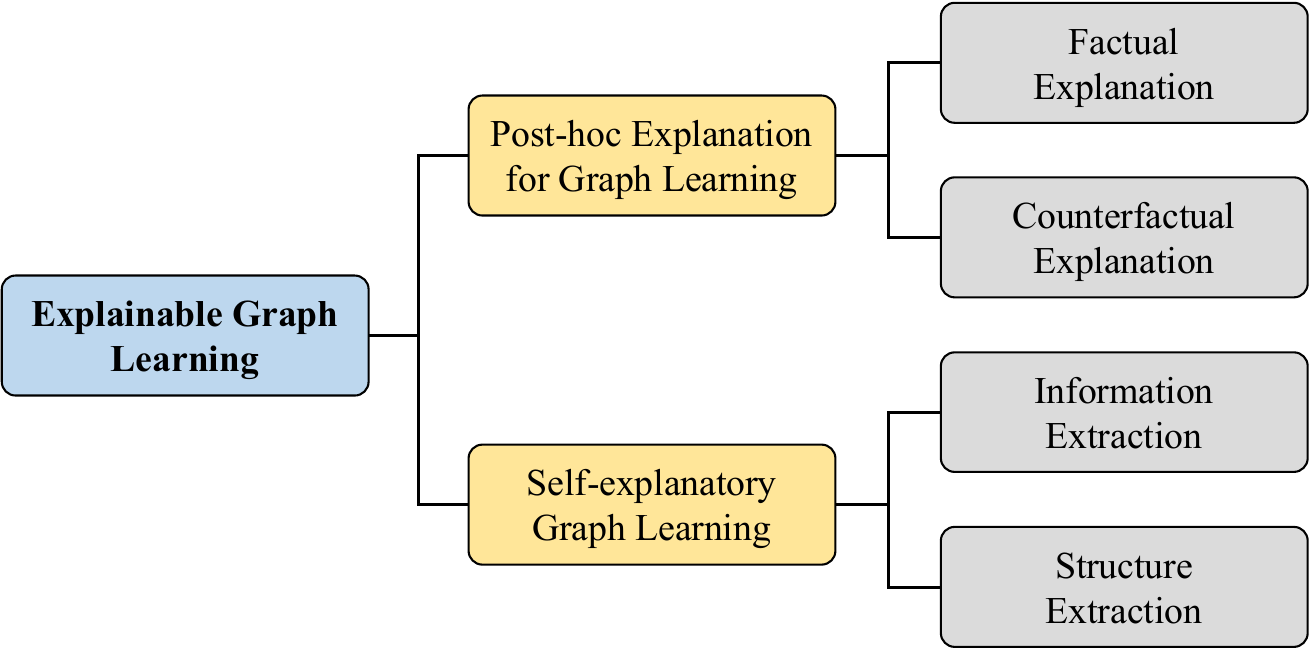}
	\caption{The taxonomy for explainable graph learning.}
	\label{The taxonomy for explainable graph learning.}
\end{figure}

In this survey, we categorize existing explainable graph learning methods into post-hoc explainable methods (explainers) and self-explanatory models (ante-hoc explainable methods).
Post-hoc explainable methods explain the predictions of a graph learning model by constructing an explainer independent of the model, which explains the predictions without accessing the internal parameters of the model.
For post-hoc explainable methods, we further divide them into factual explanation and counterfactual explanation based on the form of explanation.
In contrast, self-explanatory models modify the architecture of the original graph learning model or introduce explainable modules to make the model itself “white-box”.
For self-explanatory methods, we divide them into information extraction-based methods and structure extraction-based methods according to the manner in which they are implemented. 
Figure~\ref{The taxonomy for explainable graph learning.} illustrates the proposed taxonomy for explainable graph learning.

\label{c-intro} 
\section{Post-hoc Explanation for Graph Learning}
\begin{figure}[ht]
	\centering
	\includegraphics[width=0.9\linewidth]{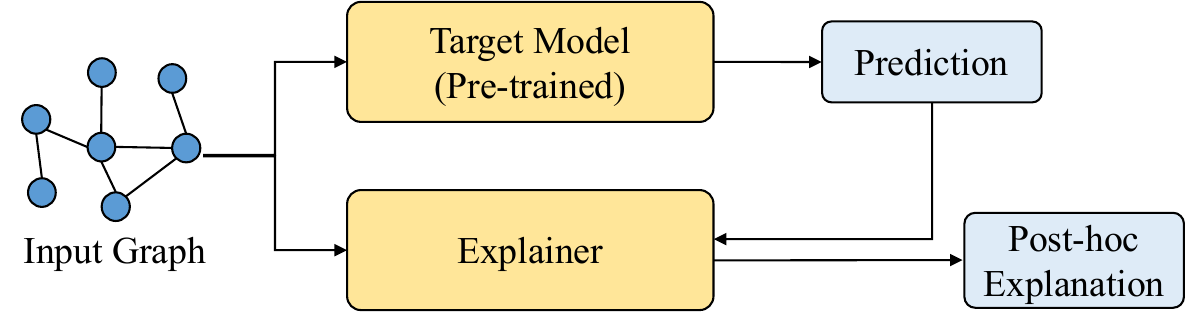}
	\caption{Post-hoc Explanation for Graph Learning.}
	\label{Post-hoc Explanation for Graph Learning.}
\end{figure}

The aim of post-hoc explanation for graph learning is to explain the predictions of the graph learning model (target model) without modification.
The post-hoc explanation is provided by a model-agnostic explainer, which defaults to the parameters of the target model being inaccessible.
The generation process of the post-hoc explanation is shown in Figure~\ref{Post-hoc Explanation for Graph Learning.}. The explainer receives input samples and predictions from the target model to identify input topological information that significantly affects the prediction, and feeds this information back to the human as a post-hoc explanation.

\begin{figure}[ht]
	\centering
	\includegraphics[width=1.0\linewidth]{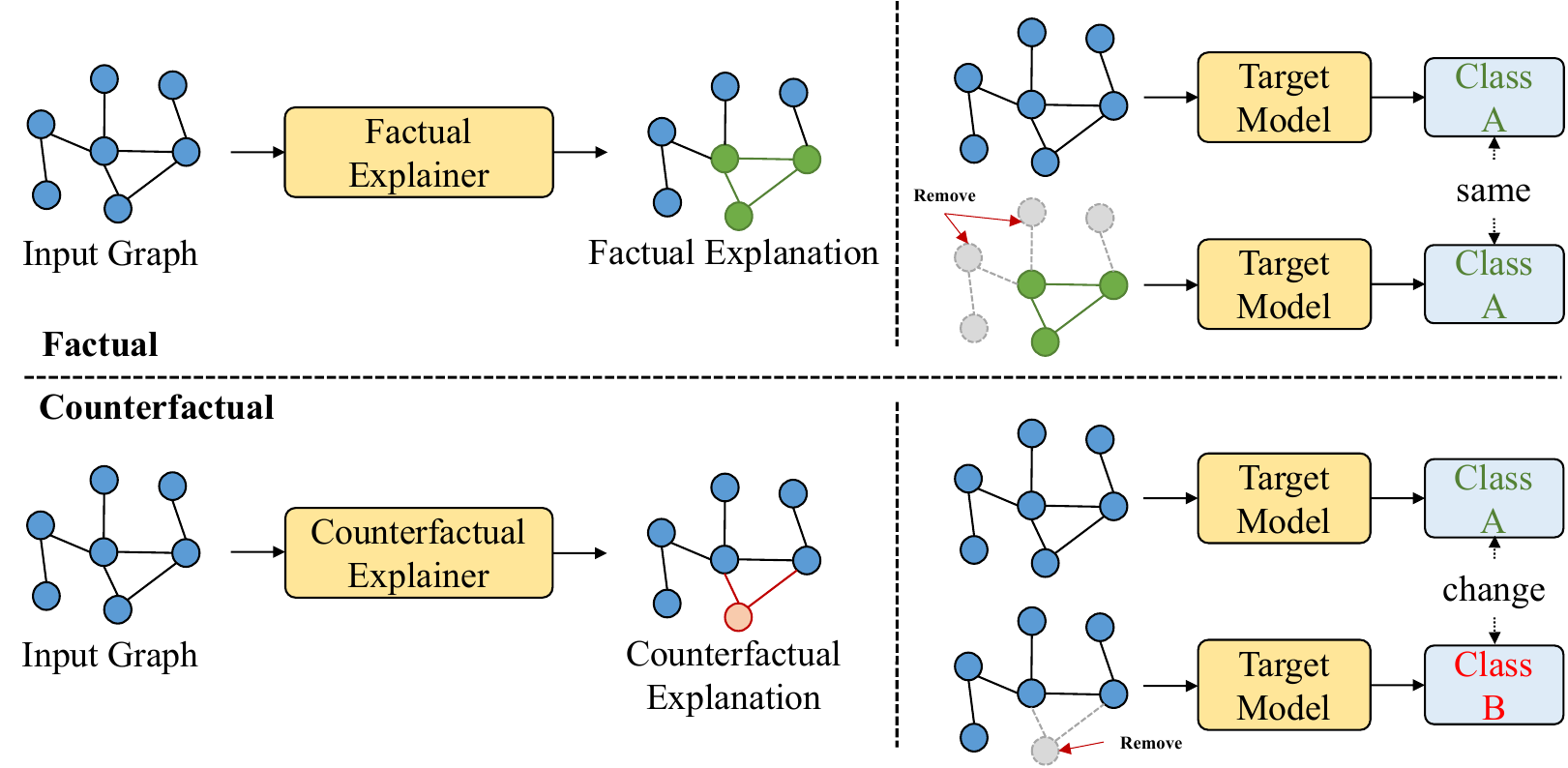}
	\caption{Factual and counterfactual explanation. The factual explanation retains key information from the input graph in order to maintain the same prediction as the original input (upper half), while the counterfactual explanation removes key information in order to change the prediction (lower half).}
	\label{Factual and counterfactual explanation.}
\end{figure}

Furthermore, we classify post-hoc explanations into two types: the factual explanation and the counterfactual explanation.
As shown in Figure~\ref{Factual and counterfactual explanation.}, the factual explanation focuses on capturing a subset of key topological information on the input graph that retains the predictions of the target model.
In contrast, the counterfactual explanation focuses on capturing a minimal subset of topological information on the input graph that significantly changes the predictions.
\subsection{Factual Explanation}
The mainstream paradigm for generating factual explanations is to quantify the importance of edges and nodes on the input graph to the prediction and filter the important edges and nodes as explanations for the prediction.
The manner in which an explainer quantifies importance can be categorized into three types: generative-based, search-based, and proxy-based.

\textit{Generative-based}.
Generative-based explainers typically utilize generative models to generate new graph explanations.
The generative model learns representations of input graphs and constrains the maximization of mutual information between the graph representations and predictions to fit a probability distribution of the target model.
Furthermore, the generative model reconstructs the explainable subgraphs based on the learned graph representations.
Numerous existing explainers fall into this type.
For instance, \citet{lin2021generative}~utilized the variation in model loss caused by removing edges from the input graph to assess the causal effect of edges on predictions, which in turn guides the generation of explanatory subgraphs.
In addition, some studies attempt to optimize graph editing strategies with reinforcement learning for guiding the generation of explanatory subgraphs. \citet{yuan2020xgnn}~utilized reinforcement learning to guide the editing of the input graph (removing or adding edges) for constraining the explainer to generate subgraphs that contain critical information.
Similarly, \citet{NEURIPS2021_be26abe7}~utilized reinforcement learning to guide the decoder for reconstructing explanatory subgraphs, and leverage a fully connected layer to capture the representation of the input graph.
Besides reinforcement learning methods, some studies~\citep{wang2023gnninterpreter, li2023dag}~constrain the explanatory subgraphs to contain critical topological information by learning probabilistic generative graph distributions to approximate the spatial distance between the representation of the explanatory subgraph and the representation of the input graph.
Despite the superior generalization and computational efficiency of generative explainers, subgraph generation processes are still black-box since they require learning the graph representation.

\textit{Search-based}.
In contrast to black-box generative explainers, search-based explainers identify the importance of edges and nodes in the input graph through a white-box search strategy without learning the representation of the input graph.
For instance, some studies~\citep{ying2019gnnexplainer, luo2020parameterized, wang2021towards, funke2022zorro, yuan2021explainability, zhang2022gstarx}~inject perturbations to each edge and node on the input graph for searching subgraph structures that significantly influence the prediction.
The explainer measures the influence on predictions by perturbing different edges and nodes to capture important topological information.
Inspired by the change rate of gradient representations,~\citet{pope2019explainability}~measured the sensitivity of predictions to the change rate of gradient representations from topological information.
Theoretically, predictions are sensitive to gradient changes in key topological information.
Therefore, the explainer quantifies such sensitivity to capture the key explanatory subgraphs.
In addition, some explainers~\citep{feng2022degree, xiong2023relevant}~decompose the predictions of the target model and back-propagate the decomposition values layer by layer up to the input layer. The input layer searches for a feature subset that critically influences the prediction according to the decomposition values.

\textit{Proxy-based}.
Proxy-based explainers aim to fit the target model with a white-box model as the proxy model.
The white-box models employed by such explainers are typically linear, and they simulate predictions by fitting probability distributions of the target model.
The explainer filters the key topological information by extracting the decision-making process of the white-box model.
\subsection{Counterfactual Explanation}
The purpose of the counterfactual explanation is to change the original prediction by minimally modifying the topological information of the input graph.
There are two methods for generating the counterfactual explanation: optimization-based methods and learning-based methods.

\textit{Optimization-based}.
The optimization-based method identifies a subset of features in the topological space of the input graph by minimizing modifications to the graph features while maximizing the impact on the model prediction. This ensures that removing this subset from the input graph leads to a significant change in the model output.
This method involves solving an optimization problem that maximizes the objective function, without the need to train a new model.
For example, some methods modify edges and nodes by perturbing the adjacency matrix of the input graph and optimize the perturbation by minimizing the probability of the original prediction and the magnitude of the perturbation~\citep{tan2022learning,kang2024unrexplainer,qiu2024generating,yang2023counterfactual}.
Additionally, some methods aim to search the minimal subset of topological information within the topological space of the input graph that can alter the prediction~\citep{wellawatte2022model,chhablani2024game,huang2023global}.
While optimization-based methods are effective, their computational overhead is prohibitive, particularly for large-scale graph data tasks, which restricts their applicability in real-world scenarios.

\textit{Learning-based}.
Learning-based methods employ generative models to capture the feature distribution of input graphs and reconstruct counterfactual graphs that closely resemble the original. Given their ability to handle complex graph data, these methods offer excellent scalability and generalization. 
Consequently, numerous learning-based counterfactual explainers have been proposed.
For instance, \citet{bajaj2021robust} proposed an unsupervised learning method to capture the decision boundaries of each category, thereby fitting a common decision logic for the target model across categories. By learning this decision logic, the explainer identifies subsets of edges that influence the prediction. These edge subsets are then removed from the input graph to generate counterfactual explanations.
\citet{ma2022clear} utilized a graph variational auto-encoder to encode the input graph representation and reconstruct the graph representation into a counterfactual explainable graph.

\section{Self-explanatory Graph Learning}
Unlike post-hoc explainability methods, self-explanatory graph learning methods integrate a self-explanatory module directly within the graph learning model. 
The self-explanatory module actively participates in the model training process and autonomously generates explanations during inference.

\begin{figure}[ht]
	\centering
	\includegraphics[width=0.9\linewidth]{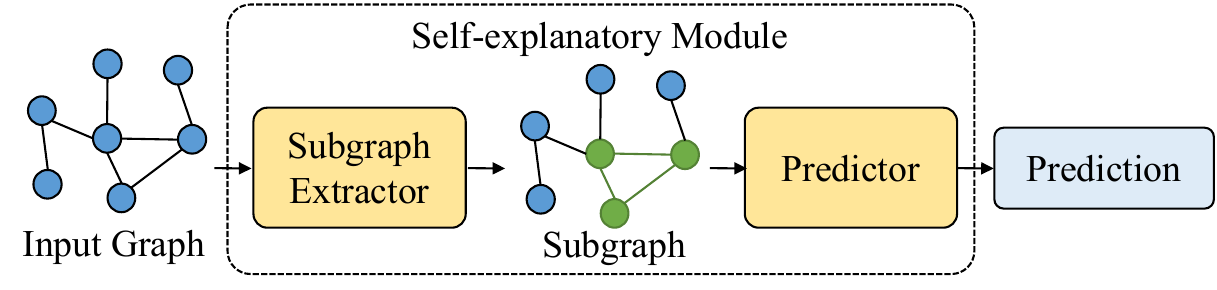}
	\caption{Self-explanatory Graph Learning.}
	\label{Self-explanatory Graph Learning.}
\end{figure}

As illustrated in Figure~\ref{Self-explanatory Graph Learning.}, self-explanatory graph learning proceeds in two stages. First, the self-explanatory module identifies and extracts key subgraphs within the input samples that are most influential to the model prediction. Second, the predictor utilizes these key subgraphs as input to produce the final prediction. These extracted subgraphs serve as intrinsic explanations for the model decision.
Based on the methods of subgraph extraction, existing approaches for self-explanatory graph learning can be categorized into two types: information extraction and structural extraction.

\subsection{Information Extraction}
\begin{figure}[ht]
	\centering
	\includegraphics[width=0.9\linewidth]{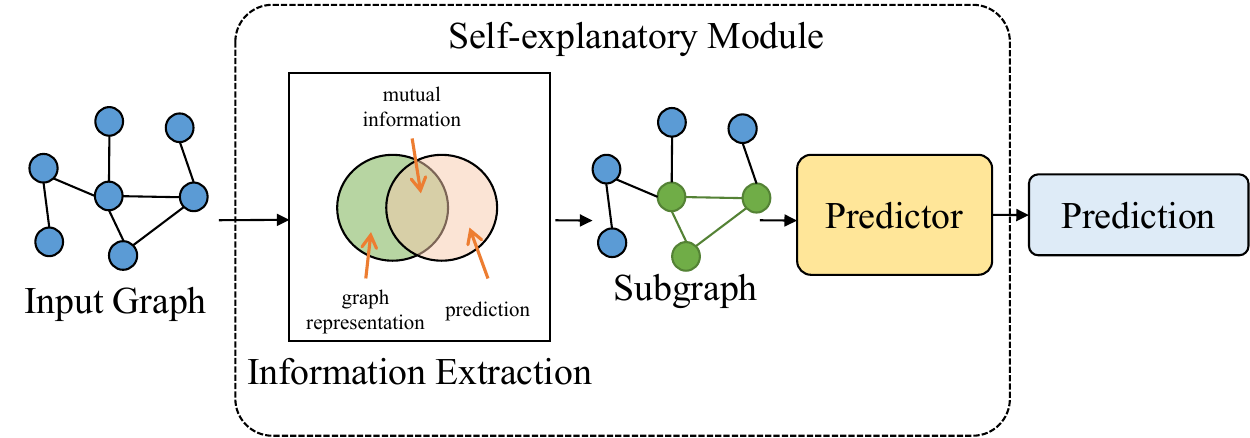}
	\caption{Information Extraction Based Self-explanatory Graph Learning.}
	\label{Information Extraction Based Self-explanatory Graph Learning.}
\end{figure}
As shown in Figure~\ref{Information Extraction Based Self-explanatory Graph Learning.}, information extraction approaches typically use mutual information as a constraint to guide subgraph extraction. 
Specifically, self-explanatory methods based on information extraction aim to maximize the mutual information between subgraphs and the final prediction while minimizing it between subgraphs and the input graphs. 
These methods encourage subgraphs to discard irrelevant features from the original graph, retaining only those key features that significantly impact prediction. 
For example, \citet{DBLP:conf/icml/MiaoLL22} proposed a method namely GSAT (Graph Stochastic Attention) that uses a stochastic attention mechanism to calculate the variational upper bound for $I(G,G_s)$, where $G_s$ is an informative subgraph from the input graph $G$. Similar to GSAT, LRI (Learnable Randomness Injection)  ~\citep{DBLP:conf/iclr/0001LLL23} uses both Bernoulli and Gaussian distribution as the prior distribution. LRI separately calculates the existence and location importance of points to study how perturbing the location of the point will affect the predictions. \citet{DBLP:conf/iclr/YuXRBHH21} proposed a bi-level optimization method, which uses the Donsker-Vardhan KL representation to estimate $I(G,G_s)$ in the inner loop and optimizes the GIB (Graph Information Bottleneck) objective in the outer loop. To improve the efficiency and stability of this bi-level training process, \citet{DBLP:conf/cvpr/Yu0H22} proposed another method named VGIB (Variational Graph Information Bottleneck). VGIB uses a different compression technique by injecting noise into the node representations via a learned probability for each node, and the higher classification loss stands for the informative substructure injected with noise. 
Thus, the informative substructure is less likely to be injected with noise compared to label-irrelevant substructures.
\citet{10.1145/3589334.3645674} proposed a self-explanatory model based on graph distribution matching, which reconstructs a new graph by matching the distributions of the original graph and the explanatory subgraph in its feature space during model training to capture the most informative subgraph.
\citet{NEURIPS2023_f224f056} proposed a prototype-based graph information bottleneck that uses prototype learning to capture key subgraphs in the input graph, which significantly influence the prediction, thereby enabling the prototypes to encapsulate richer key substructures.

\subsection{Structure Extraction}
\begin{figure}[ht]
	\centering
	\includegraphics[width=0.9\linewidth]{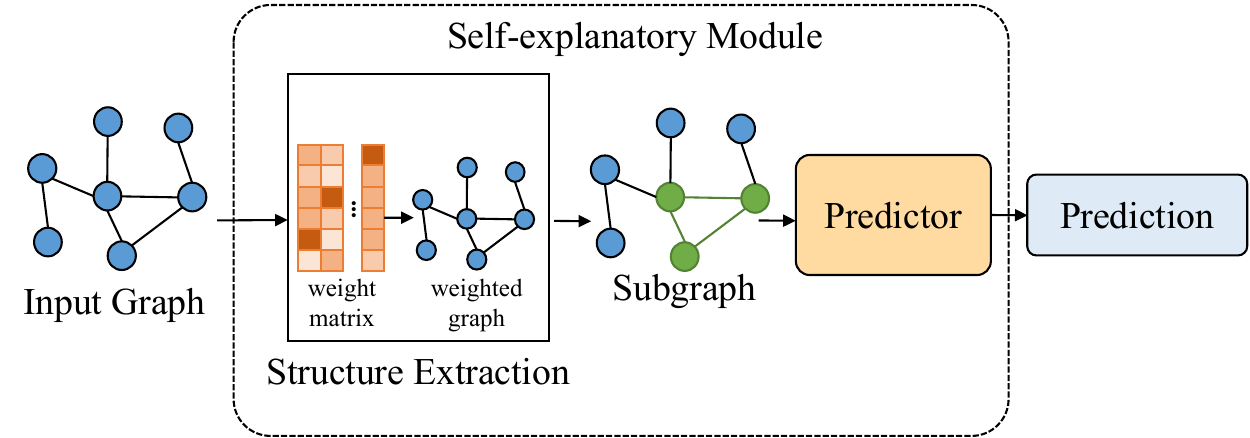}
	\caption{Structure Extraction Based Self-explanatory Graph Learning.}
	\label{Structure Extraction Based Self-explanatory Graph Learning.}
\end{figure}
Unlike information extraction, structure extraction focuses on constraining the topology of the graph instead of mutual information. 
As shown in Figure~\ref{Structure Extraction Based Self-explanatory Graph Learning.}, structure extraction-based self-explanatory approaches directly constrain the topology in the input graph to identify key subgraphs that maximize the accuracy of the predictions made by the graph learning model. For example, 
\citet{DBLP:conf/iclr/WuWZ0C22} proposed to generate explanations by dividing the input into two parts: a causal component $(C)$ and a non-causal component $(S)$. The method is trained to minimize classification loss while ensuring that the prediction $Y$ is conditionally independent of $S$ given $C$. This is accomplished by constructing several interventional distributions during training, where the features that remain stable across these interventions are identified as causal. The process begins by selecting the top-$k$ edges from the input graph to form the causal portion, leaving the rest as non-causal. A distribution intervention module then perturbs the data to help uncover invariant causal patterns. Lastly, two classifiers are applied to produce a combined prediction based on both parts.
\citet{DBLP:conf/aaai/ZhangLWLL22} proposed a method, namely ProtoGNN (Prototype Graph Neural Network), that combines prototype learning~\citep{DBLP:journals/ijmi/SchmidtMBPG01} with GNNs. Instead of relying on post-hoc explanation techniques, ProtoGNN makes predictions by comparing input graphs to a set of learned prototypes in the latent space, and allows explanations to emerge directly from the case-based reasoning process. 
\citet{DBLP:conf/cikm/DaiW21} introduced a method that identifies the $k$ nearest labeled nodes based on both structural and feature similarities. These nearest neighbors are leveraged not only for making predictions but also for producing explanations. An explanation subgraph is constructed by selecting nodes whose importance exceeds a predefined threshold.
\citet{DBLP:conf/aaai/FengY0T22} proposed KER-GNN (Kernel Graph Neural Networks), a framework that incorporates graph kernels into the GNN message-passing mechanism. In this approach, node representations are updated by measuring the similarity between each node’s local subgraph and a set of trainable hidden graph filters. These learnable filters capture key structural patterns within the data, and the resulting node embeddings can be further used to identify significant substructures for interpretation.

%% file: chapters/7-rgl.tex

\par Graph Learning has been increasingly applied in sensitive domains, making responsible usage crucial. 
Protecting personal data is vital for building trust and ensuring the well-being of vulnerable individuals and communities~\citep{zhang2024trustworthy}. 
For instance, privacy-preserving graph learning must guarantee that confidential information, including model parameters and graph structures, remains secure. 
Therefore, how to simultaneously provide accurate individual-level information prediction and user privacy protection is a research problem of great value.
A practical example would be a medical record system with a graph illustrating the social connections among patients affected by COVID-19~\citep{fu2024privacy}. 
In such cases, it is imperative to protect patients' sensitive personal data while training graph learning models on these graphs.

\par As an extension of neural networks on graph-structured data, GNNs are also vulnerable to adversarial attacks~\citep{tang2025rmtrans}. 
Due to their unique message-passing mechanism and graph structure, GNNs can be negatively affected by adversarial perturbations on both graph structures and node attributes.
The vulnerability of GNNs has raised tremendous concerns about their adoption in safety-critical domains such as credit estimation and healthcare. 
For instance, fraudsters could create several transactions with deliberately chosen high-credit users to escape GNN-based fraud detectors, which could cause significant losses to individuals and institutions. 
Hence, developing robust GNNs is another important aspect of trustworthiness, and many efforts have been taken. 
This area is mature enough that several comprehensive surveys about adversarial attacks and defenses on graphs already exist. 
In this section, we provide a brief overview of adversarial learning on graphs but focus more on methods in emerging directions such as scalable attacks, graph backdoor attacks, and recent defense methods.

\par Fairness in graph learning refers to the elimination of any bias or favouritism toward specific individuals or groups in model predictions~\citep{dong2023fairness,luo2025fairgp}.
Therefore, both individual fairness and group fairness need to be taken into account. To this end, different types of datasets have been proposed for studying different types of fairness.

As an extension of deep neural networks to the graph domain, GNNs are typically data-driven and aimed at learning mappings from graph-based data. 
However, the learning process, particularly mechanisms like message-passing, can inadvertently reinforce existing biases in the data, potentially leading to unfair predictions influenced by sensitive attributes such as race or gender~\citep{li2024rethinking}.
Therefore, how to tackle unfairness issues in graph learning algorithms naturally becomes another crucial responsible problem. 
In this survey, we review responsible graph learning with a particular focus on Privacy-Preserving Graph Learning and Fairness in Graph Learning.

\section{Privacy-Preserving Graph Learning}

The growing use of graph learning in areas involving sensitive data has intensified privacy concerns~\citep{guan2024graph}. 
For example, attackers can perturb a graph’s data to undermine a model’s effectiveness, or they can steal the model’s data and/or parameters, thus threatening the privacy of the model.
Given the potential privacy risks linked to graph learning systems, a variety of privacy-enhancing techniques have been developed to address these challenges. 
This section will explore some of the most widely adopted approaches aimed at safeguarding privacy in graph learning.

\subsection{Differential Privacy Based Methods}
Differential Privacy (DP)~\citep{wei2024poincare}, a well-established technique for privacy-preserving, ensures that an attacker cannot confidently infer private information about any specific training data from a released model. 
This guarantee of privacy is essential when developing privacy-preserving applications for graph learning.
For example, consider a social smartphone application where the server stores information about users' social interactions. 
While the server may wish to leverage private user data (e.g., usage logs) to enhance graph learning models for improved services (e.g., recommendation systems), it must ensure that raw user data remains inaccessible due to privacy concerns.
Specifically, randomization is added to the data so that an attacker cannot infer private information from any differences in the query result.
The formal definition of differential privacy is as follows~\citep{zhang2024a}:
\begin{equation}
    \mathbb{P}[R(\text{D}) \in \textbf{e}] \leq e^\epsilon \mathbb{P}[R(\text{D}') \in \textbf{e}] + \delta,
\end{equation}
where $R$ is a randomization algorithm, D and D' are two different datasets, $\textbf{e}$ is all events, $\epsilon$ is the privacy budget, and $\delta$ is a perturbation.
This formula holds that a randomized algorithm acting over $R$ on two neighboring data sets D and D' should yield an output event $\textbf{e}$ with about the same probability.
To use DP algorithms for GNN data, the properties of the neighboring datasets need to be formally defined~\citep{mueller2022differentially}. 
These neighboring datasets are divided into three categories according to what is being protected: removing or adding a node and its adjacent edges (node-level DP), removing or adding an edge (edge-level DP), and removing or adding an entire graph (graph-level DP).

\citet{xiang2024preserving} presented a solution that focuses on node-level privacy through a two-component protocol: 1) Heter-Poisson, a specialized node sampling strategy that generates sub-graphs with desired properties, and 2) a randomization routine utilizing symmetric multivariate Laplace noise, offering a robust privacy guarantee. Additionally, the authors critiqued a prior approach~\citep{sajadmanesh2023gap} that utilizes differentially private node embeddings, highlighting fundamental privacy flaws and proving the impossibility of achieving both strong privacy and acceptable utility with such embeddings, thus underscoring intrinsic utility barriers in enforcing differential privacy.

To mitigate these privacy concerns, DP protocols are typically used. 
Standard DP approaches (e.g. DP-SGD~\citep{abadi2016deep}) are not suitable for GNNs due to potential privacy leakage from node label predictions based on neighboring attributes, and varying privacy requirements for node attributes and graph topology. 
To overcome these limitations, Graph Differential Privacy (GDP)~\citep{chien2023differentially} is designed specifically for graph learning, which ensures private model parameters and predictions while offering a unified notion of graph dataset adjacency for analyzing topology privacy. 
The proposed Differentially Private Decoupled Graph Convolutions (DPDGCs) provides a more flexible and efficient alternative to traditional graph convolutions, significantly improving privacy-utility trade-offs in extensive experiments across seven node classification benchmarks and synthetic datasets compared to existing DP-GNNs.

\subsection{Federated Learning Based Methods}

Federated Learning (FL) has become a widely adopted framework that allows decentralized entities, such as mobile devices, to collaboratively train machine learning models without exposing their raw data~\citep{pei2024a}. 
In FL, the data owned by these individuals is treated as private, and it is processed and trained locally, without being shared externally. 
To ensure privacy during training, a central server periodically collects and aggregates parameters (such as gradients or model weights) from each individual client until the model reaches satisfactory performance. 
Recently, there has been a growing interest in applying FL to the context of graph learning, with several studies exploring its potential.

FL-based graph learning safeguards the privacy of individuals by transmitting model parameters rather than raw data between participants and the central server during training~\citep{fu2024federated}. 
This approach ensures that private data remains local and inaccessible to others.
Specifically, each individual trains a graph learning model on their own data and subsequently uploads their local model to a server, which performs an aggregation function (e.g., FedAvg) to build a global GNN model. 
Notably, the aggregation process utilizes GNN parameters inductively, meaning the model remains independent of its underlying structure. 
Consequently, topological information about local graph data is not required, allowing it to remain private.
However, unlike vanilla FL tasks (e.g. relies on Non-i.i.d data), FL-based graph learning presents unique challenges due to the nature of graph data. 
These challenges often necessitate carefully designed mechanisms during training and implementation, as outlined below:

i) \textbf{Non-i.i.d. individual data}: FL-based graph learning relies on stochastic gradient descent (SGD), making the training process vulnerable to non-i.i.d. data distributions~\citep{lu2024federated,zhang2025fairness}. 
In practice, the heterogeneity in graph structures and node features across individuals introduces significant challenges. 
The non-i.i.d. nature of the local graph data (i.e., divergent graph structure and feature distributions) poses a major issue for FL with graphs. 
Consequently, many FL methods tailored for GNNs aim to mitigate the effects of non-i.i.d. data on training performance. 

Federated learning has become a vital approach for training machine learning models across various domains, including graph-level tasks like graph classification, where graphs are treated as distinct data samples stored in separate local systems. 
Real-world graphs from multiple domains share statistically significant properties compared to random graphs, while different sets of graphs, even within the same domain, exhibit non-i.i.d. characteristics in terms of graph structures and node features. 
To address these challenges, SpreadFGL~\citep{zhong2024spreadfgl} is a novel framework that addresses key challenges in multi-client training, such as missing inter-client topology information and high training costs.
SpreadFGL uses an adaptive graph imputation generator to discover potential relationships between clients without data sharing and incorporates a negative sampling mechanism to enhance task performance. 
The framework also supports distributed training for load balancing, enabling faster model convergence.

ii) \textbf{Graph isolation}: In horizontal intra-graph FL, local graph data can be viewed as isolated subgraphs from a latent global graph~\citep{yang2024distributed}. 
Since representation learning in graph learning relies on message passing between connected nodes, the inability to access private information from other subgraphs negatively impacts the accuracy. 
Current FL methods designed for graph learning address this issue by transmitting high-level domain representations rather than raw connections, or by employing generators to produce missing neighbors across distributed subgraphs.

FedSpray~\citep{fu2024federated} learns local class-wise structure proxies in the latent space and aligns them to form global structure proxies at the server, aiming to provide reliable and unbiased neighboring information for improved node classification.
FGL focuses on developing graph learning models across graph data distributed among multiple data owners, applicable in fields like social recommendation and financial fraud detection. 
Similar to traditional FL, FGL faces data heterogeneity issues, where label distributions can vary significantly across clients, leading to divergent local objectives that hinder convergence, particularly in node classification tasks. 
Additionally, FGL encounters challenges related to biased neighboring information for nodes belonging to minority classes, which affects the ability of GNNs to learn effective embeddings. 
FedSpray addresses these challenges. 

\subsection{Adversarial Privacy Preservation}
To defend against attacks that attempt to infer sensitive attributes or links, adversarial learning is commonly employed in privacy-preserving graph learning. 
The main idea behind this approach is the introduction of an adversary, whose goal is to deduce sensitive information from the node representations produced by the encoder~\citep{liao2024value}. 
Meanwhile, the encoder is designed to generate node representations so that the adversary cannot infer sensitive attributes successfully. 
This process minimizes the shared information between the learned representations and the sensitive attributes, protecting against privacy leaks.

The balance between maintaining the utility of the learned model and preventing sensitive information leakage is achieved by minimizing a utility loss function, which ensures the graph learning performance in tasks like classification or reconstruction while maximizing the adversarial loss function to reduce the adversary's ability to predict sensitive attributes~\citep{zhao2023adversarial}.
Adversarial privacy preservation is first introduced to address the attribute inference problem, combining link prediction and node attribute prediction loss functions to maintain utility. 
In this approach, node attributes are predicted based on the learned representations, and losses from both link prediction and attribute reconstruction are combined to ensure the model is both useful and private. 

\citet{gosch2023adversarial} addressed fundamental theoretical and practical limitations of previous approaches in the graph learning setting. 
It reveals that flexible GNNs, which utilize learnable graph diffusion, can adapt to adversarial perturbations while maintaining an interpretable message-passing scheme. 
Additionally, the authors introduced a novel attack for structure perturbations that targets multiple nodes simultaneously and accommodates both global (graph-level) and local (node-level) constraints, ultimately demonstrating that adversarial training is a leading defence against such perturbations.

\section{Robustness of Graph Learning}
\par Graph learning models are increasingly vulnerable to adversarial attacks, making the development of robust graph learning a critical area of research. 
These defense strategies are generally categorized into three main approaches: Adversarial Training, Graph Denoising, and Certifiable Robustness. 
The following sections introduce key methods within each category, along with other notable defense techniques that fall outside these classifications.

\subsection{Adversarial Training}
\par This approach involves fortifying a graph learning model by exposing it to adversarial samples during the training process, a practice inspired by its success in computer vision. 
The core idea is to improve the model's ability to resist attacks by explicitly incorporating adversarial perturbations into the training loop. 
This can be achieved in two main ways:

\par Min-Max Optimization: 
A graph learning is trained to simultaneously minimize its loss on both clean and adversarially perturbed data.
This process can be formally expressed by the following equation:
\begin{equation}
\min _\theta \max _{\Delta_{\mathbf{A}} \in \mathcal{P}_{\mathbf{A}}, \Delta_{\mathbf{x}} \in \mathcal{P}_{\mathbf{X}}} \mathcal{L}_{\text {train }}\left(f_\theta\left(\mathbf{A}+\Delta_{\mathbf{A}}, \mathbf{X}+\Delta_{\mathbf{X}}\right)\right),
\end{equation}
where $\mathcal{L}_{\text{train}}$ is the classification loss on the labeled nodes.
$\Delta_{\mathbf{A}}$ and $\Delta_{\mathbf{X}}$ represent the perturbations on the topology structure and node attributes, respectively.
$\mathcal{P}_\mathbf{A}$ and $\mathcal{P}_\mathbf{X}$ denote the allowable perturbations within the attack budget.

\par This approach forces the model to learn a more robust decision boundary that is less susceptible to small, malicious changes in the input. 
For example, method like PerturbEmbedding adds an adversarial term to the loss function, encouraging the model to become more resistant to both structural and feature-based attacks.
\citet{yang2025unifying} proposed PerturbEmbedding that investigates the vulnerability of graph learning models to adversarial attacks that target both node features and graph structure. 
While acknowledging that existing adversarial training methods have shown promise in improving model robustness, they note that the application of these methods is often limited to specific datasets and model types. 
To address this limitation, they introduce a novel method, PerturbEmbedding, which applies adversarial perturbations directly to the hidden embeddings of graph learning models. 
This approach provides a unified and effective framework that enhances both the robustness and generalization of a wide range of graph learning models. 
By perturbing the hidden embeddings, the method forces the model to be resilient not just against input-level attacks, but also against malicious changes at intermediate layers, thereby creating a more robust and adaptable model.

\par Robust Pre-training: 
Some methods focus on pre-training the model on a large number of synthetically generated adversarial graphs before fine-tuning it on the clean graph. 
This allows the model to learn general robust features from a diverse set of attack scenarios.
For example, a recent study proposes LLM4RGNN, which leverages the inference capabilities of LLMs to enhance a graph learning model's robustness~\citep{zhang2025can}. 
This framework distills the knowledge of a powerful model like GPT-4 into a local LLM to identify and remove malicious edges from the graph structure. 
It also uses an additional language model to predict and recover important missing edges, effectively "denoising" the graph to make it more resilient to topology-based attacks.

\subsection{Graph Denoising}
\par The goal of graph denoising is to purify the input graph before or during the graph learning model's forward pass, effectively neutralizing adversarial perturbations. 
This approach is based on the assumption that malicious perturbations are a form of noise that can be filtered out. 
Denoising methods typically focus on cleaning either the graph structure or the node features.

\par Structure Denoising: These methods aim to identify and remove suspicious or harmful links in the graph.
The process can be formulated as a joint optimization problem to learn a clean adjacency matrix, $\mathbf{S}$, as shown below:
\begin{equation}
\min _{\theta, \mathbf{S}} \mathcal{L}_{\text {train }}\left(f_\theta(\mathbf{X}, \mathbf{S})\right)+\alpha\|\mathbf{A}-\mathbf{S}\|_F^2+\beta\|\mathbf{S}\|_*+\lambda \operatorname{tr}\left(\mathbf{X}^T \hat{\mathbf{L}} \mathbf{X}\right),
\end{equation}
where $\mathcal{L}_{\text{train}}(f_\theta(\mathbf{X},\mathbf{S}))$ is the classification loss.
The term $|\mathbf{S}\|_*$ is the nuclear norm of the learned adjacency matrix $\mathbf{S}$, which encourages it to be low-rank.
The final term, $\operatorname{tr}\left(\mathbf{X}^T \hat{\mathbf{L}} \mathbf{X}\right)$, encourages the learned adjacency matrix to form links between nodes with similar features, where $\hat{\mathbf{L}}$ is the Laplacian matrix of $\mathbf{S}$.

\par For instance, in real-world graphs, both structural noise (e.g., noisy edges) and a limited number of labeled nodes can significantly impair the performance of graph learning models. 
A recent framework proposed by \citet{dai2022towards} addresses this by using noisy edges as a form of supervision to learn a denoised and dense graph. 
This process effectively eliminates or down-weights noisy edges while simultaneously creating new, more meaningful connections that facilitate the message-passing mechanism. 
The framework then uses these newly generated edges to regularize predictions for unlabeled nodes, improving the overall training and robustness of the graph learning model, even when faced with significant data imperfections.

\par Feature Denoising: 
This involves applying a denoising operation to the node features to remove adversarial noise. 
For instance, some approaches use a pre-processing step to project the node features onto a low-dimensional subspace, where the adversarial perturbations are assumed to be less significant.
A novel framework, DiffSP (Diffusion-based Structure Purification), offers a prior-free approach to this problem~\citep{luo2025robust}. 
Instead of relying on assumptions about the attack, DiffSP uses a graph diffusion model to learn the inherent distribution of clean graphs. 
It then purifies a perturbed graph through a forward diffusion process (adding noise) and a reverse denoising process, where it removes adversarial noise under the guidance of the learned patterns.
To avoid losing important information, DiffSP uses an LID-driven non-isotropic diffusion mechanism to selectively inject noise. 
The framework also promotes semantic alignment between the clean and purified graphs via a graph transfer entropy guided denoising mechanism, ultimately achieving robust graph learning without needing prior knowledge of the attack.

\subsection{Certifiable Robustness}
\par Unlike the first two categories, which provide empirical defense, certifiable robustness offers a mathematical guarantee that a graph learning model's prediction will remain unchanged within a certain radius of perturbation. 
This provides a concrete measure of the model's security.
For a given node $v \in \mathcal{V}$ with label $y$, its robustness can be certified by solving the following optimization problem:
\begin{equation}
m(v ; \theta)=\min _{\hat{\mathcal{G}} \in \Phi(\mathcal{G})} \max _{i \neq y} f_\theta(\hat{\mathcal{G}})_{v y}-f_\theta(\hat{\mathcal{G}})_{v i},
\end{equation}
where $f_\theta(\hat{\mathcal{G}})_{v i}$ is the predicted logit of node $v$ in class $i$, and $\Phi(\mathcal{G})$ includes all allowable perturbed versions of the graph.
A model is certifiably robust for node $v$ if $m(v)>0$.
Key methods in this area include:

\par Randomized Smoothing: 
This technique, adapted from computer vision, involves adding random noise to the input graph and averaging the model's predictions over many noisy samples. 
By doing so, it provides a provable robustness guarantee for the model's output, ensuring that the prediction remains consistent as long as the perturbation is within a specified bound.
Building on this idea, the node-aware bi-smoothing framework extends this technique to address a new challenge: certified robustness against graph injection attacks (GIAs)~\citep{lai2024node}. 
This framework is the first to provide a verifiable defense for general node classification tasks against GIAs, making it applicable for both evasion and poisoning attacks. 
The method is also model-agnostic, meaning it can be used with various graph learning architectures. 
The authors of this work provide a rigorous theoretical analysis to establish the certifiable conditions of their smoothing scheme and demonstrate its practical effectiveness as a defense against real-world attacks.

\par Interval Bound Propagation (IBP): 
This approach, inspired by IBP methods for neural networks, aims to compute a tight bound on a graph learning model's output for all possible inputs within a given perturbation set. 
By ensuring that this output range does not cross a decision boundary, it can certify the model's robustness against a wide range of attacks.
A recent study addresses this challenge by applying a dual backpropagation network to deep graph learning~\citep{zeng2025robustness}. 
The authors propose a novel method for calculating intermediate activation bounds based on a linear approximation, which effectively enhances verification accuracy and scalability. 
Their approach is shown to be capable of verifying the robustness of deep graph learning with up to $32$ layers, demonstrating a significant improvement in the maximum perturbation value that correctly classified nodes can withstand.

\par In addition to these three main categories, other notable defense methods are emerging, such as those that leverage graph diffusion to disperse adversarial information or knowledge distillation to transfer robustness from a large, robust model to a smaller one. 
The continuous evolution of attack and defense strategies highlights the dynamic and critical nature of building trustworthy and robust graph learning systems.

\section{Fairness in Graph Learning}
\begin{figure*}[htbp]
    \centering
    \includegraphics[width=1\linewidth]{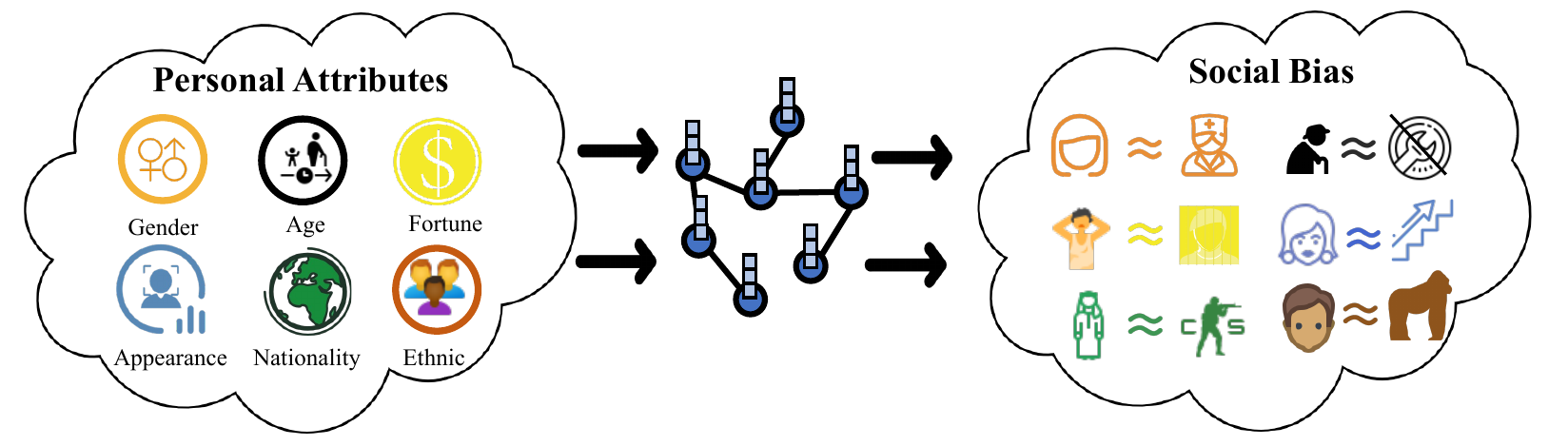}
    \caption{Reinforcing social stereotypes.}
    \label{RSS}
\end{figure*}

\begin{figure*}[htbp]
    \centering
    \includegraphics[width=1\linewidth]{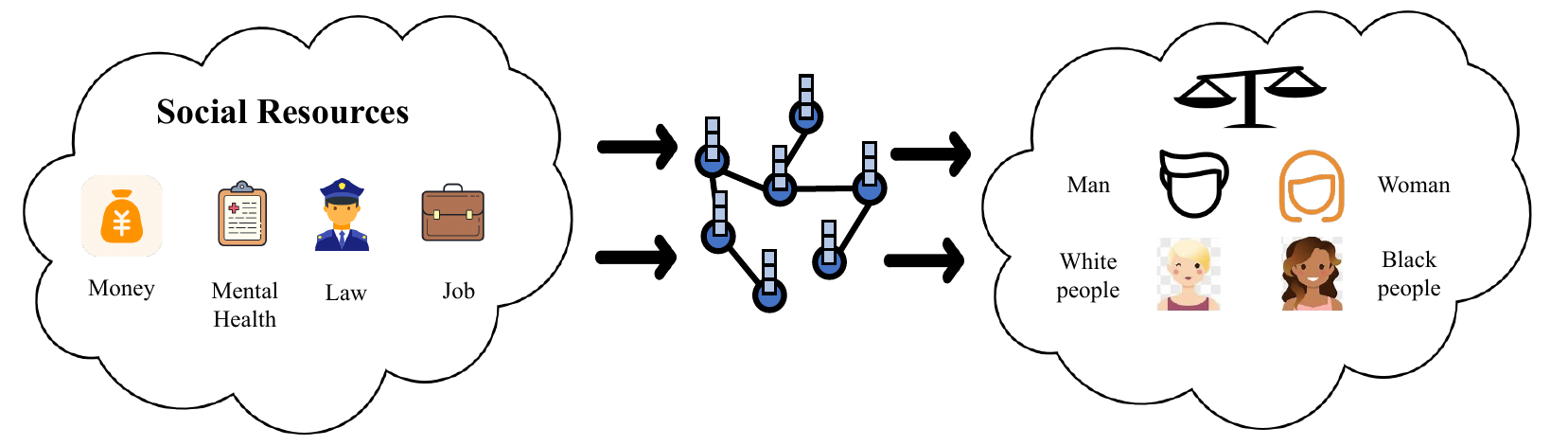}
    \caption{Causing inequitable distribution of resources.}
    \label{CIDR}
\end{figure*}

Despite the success of various graph learning algorithms, many of them do not adequately address fairness considerations. 
As a result, these algorithms can produce discriminatory outcomes against specific populations when applied in human-centered contexts~\citep{luo2024algorithmic}. 
For instance, a job recommendation system based on social networks may inadvertently offer fewer opportunities to individuals of a particular gender or to those from underrepresented ethnic groups, as shown in Figures~\ref{RSS} and~\ref{CIDR}. 
Given the widespread application of graph mining algorithms, similar discriminatory practices may emerge in other high-stakes scenarios, such as disaster response, criminal justice, and loan approvals. 
In these situations, critical decisions that can significantly impact individuals’ lives are often made. 
Therefore, addressing fairness issues in graph learning algorithms is an essential challenge.

\subsection{Group Fairness}
In many high-stakes applications, such as loan approval systems, certain features—like race and gender—are legally protected to prevent misuse.
Furthermore, in contexts like online social networking, users are often reluctant to share certain features, such as occupation and age. 
These features are classified as protected or sensitive attributes. 
Based on these attributes, populations can be categorized into different demographic subgroups, referred to as sensitive subgroups. 
Group fairness is then defined in relation to these sensitive subgroups, stipulating that algorithms should not produce discriminatory predictions or decisions against individuals belonging to any specific sensitive subgroup. \citet{luo2025fairgp} introduced a definition of fairness for a binary label $y \in {0,1}$ and a binary sensitive feature $s \in {0,1}$, where $\hat{y} \in {0,1}$ represents the predicted class label.

\textbf{Statistical Parity}~\citep{dwork2012fairness} (also known as Demographic Parity or Independence) requires that predictions be independent of the sensitive feature $s$. 
This can be formally expressed as: 
\begin{equation} 
    \mathbb{P}(\hat{y}|s=0) = \mathbb{P}(\hat{y}|s=1). 
\end{equation}

When both the predicted labels and sensitive features are binary, the degree of statistical parity is quantified by $\Delta_{\text{SP}}$, defined as: 
\begin{equation} 
    \Delta_{\text{SP}} = |\mathbb{P}(\hat{y}=1|s=0) - \mathbb{P}(\hat{y}=1|s=1)|. 
\end{equation}

$\Delta_{\text{SP}}$ measures the difference in acceptance rates between the two sensitive subgroups.

\textbf{Equal Opportunity}~\citep{hardt2016equality} demands that individuals in the positive class (i.e., those for whom $y = 1$) have an equal chance of receiving a positive prediction, regardless of their sensitive feature. 
In other words, for individuals with a positive ground truth label, sensitive features should not influence positive predictions. 
This can be formulated as: 
\begin{equation} 
    \mathbb{P}(\hat{y} = 1|y = 1, s = 0) = \mathbb{P}(\hat{y} = 1|y = 1, s = 1). 
\end{equation}

Fairness-aware algorithms are designed to prevent biased outcomes based on an individual's membership in a sensitive subgroup~\citep{hardt2016equality}. 
The degree to which predictions deviate from equal opportunity is quantified by $\Delta_{\text{EO}}$, defined as: 
\begin{equation} 
    \Delta_{\text{EO}} = |\mathbb{P}(\hat{y} = 1|y = 1, s = 0) \ - \mathbb{P}(\hat{y} = 1|y = 1, s = 1)|.
\end{equation}

Both $\Delta_\text{SP}$ and $\Delta_\text{EO}$ are evaluated on the test set to assess the fairness of the model~\citep{hardt2016equality}.

Graph learning can produce biased predictions based on protected sensitive attributes, such as skin color and gender, due to historical biases in the training data. 
This discrimination can be exacerbated by the graph structures and the message-passing mechanisms inherent in graph learning, particularly in high-stakes applications like crime rate prediction. 
While extensive research has focused on fair classification with independently and identically distributed (i.i.d.) data, approaches for non-i.i.d. scenarios remain limited, especially since sensitive attributes are often scarce in real-world graphs. 
To tackle this issue, FairGNN~\citep{dai2023learning} has been proposed to mitigate bias while maintaining node classification accuracy by leveraging limited sensitive information, and it employs local differential privacy to protect sensitive attributes. 
An extension, NT-FairGNN, further addresses the challenges of achieving fairness and privacy simultaneously, with theoretical analysis and empirical results demonstrating their effectiveness.

\begin{figure}[H]
    \centering
    \includegraphics[width=1\linewidth]{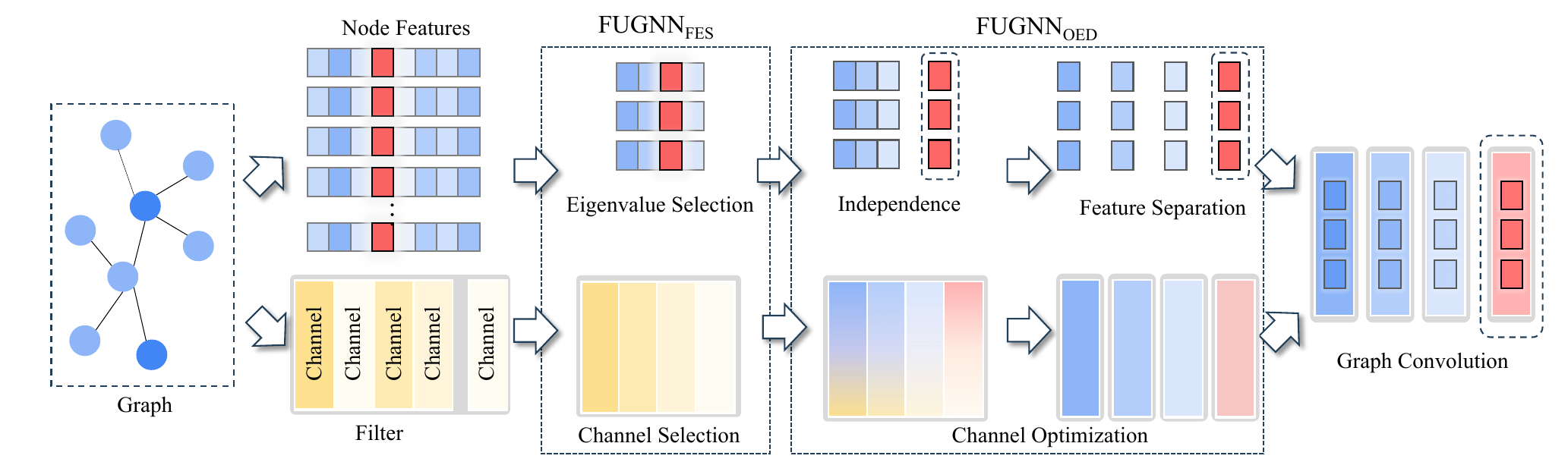}
    \caption{The framework of FUGNN. The model applies spectral truncation via eigenvalue selection and eigenvector distribution
optimization with fairness considerations. The top and bottom illustrate the two stages from the perspective of feature-level
representations and feature channels, respectively.}
    \label{fig:fugnn}
\end{figure}

In addition, fairness-aware GNNs often navigate a complex trade-off between achieving fairness and maintaining utility. 
FUGNN~\citep{luo2024fugnn} re-evaluates fairness using spectral graph theory, examining the relationship between sensitive features and the spectrum in GNNs, as shown in Figure~\ref{fig:fugnn}. 
Through theoretical analysis, it identifies how directional similarity in eigenvectors associated with the largest eigenvalue can diminish the influence of sensitive features. 
Building on these insights, the proposed FUGNN harmonizes fairness and utility by optimizing eigenvector distribution and truncating the spectrum during encoding, ensuring that convolution has a reduced impact on sensitive features without significantly sacrificing performance. 

In addition, Graph Transformers (GTs) typically neglect fairness considerations, leading to biased outcomes for certain sensitive subgroups. 
Due to their lack of message-passing mechanisms, traditional fairness-aware graph learning methods are not applicable to GTs. 
To address this, FairGT~\citep{luo2024fairgt}, a Fairness-aware Graph Transformer, is designed to alleviate fairness issues in GTs. 
FairGT utilizes a careful structural feature selection strategy and a multi-hop node feature integration method to ensure the independence of sensitive features, thereby enhancing fairness. 
The effectiveness of its fair structural topology encoding, which incorporates adjacency matrix eigenvector selection and multi-hop integration, is theoretically validated. 

\subsection{Individual Fairness}
Compared to group fairness, Individual Fairness does not take sensitive features into account; instead, it emphasizes fairness at the individual level, such as for each node in graph data. 
The core principle of individual fairness is that similar individuals should be treated similarly, making it a more granular fairness notion than group fairness. 
However, there are currently only a limited number of graph mining algorithms that explicitly address individual fairness.
Individual fairness is based on the understanding that similar individuals should be treated similarly.

\textbf{Fairness Through Awareness}~\citep{dwork2012fairness} mandates that any two individuals who are similar should receive similar algorithmic outcomes from an algorithm.
Let $u,v \in \mathcal{X}$ be two data points in dataset $\mathcal{X}$, and let $f(\cdot)$ denote a mapping function.
The formulation of fairness through awareness can be expressed as follows:
\begin{equation}
    \text{If} \quad sim(u,v) \geq \epsilon, \text{then} \quad |f(u), f(v)| \leq \delta,
\end{equation}
where $sim(\cdot, \cdot)$ is a similarity measure, $\epsilon$ is a threshold for similarity, and $\delta$ is the allowable difference in outcomes for similar individuals.

Algorithmic bias and fairness in graph mining have received limited attention, primarily focusing on group-based fairness notions. 
\citet{kang2020inform} introduced the first comprehensive study of Individual Fairness in Graph Mining (InFoRM), proposing a generic definition of individual fairness that leads to a quantitative measure of bias in graph mining results. 
It presents three complementary algorithmic frameworks to mitigate individual bias: debiasing the input graph, the mining model, and the mining results, all formulated from an optimization perspective. 
Additionally, they analyzed the impact of incorporating individual fairness on original mining results, developing an upper bound to characterize the associated costs.

REDRESS~\citep{dong2021individual} approaches individual fairness from a ranking perspective by utilizing a similarity matrix $\mathbf{S}$ to define pair-wise similarities among individuals in the input space. 
Each individual is associated with a ranking list $\mathbf{R}1$ that reflects its relative similarity to others, while a corresponding list $\mathbf{R}2$ is generated based on pair-wise individual similarities in the output space. 
Individual fairness is achieved when $\mathbf{R}1$ and $\mathbf{R}2$ match for each individual; however, this criterion is challenging to fulfill. 
As a practical measure, the average top-k similarity between $\mathbf{R}1$ and $\mathbf{R}2$ across all individuals is used to assess individual fairness, employing common ranking similarity metrics such as $NDCG@k$ and $ERR@k$.

\citet{munagala2024individual} addressed classic randomized low-diameter decomposition procedures for planar graphs, which produce connected clusters by grouping nearby nodes with high probability. 
The authors introduced an additional fairness constraint: nodes at comparable distances should have similar separation probabilities. 
They demonstrated that traditional decomposition methods fail to meet this fairness criterion and propose novel algorithms that balance individual fairness with cluster connectivity and optimality. 
Additionally, the authors tied potential improvements in fairness bounds to a key unresolved question in metric embeddings and validated the proposed algorithms on real-world planar networks, including congressional redistricting tasks.

\subsection{Graph Structure Fairness}
Graph structural fairness is a relatively new concept in the graph mining community, distinguishing itself from traditional fairness notions. 
In networked data, connections between nodes can create dependencies that are informative for various graph mining tasks. 
However, low-degree nodes often provide limited information about their dependencies, making it challenging for graph mining algorithms to capture critical information compared to high-degree nodes effectively. 
For instance, the performance of GNNs in tasks like node classification typically varies significantly between high-degree nodes (e.g., a celebrity with many followers) and low-degree nodes (e.g., an average person with few followers). 
Graph structural fairness posits that nodes should achieve similar utility, such as classification accuracy, regardless of their structural characteristics. 

GCNs are crucial in numerous real-world applications, yet they often suffer from performance disparities related to node degrees, resulting in lower predictive accuracy for low-degree nodes. 
\citet{buchanan1976hobbesian} approached the issue of degree-related performance disparity in GCNs through the lens of the Rawlsian difference principle from distributive justice theory, aiming to balance utility between low-degree and high-degree nodes while minimizing task-specific loss.
By analyzing the gradients of weight matrices in GCNs, \citet{kang2022rawlsgcn} identified the root cause of this unfairness and proposed two methods: RawlsGCN-Graph for pre-processing and RawlsGCN-Grad for in-processing, both of which enhance predictive accuracy for low-degree nodes without altering the GCN architecture or adding parameters. 
Extensive experiments on real-world graphs demonstrate that RawlsGCN significantly reduces degree-related bias while maintaining comparable overall performance.

DegFairGNN~\citep{liu2023on} highlights the fairness issues in conventional GNNs that arise from neighborhood structures, in addition to sensitive attributes. 
While previous approaches have addressed bias from node attributes, they often overlook the impact of varying neighborhood structures, especially differences in node degrees, which can lead to biased node representations. 
To address this, DegFairGNN generalizes degree bias as a measure of multi-hop structural differences around nodes.

%% file: chapters/8-et.tex

As graph learning matures, a range of emerging topics are expanding its scope and deepening its integration with cutting-edge developments in AI. Moving beyond traditional tasks like node classification and link prediction, researchers are now exploring more versatile, scalable, and intelligent approaches that push the boundaries of what graph-based models can achieve~\citep{shehzad2024graph,yu2025graph2text}. These emerging directions not only address new technical challenges but also reflect the broader AI agenda of building more generalizable, distributed, and knowledge-driven systems.

This section discusses several key frontiers in graph learning. We begin with graph foundation models, which aim to develop large-scale, pretrained graph encoders capable of generalizing across diverse tasks and domains. We then explore graph reinforcement learning, where graph structures are used to model complex environments and interactions for sequential decision-making. We also discuss federated graph learning, which enables collaborative model training across decentralized graph data sources while preserving privacy. Additionally, we cover advances in learning on knowledge graphs (KGs) and knowledge-infused graph learning, which leverage symbolic information to enhance reasoning and generalization. Finally, we examine the emerging field of quantum graph learning, which explores quantum computing techniques for more efficient and expressive graph representations. These topics represent the next wave of innovation, positioning graph learning as a central pillar in the development of intelligent, explainable, and scalable AI systems.

\section{Graph Foundation Models}

Graph Foundation Models (GFMs) are large-scale pre-trained models specifically designed for processing graph-structured data. These models are generally pre-trained on vast amounts of data across various domains~\citep{xia2022asurvey,lengthiu2023towardsgraph}. Besides, GFMs can adapt to varying demands with minimal fine-tuning or even in few-shot/zero-shot scenarios. There exist many open issues to be addressed, and we discuss them from three distinct perspectives: pre-training, architecture design, and adaptation strategies.

\subsection{Pre-training Strategies}

Pre-training involves training a model on large-scale data in advance. The goal is to create a model with transferable knowledge that can be easily fine-tuned for specific downstream tasks, improving performance even with limited target task data. Current GFMs mainly adopt supervised pre-training and self-supervised pre-training.

\begin{itemize}
	\item \textbf{Supervised pre-training}: In supervised pre-training, models are trained on labeled datasets, often incorporating multi-task learning to optimize performance across various tasks. This approach allows the model to learn representations that are specific to particular tasks, enhancing its effectiveness in downstream applications.
	\item \textbf{Self-supervised pre-training}: Self-supervised learning has become a dominant strategy in graph foundation models, as it allows for training without labeled data. This approach includes techniques like contrastive learning, where the model learns by distinguishing between positive and negative samples, and generative models, where the model learns to generate or reconstruct graph structures.
\end{itemize}

While pre-training has shown promise, several challenges remain that hinder its broader applicability. Below, we highlight key areas that require further exploration and innovation.

\begin{itemize}
	\item \textbf{Task Design}. Current pre-training tasks tend to be tailored to specific domains, which limits the ability of models to perform effectively on diverse types of graph data. The development of more universal pre-training tasks that enable models to learn transferable features applicable across various graph domains remains a critical area of research.
	\item \textbf{Data Efficiency}. Pre-training often demands large, high-quality datasets, which are not always accessible. Enhancing the data efficiency of pre-training models is crucial, particularly in data-scarce environments. This challenge involves optimizing the use of available data, including exploring large-scale datasets, leveraging cross-modal data, and implementing robust data augmentation techniques. Additionally, assessing and ensuring the quality of pre-training data is essential to maximize model performance.
	\item \textbf{Graph Complexity}. The complex nature of graph structures poses troubles for pre-training. Designing effective pre-training methods for various types of graphs, such as dynamic, homogeneous, heterogeneous graphs, or even hypergraphs, is an area that requires further exploration.
\end{itemize}

\subsection{Architecture Design}

GFMs have evolved through a variety of architectural paradigms, each designed to capture different aspects of graph data. From classic message-passing frameworks to more recent innovations like Graph Transformers and LLMs, these architectures have significantly advanced. The existing architectures for GFMs are as follows.

\begin{itemize}
	\item \textbf{Message-Passing Models}: Message-passing models are the most common designs in graph models. These architectures capture local graph structures by iteratively passing and aggregating information between nodes. Classic models in this category include GCN~\citep{KipfW17gcn}, GAT~\citep{velickovic2018graph}, and GraphSAGE~\citep{hamiltonYL17graphsage}. 
    Despite the success, message-passing struggles with expressivity, often failing to capture long-range dependencies due to over-smoothing and limited receptive fields. 
     \item \textbf{Graph Transformer}: Inspired by the success of transformers in natural language processing, Graph Transformers (GTs)~\citep{ying2021transformers,chen2022structure,ma2024polyformer} have been extended to graph-structured data. These models apply global attention mechanisms to capture long-range dependencies in graphs, making them well-suited for handling complex relations. Their token-based graph representations are well-suited for pretraining tasks, although scalability and structural priors remain open challenges.
    \item \textbf{Generative Models}: Generative graph models, such as VAEs~\citep{simonovsky2018graphvae} and GAN-based methods~\citep{wang2018graphgan}, learn latent graph distributions to support tasks like generation, completion, and anomaly detection. Besides, recent diffusion and autoregressive approaches~\citep{kong2023autoregressive} enable high-quality graph synthesis, capturing structural and semantic priors that enhance the versatility of Graph Foundation Models through generative pretraining.
    \item \textbf{Large Language Models}: Recently, LLMs have been adapted for graph data by encoding graphs into tokens~\citep{mao2024position, tang2024graphgpt}. This approach leverages the power of LLMs to process graphs, opening up new avenues for graph learning. While promising, this line of work raises open questions on how to design good tokenization schemes, inject structural priors, and mitigate hallucination. As GFMs evolve, the convergence of LLM-based approaches and GTs may play a central role in more areas.
\end{itemize}

However, several challenges in architectural design continue to limit their effectiveness. Below, we outline key areas that require further innovation.
\begin{itemize}
	\item \textbf{Dynamic Graph Modeling}. Real-world graphs are often dynamic, with nodes and edges evolving over time. Designing GFMs capable of handling dynamic graphs, especially in scenarios where the graph structure continuously updates, remains an area requiring further exploration.
	\item \textbf{Scalable Architecture Design}. Maintaining GFMs' effectiveness and efficiency becomes increasingly challenging as graph data scales to millions of nodes and edges. Future research can focus on developing scalable architectures that perform well on large-scale graphs without compromising on computational efficiency.
	\item \textbf{Balancing Global and Local Information}. Current graph learning architectures often focus on either local or global structural information, rather than integrating both. Developing methods that effectively combine global and local information in GFMs is a promising direction for performance gain.
	\item \textbf{Handling Complex Graph Structures}: Most existing graph models are designed for simpler graph structures, such as homogeneous graphs. However, many real-world graphs are heterogeneous or hypergraphs, which present additional challenges. Developing GFMs to process these complex structures remains challenging.
\end{itemize}

\subsection{Tuning, Adaptation and Compression}

GFMs for specific downstream applications requires a combination of adaptation strategies. This means that pre-trained models can be effectively fine-tuned for targeted tasks and efficiently deployed in various environments. The existing transfer strategies for GFMs are as follows:

\begin{itemize}
	\item \textbf{Fine-Tuning and Adaptation}: Fine-tuning is a common approach for applying pre-trained GFM to specific downstream tasks. By further training the model parameters on a small amount of task-specific data, fine-tuning enhances the model's performance on the target task. This can be achieved through full-parameter fine-tuning or by freezing most of the model's layers and only fine-tuning a few selected layers.
	\item \textbf{Model Compression}: In resource-constrained environments, model compression techniques are crucial for deploying GFMs. Compression methods include pruning, quantization, and knowledge distillation. These techniques aim to reduce the model's parameters or complexity, thereby lowering computational and storage requirements while retaining as much performance as possible.
\end{itemize}

Despite the advances in tuning, adaptation, and compression strategies for GFMs, several open issues persist that limit their effectiveness in applications.

\begin{itemize}
	\item \textbf{Generalizable Fine-Tuning Methods}. While fine-tuning is effective for specific tasks, developing methods that generalize across a wide range of downstream applications for GFMs remains challenging. The key issue is finding the right balance between retaining pretrained knowledge and adapting quickly to the unique requirements of new tasks.
	\item \textbf{Automated Fine-Tuning}. Future research should explore automated tools like AutoML~\citep{he2021automl} to streamline the GFM fine-tuning process. Automation has the potential to reduce human intervention, enhance efficiency, and produce more optimal tuning outcomes, making fine-tuning more accessible and effective.
	\item \textbf{Maintaining Performance After Compression}. Model compression techniques, such as pruning and quantization, often lead to performance degradation, particularly in complex graph tasks. A critical area of research is developing compression strategies on GFMs that minimize this loss of performance while ensuring that compressed models remain effective in practical applications.
	\item \textbf{Few- and Zero-Shot Scenatios}. In many real-world scenarios, labeled data is limited. Effectively fine-tuning pre-trained models in few-shot settings is a significant challenge. Exploring how pre-trained models can be leveraged for zero-shot or few-shot learning, particularly in data-scarce environments, will be an important focus for future GFM tuning strategies.
\end{itemize}

GFMs are emerging as a unified framework for graph-based learning, combining large-scale pretraining, GTs, and flexible fine-tuning to support diverse applications from social networks to scientific discovery. As the field matures, GFMs may become core components of graph learning and become a unified tool across graph tasks.

\section{Graph Reinforcement Learning}

Deep Reinforcement Learning (DRL) is a subfield of machine learning that integrates deep learning with reinforcement learning~\citep{landers2023deep}. DRL training agent to make decisions by learning optimal policy through interactions with the environment, using deep neural networks to handle high-dimensional state spaces and approximate complex value functions~\citep{milani2024explainable}. The sequential decision-making process can be modeled using a Markov Decision Process (MDP) or an extended MDP, such as a partially observable Markov decision process (POMDP) or Semi-MDP. Specifically, an MDP can be defined as $<S,A,T,R>$, where $S$ is state space, $A$ is the action space, $T$ is state transition probability, and $R$ is the reward value obtained after actor $a$ is performed. This approach enables the agent to maximize cumulative rewards over time in sequential decision-making tasks. 

Graph Reinforcement Learning (GRL) is a novel method that combines DRL with graph representation, enabling the solution of decision-making problems based on graph data~\citep{munikoti2023challenges}. In DRL problems, graph-structured data provides rich relational information, especially in multi-agent scenarios, which is unavailable in non-graph-structured states. Additionally, incorporating DRL into graph learning introduces a new training approach that differs from existing graph learning tasks, which often require substantial expert knowledge. DRL allows for learning without prior knowledge. We will introduce GRL from two aspects: graph-enhanced reinforcement learning and reinforcement learning enhanced graph learning.

\subsection{Graph Enhanced DRL}

In multi-agent DRL, a group of agents cooperate or compete to achieve a common goal, where the key is to understand the mutual interplay between agents. This architecture is widely used in various tasks, such as traffic signal control~\citep{devailly2021ig}, vehicle control systems~\citep{yang2023hierarchical}, and network routing~\citep{xu2024graph}. The communication between agents provides information about the states and environments of other agents. Various attention-based methods have been proposed to address communication issues in multi-agent systems. In G2ANet~\citep{liu2020multi}, the authors used a complete graph to model the relationships between agents in large-scale multi-agent scenarios and designed a two-stage attention model as the communication model. Similarly, in Graphocomm~\citep{shen2021graphcomm}, the authors applied a relational graph module to model explicit and implicit relationships through explicit and implicit relation layers. Specifically, explicit relationships are built from prior knowledge, while implicit relationships are learned through interactions among agents. In SRI-AC~\citep{zhang2021structural}, the authors deployed a Variational Autoencoder (VAE) model to predict interactions and learn a state representation from observational data.

In multi-task DRL, states and actions can vary significantly across tasks. However, this challenge can be effectively addressed by leveraging the generalization capabilities of GNNs, which facilitate efficient multi-source policy transfer learning in settings with state-action mismatches. In TURRET~\citep{kong2024mag}, GNNs are used to learn the intrinsic properties of agents, creating a unified state embedding space for all tasks. This approach enables TURRET to achieve more efficient transfer and stronger generalization across tasks, making it easily compatible with existing DRL algorithms.

\subsection{DRL Enhanced Graph Learning}

Most graph representation methods either use global pooling techniques to generate a single global representation vector, which overlooks the semantics of substructures, or rely on expert knowledge to extract local substructures~\citep{lanciano2024survey}. This often results in poor interpretability and limited generalization ability~\citep{ribeiro2021survey}. By utilizing reinforcement learning, it is possible to capture the most significant subgraphs without the need for expert knowledge~\citep{sun2021sugar}, offering a more flexible and adaptive approach to graph representation. By defining the update value $k$ within the MDP framework and using Q-learning to learn the optimal policy, where $k$ serves as the pooling ratio in top-k sampling. Similarly, the identification of an optimal subset can be formulated as a combinatorial optimization problem, where the goal is to select the best combination of elements from a larger set. Deep Q-learning can be applied to solve this problem by learning the value of each possible action and selecting the combination that maximizes the overall reward~\citep{kong2024mag}.

Optimizing the GNN framework and hyperparameters can significantly enhance learning capabilities. However, modifying these components requires extensive domain knowledge. Utilizing Neural Architecture Search (NAS) allows for the selection of suitable GNN models and their corresponding hyperparameters tailored to specific tasks. Due to the highly diverse nature of graph classification applications, designing data-specific pooling methods based on human expertise presents a significant challenge~\citep{gao2021graph}. To address this, PAS~\citep{wei2023neural} introduces a unified pooling framework that simplifies the design of the search space. Two variants, PAS-G and PAS-NE, are also proposed, each offering distinct approaches to improve the effectiveness of pooling in different scenarios.

\section{Federated Graph Learning}

Although graph learning techniques have significantly progressed across various domains in recent years, most existing graph neural networks still rely on centralized storage of large-scale graph data for training. However, growing concerns over data security and user privacy make this requirement impractical in real-world scenarios~\citep{liu2024federated}. In many practical applications, graph data is often distributed across multiple data holders, and due to privacy concerns and regulatory constraints, no party can directly access the data of others. This data distribution across different devices and owners, coupled with privacy concerns, makes it increasingly challenging to learn a global model while preserving data privacy~\citep{ge2024review}.

For instance, in the financial sector, a third-party company may need to train a graph learning model for multiple financial institutions to detect potential financial crimes and fraudulent activities. Each financial institution holds its local customer dataset, including demographic information and transaction records between customers. Based on this data, each institution forms a customer graph, where the edges represent transaction relationships. Due to strict privacy policies and industry competition, these local customer datasets cannot be directly shared with third-party companies or with other institutions. At the same time, there may be interactions between different financial institutions, representing structural information across institutions. The main challenge for the third-party company lies in training a graph learning model for financial crime detection based on both the local customer graphs and the structural information between institutions, without directly accessing the local data of each institution~\citep{fu2022federated}.

To address this challenge, federated learning (FL) has emerged as a promising distributed learning framework that tackles the issues of data isolation and privacy preservation~\citep{mcmahan2017communication}. FL allows participants to collaboratively train a global model without sharing their local data, and it has been widely applied to Euclidean data tasks, such as image classification, by aggregating model updates from multiple clients~\citep{li2020federated, kong2023fedawr}. However, the standard FL framework struggles to handle the complex relationships inherent in graph-structured data, limiting its direct application to graph learning scenarios~\citep{kong2024federated}.

In response, Federated Graph Learning (FGL) has emerged as a promising solution~\citep{tan2023federated, li2024openfgl, liu2025taimodel}. FGL combines the strengths of federated learning and graph neural networks, enabling multiple data holders to collaboratively train a graph learning model without sharing their local graph data. This method is particularly suitable to distributed graph data scenarios, such as analyzing customer transaction data across multiple financial institutions. Each institution holds a local customer graph, and the interactions between institutions form cross-institutional structural information. By integrating local graphs with inter-institutional structural information while ensuring data privacy, FGL can construct a global graph model, significantly enhancing the detection of financial crimes and other related tasks.

FGL can be further categorized into three types based on how graph structure information is distributed among participants: graph federated learning, subgraph federated learning, and graph structured federated learning. By incorporating structural information from graph data, FGL not only enhances distributed graph learning but also boosts performance in cross-institutional and cross-domain collaboration, while maintaining privacy protection~\citep{fu2022federated, liu2024federated}.

\subsection{Graph Federated Learning}

Graph federated learning is a natural extension of standard FL, where each participant locally holds graph structured data, and the global model is trained to perform graph level tasks. A typical application of graph federated learning is in AI4Science, such as drug discovery and molecular property prediction. In these cases, GNNs are used to study the graph structure of molecules, where molecules are represented as graphs, with atoms as nodes and chemical bonds as edges~\citep{li2024openfgl}. Each bio-pharmaceutical company has its private dataset, containing molecular nodes and their corresponding properties. Historically, commercial competition and data privacy concerns have hindered collaboration between these companies. However, with the graph federated learning framework, it is now possible to jointly train molecular structure models, enabling clients to collaborate on building robust models while preserving data privacy.

Formally, each client $k$ owns its private local graph data $\mathcal{D}_k=\left\lbrace \mathcal{G}_1, \mathcal{G}_2, \cdots \right\rbrace $, where each $\mathcal{G}_i=(\mathcal{V}, \mathcal{E})$ is a graph comprising a set of nodes $\mathcal{V}$ and a set of edges $\mathcal{E}$. Each client collaborates with others to train a robust global graph model $\mathcal{F}$ based on its local dataset $\mathcal{D}_k$, while ensuring that $\mathcal{D}_k$ remains local and private. A significant challenge in this setting is the heterogeneity of the data across clients, which may differ significantly in terms of node features and graph structures~\citep{xie2021federated}. This data heterogeneity can lead to severe model divergence during the federated process, thereby degrading the performance of the global model. To address this, various methods have been proposed to either train a single global model with better generalization or develop personalized models that can adapt to the unique characteristics of each client’s data~\citep{tan2023federated, wan2024federated}.

\subsection{Subgraph Federated Learning}

In subgraph federated learning, each participant's local data is treated as a subgraph of a larger, more comprehensive global graph~\citep{zhang2021subgraph}. Participants focus on using the nodes and edges within their subgraphs as training samples. Without directly sharing data, the participants collaboratively train a global graph model. Subgraph federated learning has broad application prospects in fields like social network analysis, recommendation systems, and financial risk control, where graph data plays a central role. It offers a solution to handle large-scale, distributed graph data while safeguarding privacy and improving model performance.

However, subgraph federated learning faces several challenges~\citep{kong2024mitigating}. First, each client only holds a subgraph of the original global graph, and due to privacy concerns and the isolated storage of data, each node can only access the information of neighboring nodes within its subgraph, lacking access to information from other clients. This missing cross-client information leads to biased node embeddings on each client, thereby reducing the performance of the graph learning model~\citep{baek2023personalized}. A key challenge is to reconstruct the missing cross-client information to accurately compute the node embeddings. Another challenge is that the same node may belong to multiple clients. In such cases, during collaborative training, the embeddings of overlapping nodes from different clients may originate from different embedding spaces. The core idea to address this is to learn global node embeddings based on the local embeddings of each client, and an overlapping instance alignment technique has been introduced to ensure the consistency of node embeddings across clients~\citep{yao2024fedgcn}.

\subsection{Graph Structured Federated Learning}

In real-world scenarios, clients are often interconnected. For instance, in traffic flow prediction tasks, sensor devices distributed across different geographical locations may have connections, and these links often contain rich information, such as spatial dependencies~\citep{ge2024review}.

In graph structured federated learning, each client is treated as a node, and together, all the clients form a graph $\mathcal{G} = (\mathcal{V}, \mathcal{E}) $, where $\mathcal{V}$ represents the set of clients and $\mathcal{E}$ represents the edges between them. This approach takes into account the topological structure between clients and uses GNNs to aggregate models based on these topological relationships~\citep{fu2022federated,}. It is important to note that the data on the client side is not necessarily graph-structured.

Typically, clients upload their local model parameters or local embeddings to the server, just as in standard FL. The server then aggregates the uploaded models using graph-based algorithms, taking into account the client graph $\mathcal{G}$, and finally distributes the updated model parameters back to the clients~\citep{meng2021cross, xing2022big, chen2022personalized}.

\section{Learning on Knowledge Graphs}
KGs represent structural relations between entities, storing human knowledge facts in intuitive graph structures and being towards human-level intelligence~\citep{liang2024survey,DBLP:journals/tnn/JiPCMY22}. Learning on KGs refers to combining the structured data represented in knowledge graphs with graph learning techniques~\citep{DBLP:conf/sigir/LiJLGGSWC21,DBLP:conf/icde/ChenWWZCLL24,10880109}. Although KGs have made significant success in various real-world fields, how to use these semantically structured data to assist graph learning still faces challenges. Firstly, KGs are often incomplete with missing links between entities, resulting in a degradation of the quality of the knowledge graph and impacting model performance. Thus, it poses challenges in the knowledge extraction process to construct high-quality KGs. Secondly, KGs grow exponentially due to the quick updating of information in the real world. Thus, it poses challenges in proposing efficient representation methods to handle the scalability of KGs.

\subsection{Knowledge Extraction}
Knowledge extraction (or knowledge acquisition) tasks can be divided into three categories: knowledge graph completion (KGC), relation extraction, and entity discovery~\citep{DBLP:journals/tnn/JiPCMY22}. KGC aims to predict missing links in existing KGs, and embedding-based ranking, relation path reasoning, rule-based reasoning, and meta relational learning are main categories for KGC. Compared to KGC, relation extraction and entity discovery are for discovering new knowledge (e.g., relations and entities) from the text. Relation extraction models utilize attention mechanisms, GCNs, adversarial training, reinforcement learning, deep residual learning, and transfer learning. KGC, relation extraction, and entity discovery are three separate tasks for knowledge extraction. In order to simplify development and be convenient for knowledge extraction, proposing a unified framework is one of the challenges in this field. For example, \citet{DBLP:conf/aaai/Han0S18} proposed a joint learning framework with mutual attention for data fusion between knowledge graphs and text, which simultaneously solves KGC and relation extraction from text. Additionally, knowledge extraction also face other challenges in KGC, relation extraction, and entity discovery as follows.

\begin{itemize}
	\item \textbf{Challenges of KGC:} As knowledge graphs are by nature incomplete, discovering new triples in knowledge graphs is the prime mission. The first challenge is that most existing KGC methods focus on one-hop reasoning but fail to capture multi-step relationships for complex reasoning. For example, embedding-based ranking methods usually use one-hop neighbour information to predict missing entities or relations~\citep{DBLP:conf/cikm/GuanJWC18,DBLP:journals/www/NiuFYLCLZ21,DBLP:conf/kdd/ZhangLDFY18}. However, these methods cannot consider the influence of multi-hop relationships and model complex relation paths in knowledge graphs. To solve this problem, relation path reasoning methods and rule-based methods attempt to leverage path information over the graph structure~\citep{DBLP:journals/tkde/ZhuoWZW24,DBLP:journals/nn/MeiYJCGHP24}. Another two challenges of KGC are that: 1) the long-tail phenomena exist in the relations, 2) and unseen triples usually emerge because of knowledge being dynamic in the real-world scenario. Some efforts have been made to target these two challenges, and meta-relational learning is a solution to them. The principle behind meta-relational learning is based on meta-learning and local graph structures. It encodes one-hop neighbors to capture the structural information with R-GCN and then takes the structural entity embedding for multi-step matching guided by meta-learning~\citep{DBLP:conf/emnlp/ChenZZCC19}. 
	
	\item \textbf{Challenges of relation extraction:} Relation extraction usually aims to extract unknown relational facts from plain text and add them into knowledge graphs. Traditional methods highly depend on feature engineering with prior knowledge to explore the inner correlation between features~\citep{DBLP:conf/wsdm/QuOHYZ19}. However, the main challenge of relation extraction is how to learn richer representations as possible. Thus, using deep neural networks like GCN can leverage relational knowledge in graphs to effectively extract relation~\citep{DBLP:journals/www/YingMZYPL23,DBLP:conf/aaai/Li0ZZ23,DBLP:conf/kdd/WeiL22}. For example, \citet{DBLP:conf/naacl/ZhangDSWCZC19} applied GCN for relation embedding in knowledge graphs for sentence-based relation extraction.
	
	\item \textbf{Challenges of entity discovery:} Entity recognition is the first step of entity discovery, which tags entities in text. Its technique has evolved from hand-craft to applying neural architectures such as LSTM-CNN methods, attention-based methods, K-BERT and so on \citep{ehrmann2023named}. Entity typing is the step after entity recognition. However, arranging a proper entity typing is not a simple task. Because entity typing not only includes coarse types, but also contains fine-grained types. And the latter is typically regarded as multi-class and multi-label classification. Thus, label noise is inevitably introduced to the entity typing process. The challenge of entity discovery is reducing label noise in entity typing. Embedding-based approaches have been proposed to tackle this problem. JOIE and ConnectE are embedding-based methods, which explore local typing and global triple knowledge structure information to enhance joint embedding learning ~\citep{hao2019universal,DBLP:conf/acl/ZhaoZXLW20}.   
\end{itemize}

\subsection{Knowledge Representation}
Knowledge representation learning (KGL) is a critical research issue of KGs, which paves the way for many downstream applications~\citep{DBLP:journals/ai/ZhengYW24,DBLP:conf/acl/WangZLL23}. Existing KGL focuses on modeling the semantic interaction of facts and utilizing external information. GNNs are introduced to learn the connectivity structure in KGs under an encoder-decoder framework. R-GCN proposes a relation-specific transformation to model the directed nature of knowledge graphs, and GCN acts as a graph encoder~\citep{schlichtkrull2018modeling}. SACN introduces weighted GCN to define the strength of two adjacent nodes with the same relation type, capturing the structural information within KGs~\citep{DBLP:conf/aaai/ShangTHBHZ19}. 

Although embedding representations can capture relationships between knowledge elements, it is challenging to infer explicit logical relationships directly from these vector representations. Designing knowledge representation models that can provide interpretable reasoning while maintaining efficient computational performance remains an important research direction. Additionally, a unified understanding of knowledge representation is less explored. The field of knowledge representation still faces numerous challenges that hinder its broader application. Below are some key areas that require further exploration:

\begin{itemize}
	\item \textbf{Dynamic Knowledge Representation:} Knowledge in the real world is constantly changing, and the temporal changes in knowledge, along with the emergence of new knowledge, require knowledge representations to have the capability to handle dynamic changes. Designing dynamic knowledge graphs or knowledge representation models that can update over time and perform real-time reasoning is a challenging problem that needs to be addressed.
	
	\item \textbf{Cross-Domain Knowledge Representation:} Many knowledge representation methods are often constructed for specific domains, making it difficult to represent and reason across domains. The concepts and relationships in different domains may vary significantly, and how to unify and integrate cross-domain knowledge remains a challenge.
\end{itemize}

\section{Knowledge-infused Graph Learning}
Graph learning has achieved significant success in knowledge graph tasks, such as knowledge extraction and knowledge representation~\citep{DBLP:journals/tkde/ZhuoWZW24,DBLP:journals/nn/MeiYJCGHP24}. In the domain of graph learning, models are operated under the assumption that all data is of high quality, and graph learning techniques are extended to enhance the performance of downstream tasks. However, challenges related to low-quality data and poor generalization when applying these methods are encountered, leading to subpar model performance on instances not seen during training~\citep{DBLP:journals/corr/abs-2302-08261}. To address these limitations, the concept of knowledge-infused graph learning has been introduced. This approach integrates external knowledge into various components of the graph machine learning pipeline, aiming for more accurate results. Knowledge can be categorized into formal scientific knowledge (e.g., established laws or theories that govern the behaviour of target variables) and informal experimental knowledge (e.g., well-known facts derived from long-term observations or human reasoning).

Compared to traditional graph machine learning techniques, incorporating human knowledge provides several advantages: (i) it allows for the successful integration of vast amounts of information, (ii) it enhances reliability in results, and (iii) it facilitates meaningful inferences, which can assist individuals lacking domain expertise. Knowledge-infused graph learning methods can be seen as employing knowledge in two ways: as prompts or as augmentation. “Knowledge as prompts” refers to using external knowledge to guide model learning and decision-making processes. Conversely, “knowledge as augmentation” involves analyzing a graph \(\mathcal{G=(V,E)}\) in conjunction with a relevant knowledge dataset \(\mathcal{D}\).  The primary objectives of this approach are twofold: to learn a function that generates vector representations for graph elements, \(f:(\mathcal{G,D}) \rightarrow \textbf{Z}\), capturing both the structure and semantics of the graph alongside the knowledge stored in the knowledge database, and to retrieve direct and precise answers (or explanations) from the dataset \(\mathcal{D}\) to address user queries.

Despite the considerable progress of knowledge-infused graph learning in downstream applications such as large language models pre-training and drug discovery, challenges remain, including knowledge database composition and effective knowledge integration. These challenges will be discussed in the following sections.

\subsection{Knowledge as Prompt}
The prompt-based fine-tuning method originates from natural language processing and has been widely used to help pre-trained language models adapt to various downstream tasks~\citep{liu2023pre}. Knowledge, as the initial guidance or prompt in the graph learning process, aims to assist the model in understanding the structure and relationships within the data, thereby generating more effective learning outcomes. There are two main types of prompt methods:

\begin{itemize}
	\item \textbf{Discrete Graph Prompt:} In the discrete graph prompt method, prompts are explicitly marked or modified within the graph structure through specific nodes, edges, or subgraphs. These prompts are typically predefined or interpretable, directly embedded into the graph's structure, and the prompt information is discrete.
	
	\item \textbf{Continuous Graph Prompt:} In the continuous graph prompt method, prompts are embedded into the graph's features or edge weights in the form of vectors or continuous values. The prompt information does not directly change the graph structure but subtly influences the model's behavior by adjusting the graph representation or the parameters of the graph neural network.
\end{itemize}

Although knowledge as a prompt has achieved some success in the field of graph learning, several challenges still remain:

\begin{itemize}
	\item \textbf{Limited Fine-Tuning Methods:} In the domain of graph neural networks, existing research on graph prompts is still limited, and there is a lack of a general approach to cater to different downstream tasks. Some approaches have leveraged prompt-based fine-tuning of pre-trained models via edge prediction.~\citep{sun2023all,liu2023graphprompt}.
	
	\item \textbf{Generalization:} Some current research methods design specific prompt functions for pre-training tasks, but these approaches are often tailored to particular pre-trained GNN models and may not perform well on others. Future research should explore a more general prompt-based fine-tuning method such as GPF~\citep{fang2024universal}.
\end{itemize}

\subsection{Knowledge as Augmentation}
Typical graph analysis applications include node role identification, personalised recommendation, social healthcare, academic network analysis, graph classification and epidemic trend study~\citep{DBLP:journals/bdcc/ChenZP21,DBLP:conf/aaai/ZhangCNC18}. Regarding external knowledge as data augmentation helps incorporate with graph learning for more accurate performance. For example, knowledge-augmented graph learning methods have achieved more precise drug discovery with limited training data~\citep{DBLP:journals/corr/abs-2302-08261}. To illustrate it well, we will discuss challenges from these four directions: incorporating knowledge in preprocessing, pretraining, training, and interpretability.

\begin{itemize}
	\item \textbf{Incorporating Knowledge in Preprocessing:} Taking drug discovery as an example, GNN models employ a limited number of message-passing steps \(L\), that is less than the diameter of the biomedical graph \(diam(G)\). It means that entities that are more than \(L\) hops apart will not receive messages about each other, which hampers the prediction of biomedical properties that are heavily dependent on global features. Thus, how to design new features from external domain knowledge for input data \(\mathcal{X}\) is a challenge. \citet{DBLP:journals/natmi/ShenZZWQTJC21} developed a molecular feature-generation model to map molecular and fingerprint features. In addition, some studies propose capturing additional domain knowledge from external biomedical knowledge datasets to enrich the semantics of entities’ embeddings for precise drug-target interaction prediction~\citep{DBLP:conf/ijcai/LyuGTLZZ21,DBLP:journals/tbd/ShenZYLZ22}.
	
	\item \textbf{Incorporating Knowledge in Pretraining:} One of the main challenges in the use of graph learning techniques is the construction of an appropriate objective function for model training, especially in the presence of limited supervision labels in biomedical datasets. One notable example of this approach is the PEMP mode, which identifies a group of useful knowledge to enhance the capability of GNN models in capturing relevant information through pretraining~\citep{DBLP:conf/cikm/SunCMHLMML22}. Additionally, knowledge from large-scale knowledge databases can be exploited as well. 
	
	\item \textbf{Incorporating Knowledge in Training} The sparsity of training data and complexity of process modeling are significant limitations. To address these challenges, leveraging external knowledge to guide the message-passing process is a promising solution. This approach involves leveraging domain knowledge to adjust the internal processes of GNN models~\citep{hasebe2021knowledge}. However, the optimal method for adaptively utilising auxiliary knowledge by considering the uncertainty remains an open research question.
	
	\item \textbf{Incorporating Knowledge in Interpretability:} The ability to understand and explain how a model works is a crucial step towards Trustworthy Artificial Intelligence (TAI). This highlights the need to improve interpretability and trustworthiness from the user's perspective. By utilising machine-readable domain knowledge represented in KGs, relevant knowledge can be highlighted for downstream applications. \citet{yu2021sumgnn} extracted local subgraphs in an external KG to extract biomedical knowledge based on a self-attention mechanism. Nevertheless, there remains a vast scope for further research in the area of advanced interpretability, with the aim of providing more holistic and adaptable explanations, such as advanced reasoning and question-answering capabilities~\citep{zhu2022neural,wang2019multi}.
\end{itemize}

\section{Quantum Graph Learning}
Quantum mechanics is increasingly being explored to enhance graph learning techniques, giving rise to the emerging field of Quantum Graph Learning (QGL)~\citep{yu2023quantum}.
Traditional graph learning methods, which focus on understanding and predicting relationships in graph-structured data, face challenges like scalability and capturing intricate node interactions.
By integrating principles of quantum mechanics (e.g., superposition and entanglement), QGL aims to revolutionize this process.
Quantum algorithms have the potential to process complex graph structures more efficiently, enabling faster computations and deeper insights into the connectivity patterns.
As QGL is still in its infancy, researchers are actively exploring quantum models and algorithms that can be adapted to current graph learning frameworks.

Message-passing GNNs face significant limitations when it comes to efficiently capturing long-range dependencies between nodes.
As the distance between nodes increases, the efficiency of information transfer diminishes, leading to issues such as oversmoothing, where node representations become indistinguishable.
When representing data in quantum states, the quantum coherence or entanglement transfers long-distance information with a faster time.
Quantum graph learning provides a potential solution by leveraging quantum phenomena like \textit{quantum coherence} and \textit{entanglement}~\citep{tuysuz2021hybrid, ai2022decompositional}.
Quantum coherence allows quantum systems to exist in multiple states simultaneously. In this framework, entanglement plays a pivotal role by facilitating the transfer of information between distant nodes through non-local effects, allowing long-range dependencies to be captured with much greater precision and efficiency.

Non-Euclidean graph-structured data have rigorous requirements in hardware space and speed. Thus, classical hardware limitations induce bottlenecks for graph learning.
Qubits carry exponentially more information than bits, and quantum algorithms solve graphs with lower time complexity \citep{tabi2020quantum, pelofske2021decomposition, wang2022quantum}.
Additionally, quantum search algorithms, such as Grover's algorithm \citep{grover1996fast}, have lower time complexity compared to their classical counterparts, making it possible to retrieve information from graph structures much faster.
Thus, quantum hardware has exponentially accelerated access to data in hardware, and thus eliminates the fundamental hardware bottleneck for graph learning.

QGL harnesses these quantum advantages to intensify traditional graph learning processes.
Classical methods (GNNs) are time-consuming to deal non-Euclidean graph-structured data, and thus cannot utilize hardware resources sufficiently.
The use of superposition allows for the simultaneous exploration of multiple graph paths, while entanglement facilitates capturing complex, multi-node relationships with a level of intricacy that classical methods cannot easily achieve. As a result, QGL has the potential to drastically improve the efficiency and accuracy of tasks.
Researchers are exploring quantum-enhanced graph neural networks (QGNNs) \citep{chen2022novel, mernyei2022equivariant} and quantum-inspired models \citep{henry2021quantum, kishi2022graph, yan2022towards} to adapt these concepts to existing machine learning frameworks.

Graph learning has a property of black-box.  Existing models rarely provide a human-understandable explanation for tasks, and less explainable output results of models are discussed.
Quantum mechanics explains the existence of our physical world. With the help of quantum theory, the explainable or interpretable problem will be resolved in graph learning.
Graph learning criticized for its "black-box" nature, where models provide little insight into their decision-making processes. This lack of transparency can hinder trust and understanding, especially in critical applications like healthcare and finance. Quantum mechanics, known for its precise explanation of physical phenomena, offers a promising solution to this issue. By leveraging principles such as superposition and entanglement, QGL models can create features and relationships that are more comprehensible to humans. Quantum algorithms can highlight specific quantum states or transitions that correspond to patterns in the data offering clearer explanations for predictions.

%% file: chapters/9-Conclusion.tex
Graph learning has emerged as a rapidly evolving and influential field within AI, offering powerful tools for modeling complex relationships and extracting valuable insights from both graph-structured and non-graph-structured data. This survey has provided a comprehensive review of state-of-the-art graph learning methods, covering key areas such as scalable graph learning, temporal graph learning, multimodal graph learning, generative graph learning, explainable graph learning, and responsible graph learning. Additionally, we have explored emerging topics, including graph foundation models, graph reinforcement learning, federated graph learning, learning on knowledge graphs, knowledge-infused graph learning, and quantum graph learning.

Recent research trends in graph learning, beyond the commonly discussed areas like scalability, temporal dynamics, multimodality, generative models, explainability, and responsibility, focus on several innovative directions with strong ties to AI advancements. Beyond the above-mentioned emerging topics, one key trend, for example, is causal graph learning and neurosymbolic integration. Causal graph learning seeks to go beyond mere pattern recognition by uncovering the underlying causal structures within graph data, which is crucial for robust decision-making in AI systems. This direction aims to improve generalization, interpretability, and counterfactual reasoning capabilities, which are especially important in high-stakes domains such as healthcare, finance, and policy-making. Meanwhile, the fusion of graph learning with symbolic reasoning, often referred to as neurosymbolic graph learning, enables more structured, logic-aware AI by embedding domain knowledge and logical constraints into neural graph models, allowing for more trustworthy and knowledge-driven inference.

Another emerging trend is interactive and agent-centric graph learning, where the learning process is embedded within an interactive environment or driven by the actions of autonomous agents. This includes applications in reinforcement learning on graphs, active learning with graph feedback, and the use of graph-based world models for embodied AI or decision-making agents. Such models are increasingly relevant for robotics, scientific discovery, and multi-agent systems, where understanding the dynamics of relational structures is essential.

In the coming years, graph learning is poised to play a transformative role across a broad spectrum of real-world applications, driven by growing demands for, e.g., relational understanding, structured reasoning, and efficient learning. In domains such as healthcare, biology, and drug discovery, graph learning will enable more precise modeling of complex systems like molecular interactions and disease progression pathways, accelerating personalized medicine and therapeutic innovations. In transportation, urban computing, and supply chain management, graph-based models will become central to optimizing infrastructure, predicting disruptions, and enabling adaptive, intelligent planning.

Moreover, the integration of graph learning into large-scale systems such as recommender engines, fraud detection, cybersecurity, and knowledge-intensive search will expand as organizations seek AI solutions that can reason over rich, dynamic relational data. The increasing availability of heterogeneous and dynamic data sources will further fuel the adoption of graph-based AI in environmental monitoring, social behavior analysis, and scientific discovery. As these applications mature, future graph learning systems will likely become more autonomous, context-aware, and tightly coupled with decision-making processes, forming the backbone of next-generation intelligent systems.